%% file: egpaper_for_review.tex
\documentclass[10pt,twocolumn,letterpaper]{article}

\usepackage{cvpr}
\usepackage{times}
\usepackage{epsfig}
\usepackage{graphicx}
\usepackage{amsmath}
\usepackage{amssymb}

\usepackage{subcaption}
\usepackage{bbding}
\usepackage{pifont}
\usepackage[normalem]{ulem}
\usepackage{multirow}
\usepackage[table,xcdraw]{xcolor}
\usepackage{bm}
\usepackage{algorithm}
\usepackage[noend]{algpseudocode}
\usepackage{etoolbox}
\usepackage{symbols}
\usepackage{enumerate}
\usepackage[symbol,multiple]{footmisc}
\usepackage[export]{adjustbox}

\newcommand{\model}{\mbox{Two Body Network}}
\newcommand{\modelshort}{\mbox{\sc TBONE}}

\newcommand{\bV}{\bm{V}}
\renewcommand{\tilde}{\widetilde}

\usepackage[pagebackref=true,breaklinks=true,letterpaper=true,colorlinks,bookmarks=false]{hyperref}

\cvprfinalcopy 

\def\cvprPaperID{3820} 

\ifcvprfinal\fi
\begin{document}

\title{Two Body Problem: Collaborative Visual Task Completion}

\author{Unnat Jain$^{1}$\thanks{~indicates equal contributions.}\hspace{1.5mm}\footnotemark[2]
\and
Luca Weihs$^{2}$\footnotemark[1]
\and
Eric Kolve$^{2}$
\and
Mohammad Rastegari$^{2,4}$
\and
Svetlana Lazebnik$^{1}$
\and
Ali Farhadi$^{2,3,4}$
\and
Alexander Schwing$^{1}$
\and
Aniruddha Kembhavi$^{2}$
}

\twocolumn[{
\renewcommand\twocolumn[1][]{#1}
\maketitle
\vspace*{-0.5cm}
\centering
\includegraphics[width=.93\linewidth]{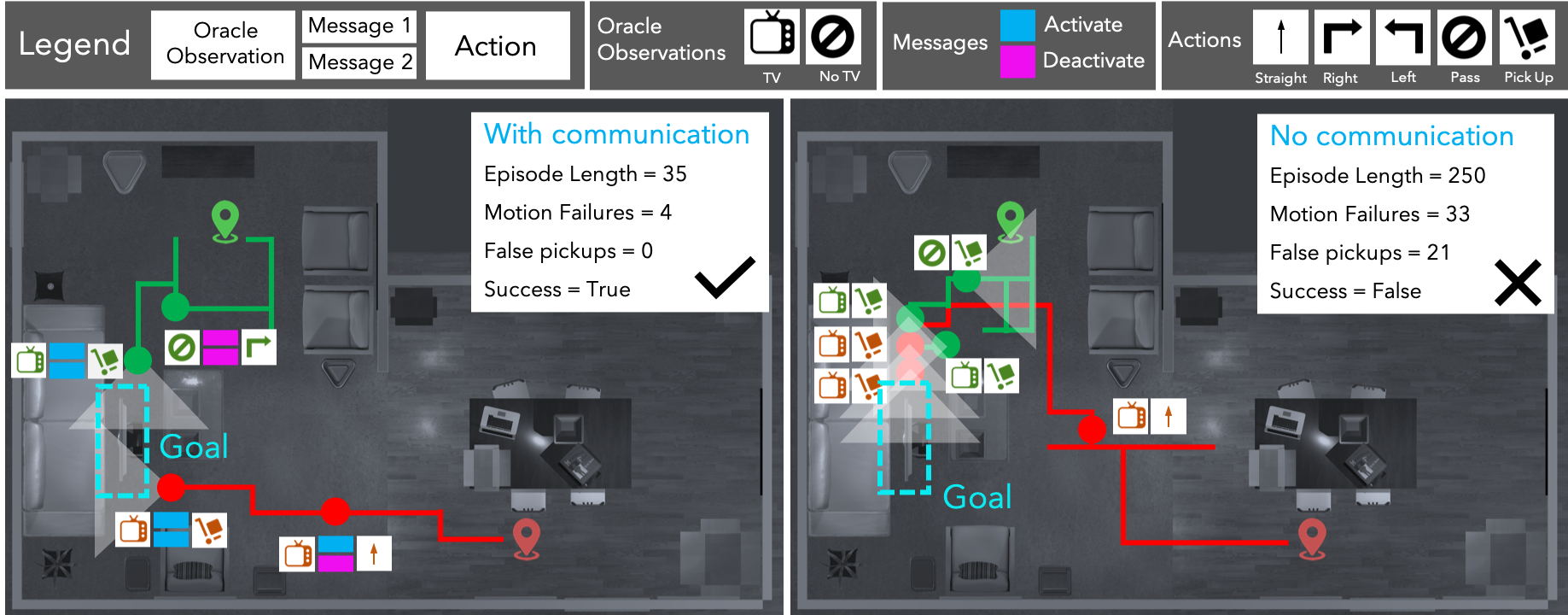}
\vspace{-.2cm}
\captionof{figure}{
Two agents learn to successfully navigate through a previously unseen environment to find, and jointly lift, a heavy TV. Without learned communication, agents attempt many failed actions and pickups. With learned communication, agents send a message when they observe or when they intend to interact with the TV. The agents also learn to grab the opposite ends of the TV and coordinate to do so.}
\label{fig:teaser1}
\vspace*{0.3cm}
}]

\maketitle

\input{abs}

\vspace{-0.5cm}
\input{intro}
\input{rel}
\input{app}

\input{exp}

\input{conc}

\noindent\textbf{Acknowledgements:} This material is based upon work supported in part by the National Science Foundation under Grants No.\ 1563727, 1718221, 1637479, 165205, 1703166, Samsung, 3M, Sloan Fellowship, NVIDIA Artificial Intelligence Lab, Allen Institute for AI, Amazon, AWS Research Awards and Thomas \& Stacey Siebel Foundation.

{\small
\bibliographystyle{ieee}
\bibliography{alex}
}
\clearpage
\input{appendix}
\end{document}

%% file: abs.tex
\begin{abstract}
\vspace*{-0.1cm}
Collaboration is a necessary skill to perform tasks that are beyond one agent's capabilities. Addressed extensively in both conventional and modern AI, multi-agent collaboration has often been studied in the context of simple grid worlds. We argue that there are inherently visual aspects to collaboration which should be studied in visually rich environments. A key element in collaboration is communication that can be either explicit, through messages, or implicit, through perception of the other agents and the visual world. Learning to collaborate in a visual environment entails learning (1) to perform the task, (2) when and what to communicate, and (3) how to act based on these communications and the perception of the visual world. In this paper we study the problem of learning to collaborate directly from pixels in AI2-THOR and demonstrate the benefits of explicit and implicit modes of communication to perform visual tasks. Refer to our project page for more details: \url{https://prior.allenai.org/projects/two-body-problem}
\end{abstract}
\footnotetext[1]{indicates equal contributions.}
\footnotetext[2]{work partially done as an intern at Allen Institute for AI}

%% file: intro.tex
\section{Introduction}
\vspace{-0.2cm}
Developing collaborative skills is known to be more cognitively demanding than learning to perform tasks independently.  
In AI, multi-agent collaboration has been studied in more conventional~\cite{GilesICABS2002, KasaiSCIA2008, BratmanCogMod2010, MeloMAS2011} and modern settings~\cite{LazaridouARXIV2016, FoersterNIPS2016, SukhbaatarNIPS2016, GuptaAAMAS2017,LoweNIPS2017,MordatchAAAI2018}. These studies have mainly been performed on grid-worlds and have factored out the role of perception in collaboration. 

In this paper we argue that there are aspects of collaboration that are inherently visual. Studying collaboration in simplistic environments does not permit to observe the interplay between perception and communication, which is necessary for effective collaboration. Imagine moving a piece of furniture with a friend. Part of the collaboration is rooted in explicit communication 
through exchanging messages,
and some part of it is done through implicit communication
through interpreting perceivable cues about the other agent’s behavior. If you see your friend going around the furniture to grab it, you would naturally stay on the opposite side to avoid toppling it over. Additionally, communication and collaboration should be considered jointly with the task itself. The way you communicate, either explicitly or implicitly, in a soccer game is very different from when you move furniture. This suggests that factoring out perception and studying collaboration in isolation (grid-world) might not result in an ideal outcome.

In short, learning to perform tasks collaboratively in a visual environment entails joint learning of (1) how to perform tasks in that environment, (2) when and what to communicate, and (3) how to act based on implicit and explicit communication. In this work, we develop one of the first frameworks that enables the study of explicitly and implicitly communicating agents collaborating together in a photo-realistic environment. 

To this end we consider the problem of finding and lifting bulky items, ones which cannot be lifted by a single agent. While conceptually simple, attaining proficiency in this task requires multiple stages of
communication.
The agents must search for the object of interest in the environment (possibly communicating their findings to each other), position themselves appropriately (for instance, opposing each other), and then lift the object simultaneously. If the agents position themselves incorrectly, lifting the object will cause it to topple over. Similarly, if the agents pick up the object at different time steps, they will not succeed.

To study this task, we use the AI2-THOR virtual environment~\cite{KolveARXIV2017}, a photo-realistic, physics-enabled environment of indoor scenes used in past work to study single agent behavior. We extend AI2-THOR to enable multiple agents to communicate and interact.

We explore collaboration along several modes: 
(1) The benefits of communication for spatially constrained tasks (\eg, requiring agents to stand across one another while lifting an object) \vs unconstrained tasks.
(2) The ability of agents to implicitly and explicitly communicate to solve these tasks. 
(3) The effect of the expressivity of the communication channel on the success of these tasks. 
(4) The efficacy of these developed communication protocols on known environments and their generalizability to new ones.
(5) The challenges of egocentric visual environments \vs grid-world settings.

We propose a \model, or \modelshort, for modeling the policies of agents in our environments. \modelshort\ operates on a visual egocentric observation of the 3D world, a history of past observations and actions of the agent, as well as messages
received from other agents in the scene. At each time step, agents go through two rounds of communication, akin to sending a message each and then replying to messages that are received in the first round.
\modelshort\ is trained with a warm start using a variant of DAgger~\cite{RossAISTATS2011}, followed by a minimization of a sum of an A3C loss and a cross entropy loss between the agents actions and the actions of an expert policy. 

We perform a detailed experimental analysis of the impact of communication using metrics including accuracy, number of failed pickup actions, and episode lengths. Following our above research questions, our findings show that:
(1) Communication clearly benefits both constrained and unconstrained tasks but is more advantageous for  constrained tasks.
(2) Both explicit and implicit communication are exploited by our agents and both are beneficial, individually and jointly.
(3) For our tasks, large vocabulary sizes are beneficial.
(4) Our agents generalize well to unseen environments.
(5) Abstracting our environments towards a grid-world setting improves accuracy, confirming our notion that photo-realistic visual environments are more challenging than grid-world like settings. This is consistent with findings by past works for single agent scenarios.

Finally we interpret the explicit mode of communication between agents by fitting logistic regression models to the messages to predict the values such as oracle distance to target, next action, \etc, and find strong evidence matching our intuitions about the usage of
messages between agents.

%% file: rel.tex
\section{Related Work}

We now review related work in the directions of visual navigation, 
navigation and language, visual multi-agent reinforcement learning (RL), and virtual learning environments employed in past works to evaluate algorithms.

\begin{figure}[t]
\centering
\includegraphics[width=\linewidth]{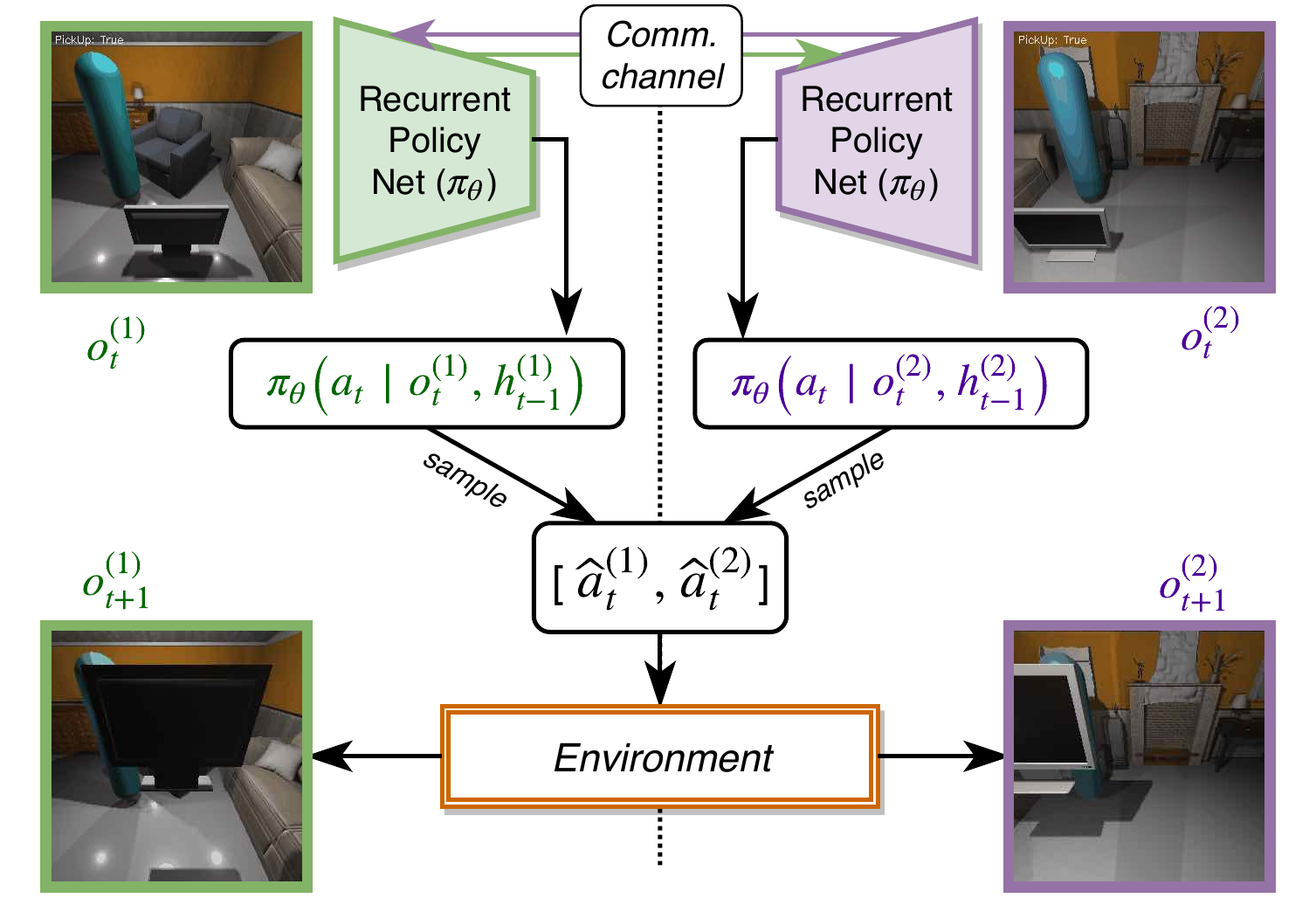}
\caption{A schematic depicting the inputs to the policy network. An agent's policy operates on a partial observation of the scene's state and a history of previous observations, actions, and messages received.}
\label{fig:overview}
\vspace{-0.5cm}
\end{figure}

\begin{figure*}[t!]
\centering
\vspace{-0.3cm}
\includegraphics[height=5.8cm]{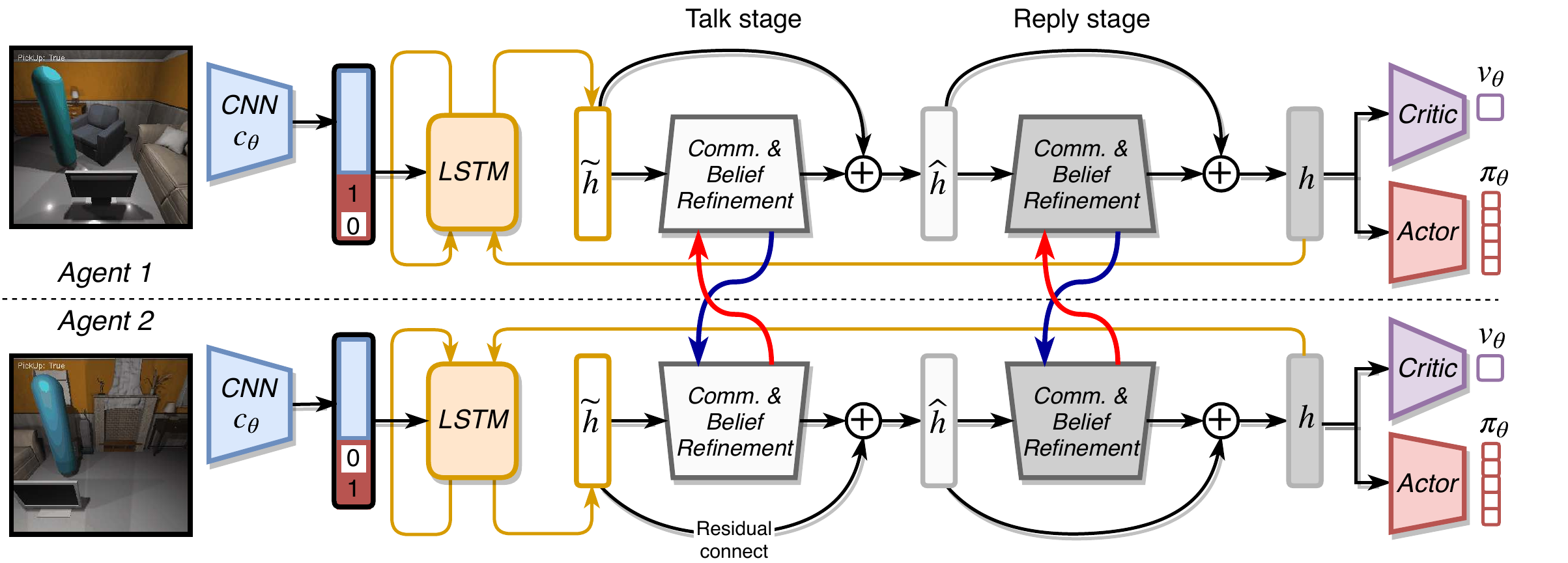}
\vspace{-0.3cm}
\caption{Overview of our \modelshort\ architecture for collaboration.
}
\label{fig:arch}
\vspace{-0.6cm}
\end{figure*}

\noindent\textbf{Visual Navigation:} A large body of work focuses on visual navigation, \ie, locating a target using only visual input. Prominent early map-based navigation methods~\cite{kim1999symbolic,borenstein1989realtime,borenstein1991the,oriolo1995online} use a global map to make decisions. More recent approaches~\cite{sim2006autonomous,wooden2006a,davison2003real,tomono20063d,kidono2002autonomous,royer2005outdoor} reconstruct the map on the fly. Simultaneous localization and mapping~\cite{tomasi1992shape,frahm2016structurefrommotion,thorpe2000structure,cadena2016past,smith1986on,smith1986estimating}  consider mapping in isolation. Upon having obtained a map of the environment, planning methods~\cite{CannyMIT1988,KavrakiRA1996,Lavalle2000} yield a sequence of actions to achieve the goal. Combinations of joint mapping and planning have also been discussed~\cite{elfes1989using,kuipers1991byun,konolige2010viewbased,fraundorfer2012visionbased,aydemir2013active}. Map-less methods~\cite{haddad1998reactive,lenser2003visual,remazeilles2004robot,saeedi2006visionbased,phillips2016fast,ZhuARXIV2016,GuptaCVPR2017} often formulate the task as obstacle avoidance given an input image or reconstruct a map implicitly. Conceptually, for visual navigation, we must learn a mapping from visual observations to actions which influence the environment. Consequently the task is well suited for an RL formulation, a perspective which has become popular recently~\cite{oh2016control,abel2016exploratory,chen2015deepdriving,daftry2016learning,giusti2016he,kahn2017plato,toussaint2003learning,mirowski2017learning,bhatti2016playing,brahmbhatt2017deepnav,zhang2016deep,duan2016rl2,GuptaCVPR2017,ZhuARXIV2017,gupta2017unifying}. Some of these approaches compute actions from observations directly while others attempt to explicitly/implicitly reconstruct a map.

Following recent techniques, our proposed approach also uses RL for visual navigation. While our proposed approach could be augmented with explicit or implicit maps, our focus is upon multi-agent communication. In the spirit of factorizing out orthogonal extensions from the model, we defer such extensions to future work.

\noindent\textbf{Navigation and Language:} Another line of work has focused on communication between humans and virtual agents. These methods more accurately reflect real-world scenarios since humans are more likely to interact with an agent using language rather than abstract specifications. Recently Das \etal~\cite{DasCVPR2018,DasECCV2018} and Gordon \etal~\cite{GordonCVPR2018} proposed to combine question answering with robotic navigation. Chaplot \etal~\cite{ChaplotARXIV2017}, Anderson~\etal~\cite{anderson2018vision} and Hill \etal~\cite{HillARXIV2017} propose to guide a virtual agent via language commands.

While language directed navigation is an important task, we consider an orthogonal direction where multiple agents need to collaboratively solve a specified task. Since visual multi-agent RL is itself challenging, we refrain from introducing natural language complexities. 
Instead, in this paper, we are interested in developing a systematic understanding of the utility and character of communication strategies developed by multiple agents through RL.

\noindent\textbf{Visual Multi-Agent Reinforcement Learning:} 
Multi-agent systems result in non-stationary environments posing significant challenges. Multiple approaches have been proposed over the years to address such concerns~\cite{TanICML1993,TesauroNIPS2004,TampuuPLOS2017,FoersterARXIV2017}. Similarly, a variety of settings from multiple cooperative agents to multiple competitive ones have been investigated~\cite{LauerICML2000,Panait2005,MatignonIROS2007,Busoniu2008,OmidshafieiARXIV2017,GuptaAAMAS2017,LoweNIPS2017,FoersterAAAI2018,MordatchAAAI2018}.

Among the plethora of work on multi-agent RL, we want to particularly highlight work by Giles and Jim~\cite{GilesICABS2002}, Kasai \etal~\cite{KasaiSCIA2008}, Bratman \etal~\cite{BratmanCogMod2010}, Melo \etal~\cite{MeloMAS2011}, Lazaridou \etal~\cite{LazaridouARXIV2016}, Foerster \etal~\cite{FoersterNIPS2016}, Sukhbaatar \etal~\cite{SukhbaatarNIPS2016} and Mordatch and Abbeel~\cite{MordatchAAAI2018}, all of which investigate the discovery of communication and language in the multi-agent setting using maze-based tasks, tabular setups, or Markov games. For instance, Lazaridou \etal~\cite{LazaridouARXIV2016} perform experiments using a referential game of image guessing, Foerster \etal~\cite{FoersterNIPS2016} focus on switch-riddle games, Sukhbaatar \etal~\cite{SukhbaatarNIPS2016} discuss multi-turn games on the MazeBase environment~\cite{SukhbaatarARXIV2015}, and Mordatch and Abbeel~\cite{MordatchAAAI2018} evaluate on a rectangular environment with multiple target locations and tasks. Most recently, Das \etal~\cite{das2018tarmac} demonstrate, especially in grid-world settings,  the efficacy of targeted communication where agents must learn to whom they should send messages.

Our work differs from the above body of work in that we consider communication for visual tasks, \ie, our agents operate in rich visual environments rather than a grid-like maze, a tabular setup or a Markov game. We are particularly interested in investigating how communication and perception support each other.

\noindent\textbf{Reinforcement Learning Environments:} As just discussed, our approach is evaluated on a rich visual environment. Suitable environment simulators are AI2-THOR~\cite{KolveARXIV2017}, House3D~\cite{WuARXIV2018}, HoME~\cite{BrodeurARXIV2017}, MINOS~\cite{SavvaARXIV2017Minos} for Matterport3D~\cite{Chang3DV2017Matterport} and  SUNCG~\cite{SongCVPR2017SUNCG}. Common to these environments is  the goal of modeling real world living environments with substantial visual diversity. This is in contrast to other RL environments such as the arcade environment~\cite{bellemare2013the}, Vizdoom~\cite{kempka2016vizdoom}, block towers~\cite{lerer2016learning}, Malmo~\cite{johnson2016the}, TORCS~\cite{wymann2013torcs}, or MazeBase~\cite{SukhbaatarARXIV2015}. Of these environments, we chose  AI2-THOR as it was easy to extend, provides high fidelity images, and has interactive physics enabled scenes, opening up interesting multi-agent research directions beyond this current work.


%% file: app.tex
\renewcommand{\compresslist}{%
\setlength{\itemsep}{1pt}%
\setlength{\parskip}{0pt}%
\setlength{\parsep}{0pt}%
}
\section{Collaborative Task Completion}

\begin{figure}
  \centering
  \includegraphics[width=\linewidth]{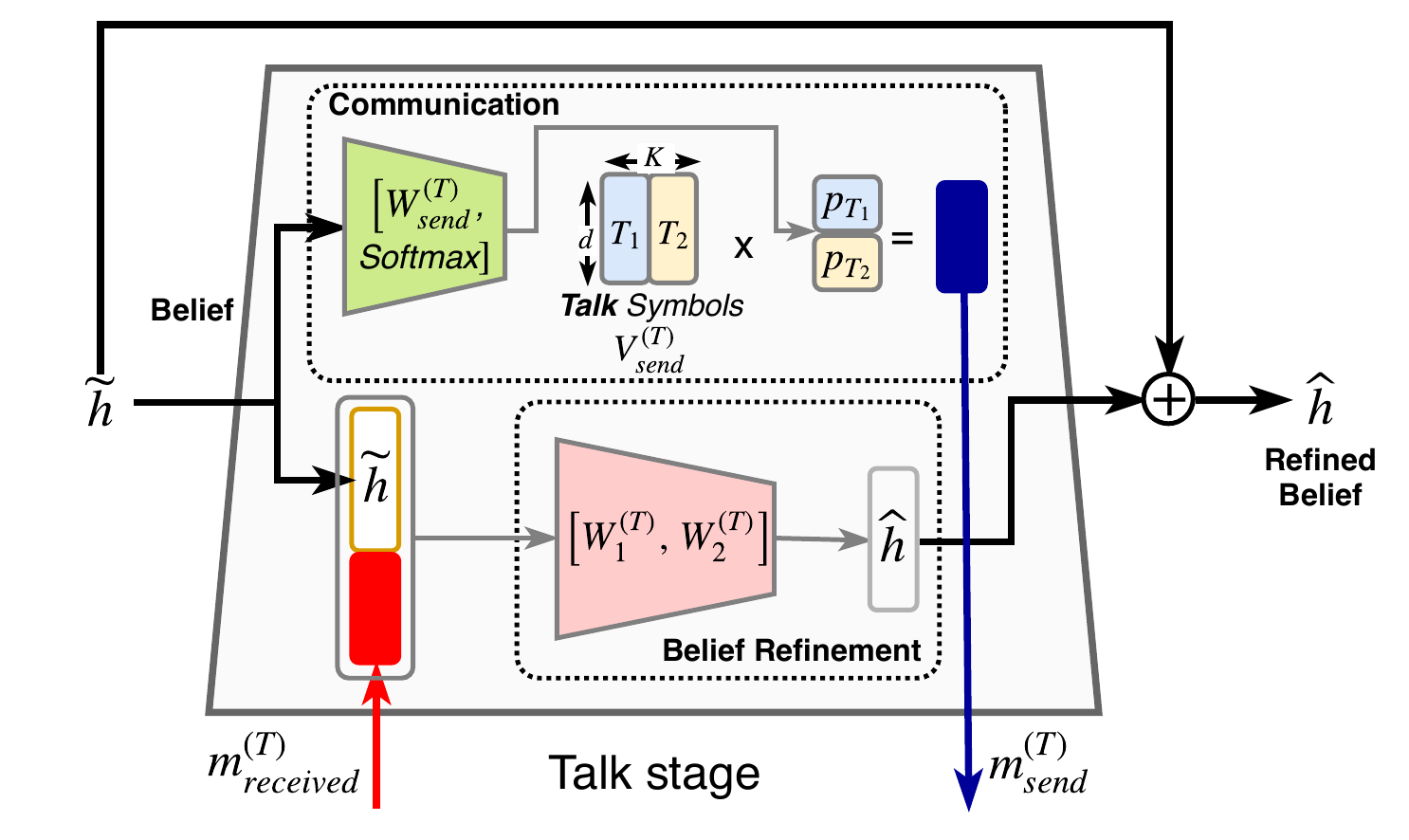}
  \vspace{-0.7cm}
  \caption{Communication and belief refinement module for the \textit{talk} stage (marked with the superscript of $(T)$) of explicit communication. Here our vocab. is of size $K=2$.}
  \label{fig:comm}
  \vspace{-0.4cm}
  \end{figure}

We are interested in understanding how two agents can learn, from pixels, to communicate so as to effectively and collaboratively solve a given task.
To this end, we develop a task for two agents which consists of two components, each tailored to a desirable skill for indoor agents. The components are:
(1) visual navigation, which the agents may solve independently, but which may also benefit from some collaboration; and (2) jointly synchronized interaction with the environment, which typically requires collaboration to succeed. The choice of these components stems from the fact that navigating to a desired position in an environment or to locate a desired object is a quintessential skill for an indoor agent, and synchronized interaction is fundamental to understanding any collaborative multi-agent setting.

We first discuss the collaborative task more formally, then detail the components of our network, \modelshort, used to complete the task.

\subsection{Task: Find and Lift Furniture}

We task two agents to lift a heavy target object in an environment, a task that cannot be completed by a single agent owing to the weight of the object. The two agents as well as the target object are placed at random locations in a randomly chosen AI2-THOR living room scene. Both agents must locate the target, approach it, position themselves appropriately, and then simultaneously lift it.

To successfully complete the task, both agents perform actions over time according to the same learned policy (Fig.~\ref{fig:overview}). Since our agents are homogeneous, we share the policy parameters for both agents. Previous works \cite{GuptaAAMAS2017,MordatchAAAI2018} have found this to train agents more efficiently. For an agent, the policy operates on (1) an ego-centric observation of the environment as well as a previous history of (a) observations, (b) actions taken by the agent, and (c)  messages sent by the other agent. At each time step, the two agents process their current observations and then perform two rounds of explicit communication. Each round of communication involves each of the agents sending a single message to the other. The agents also have the ability to watch the other agent (when in view) and possibly even recognize their actions over time, thereby using implicit communication as a means of gathering information. 

More formally, an agent perceives the scene at time $t$ in the form of an image $o_t$ and chooses its action $a_t\in\cA$ by computing a policy, \ie, a probability distribution $\pi_\theta(a_t|o_t, h_{t-1})$, over all actions $a_t\in\cA$.
In our case, the images $o_t$ are first-person views obtained from AI2-THOR. Following classical recurrent models, our policy  leverages information computed in the previous time-step via the representation $h_{t-1}$. The set of available actions $\cA$ consists of the five options \textsc{MoveAhead}, \textsc{RotateLeft}, \textsc{RotateRight}, \textsc{Pass}, and \textsc{Pickup}. The actions \textsc{MoveAhead}, \textsc{RotateLeft}, and \textsc{RotateRight} allow the agent to navigate. To simplify the complexities of continuous time movement we let a single \textsc{MoveAhead} action correspond to a step of size 0.25 meters, a single \textsc{RotateRight} action correspond to a 90 degree rotation clockwise, and a single \textsc{RotateLeft} action correspond to a 90 degree rotation anti-clockwise. The \textsc{Pass} action indicates that the agent should stand-still and \textsc{Pickup} is the agent's attempt to pick up the target object. Critically, the \textsc{Pickup} action has the desired effect only if three preconditions are met, namely both agents must (1) be within 1.5 meters of the object and be looking directly at it, (2) be a minimum distance away from one another, and (3) carry out the \textsc{Pickup} action simultaneously. Note that asking agents to be at a minimum distance from one another amounts to adding specific constraints on their relative spatial layouts with regards to the object and hence requires the agents to reason about such relationships. This is akin to requiring the agents to stand across each other when they pick up the object. The motivation to model spatial constraints with a minimum distance constraint is to allow us to easily manipulate the complexity of the task. For instance, setting this minimum distance to 0 loosens the constraints and only requires agents to meet two of the above preconditions.

In our experiments, we train agents to navigate within and  interact with 30 indoor environments. Specifically, an episode is considered successful if both agents navigate to a known object and, jointly, lift it within a fixed number of time steps. As our focus is the study of collaboration and not primarily object recognition, we keep the sought object, a television, constant. Importantly, environments as well as the agents' start locations and the target object location are randomly assigned at the start of each episode. Consequently, the agents must learn to (1) search for the target object in different environments,
(2) navigate towards it,
(3) stay within the object's vicinity until the second agent arrives,
(4) coordinate that both agents are apart from each other by at least the specified distance,
and (5) finally and jointly perform the pickup action.


Intuitively, we expect the agents to perform better on this task if they can communicate with each other. We conjecture that explicit communication will allow them to both signal when they have found the object and, after navigation, help coordinate when to attempt a \textsc{Pickup}, whereas implicit communication will help to reason about their relative locations with regards to each other and the object. To measure the impact of explicit and implicit means of communication in the given task, we train models with and without message passing as well as by making agents (in)visible to one another. Explicit communication would seem to be especially important in the case where implicit communication isn't possible. Without any communication, there seems to be no better strategy than for both agents to independently navigate to the object and then repeatedly try \textsc{Pickup} actions in the hope that they will be, at some point, in sync. The expectation that such a policy may be forthcoming gives rise to one of our metrics, namely the count of failed pickup events among both agents in an episode. We discuss metrics and results in Section~\ref{sec:exp}.

\begin{table}[]
\centering
\resizebox{0.98\columnwidth}{!}{
\begin{tabular}{ccccc}
\hline
Data & Accuracy & Reward & 
\begin{tabular}[c]{@{}c@{}}Missed\\ pickups\end{tabular} & \begin{tabular}[c]{@{}c@{}}Unsuccess.\\ pickups\end{tabular} 
\\ \hline
Visual & 59.0 $\pm$4.0 & -2.7 $\pm$0.3 & 0.3 $\pm$0.09 & 2.9 $\pm$0.8 \\ \hline
Visual$+$depth & 65.7 $\pm$3.9 & -2.0 $\pm$0.3 & 0.4 $\pm$0.1 & 3.2 $\pm$0.9\\ \hline
Grid-world & \textbf{78.2 $\pm$3.4} & \textbf{-0.6 $\pm$0.2} & \textbf{0.1 $\pm$0.05} & \textbf{0.7 $\pm$0.1} \\ \hline
\end{tabular}
}
\caption{Effect of adding oracle depth as well as moving to a grid-world setting on unseen scenes, \emph{Constrained} task.}
\label{tab:depth_gridworld}
\vspace{-0.5cm}
\end{table}

\subsection{Network Architecture} \label{sec:network-architecture}
In the following we describe the learned policy (actor) $\pi_\theta(a_t|o_t, h_{t-1})$ and value (critic) $v_\theta(o_t,h_{t-1})$ functions for each agent in greater detail. See Fig.~\ref{fig:arch} for a high level visualization of our network structure. 
Let $\theta$ represent a catch-all parameter encompassing all the learnable weights in \modelshort. At the $t$-th timestep in an episode we obtain as an agent's observation, from AI2-THOR, a $3\times 84 \times 84$ RGB image $o_t$ which is then processed by a four layer CNN $c_\theta$ into the 1024-dimensional vector $c_\theta(o_t)$. Onto $c_\theta(o_t)$ we append   an 8-dimensional learnable embedding $e$ which, unlike all other weights in the model, is not shared between the two agents. This agent embedding $e$ gives the agents the capacity to develop distinct complementary strategies. The concatenation of $c_\theta(o_t)$ and $e$ is fed, along with historical embeddings from time $t-1$, into a long-short-term-memory (LSTM)~\cite{HochreiterNC1997} cell resulting in a 512-dimensional output vector $\tilde{h}_t$
capturing
the beliefs of the agent given its prior history and most recent observation. Intuitively, we now would like the two agents to refine their beliefs via communication before deciding on a course of action. We consider this process in several stages (Fig.~\ref{fig:comm}).

\noindent\textbf{Communication:}
We model communication by allowing the agents to send one another a $d$-dimensional vector derived by performing soft-attention over a vocabulary of a fixed size $K$. More formally, let $\bW_{\text{send}}\in\mathbb{R}^{K\times 512}$, $\bm{b}_{\text{send}}\in\bR^{512}$, and $\bV_{\text{send}} \in\mathbb{R}^{d \times K}$ be (learnable) weight matrices with the columns of $\bV_{\text{send}}$ representing our vocabulary. Then, given the representation $\tilde{h}_t$ described above, the agent computes soft-attention over the vocabulary producing the message $m_{\text{send}} =\bV_{\text{send}} \ \text{softmax}( \bW_{\text{send}}\ \tilde{h}_t + \bm{b}_{\text{send}}) \in \mathbb{R}^d,$
which is relayed to the other agent.

\noindent\textbf{Belief Refinement:}
Given the agents' current beliefs $\tilde{h}_t$ and the message $m_{\text{received}}$ from the other agent, we model the process of refining one's beliefs given new information using a two layer fully connected neural network with a residual connection. In particular, $\tilde{h}_t$ and $m_{\text{received}}$ are concatenated, and new beliefs $\hat{h}_t$ are formed by computing
$\hat{h}_t = \tilde{h}_t + \text{ReLU}(\bW_2\ \text{ReLU} (\bW_1 [\tilde{h}_t\ ;\ m_{\text{received}}] + \bm{b}_{1}) + \bm{b}_{2}),$
where $\bW_1\in\bR^{512\times(512 + d)}$, $\bm{b}_{1},\bm{b}_{2}\in\bR^{512}$, and $\bW_2\in\bR^{512\times512}$ are learnable weight matrices. We set the value of $d$ to 8.

\noindent\textbf{Reply and Additional Refinement:}
The above step is followed by one more round of communication and belief refinement by which the representation $  \hat{h}_t$ is transformed into $h_t$. These additional stages have new sets of learnable parameters including a new vocabulary matrix. Note that, unlike in the standard LSTM framework where $\tilde{h}_{t-1}$ would be fed into the cell at time $t$, we instead give the LSTM cell the refined vector $h_{t-1}$.

\noindent\textbf{Linear Actor and Critic:}
Finally the policy and value functions are computed as
$\pi_\theta(a_t|o_t, h_{t-1}) = \text{softmax}(\bW_{\text{actor}}\ h_{t} + \bm{b}_{\text{actor}})$, and $v_\theta(o_t,h_{t-1}) = \bW_{\text{critic}}\ h_{t} + \bm{b}_{\text{critic}}$
where $\bW_{\text{actor}}\in\bR^{5 \times 512}$, $\bm{b}_{\text{actor}}\in\bR^{5}$, $\bW_{\text{critic}}\in\bR^{1\times 512}$, and $\bm{b}_{\text{critic}}\in\bR^1$ are learned.

\begin{figure*}
    \centering
    \begin{tabular}{@{\hskip1pt}c@{\hskip4pt}c@{\hskip4pt}c@{\hskip4pt}c@{\hskip0pt}}
    \includegraphics[height=4cm]{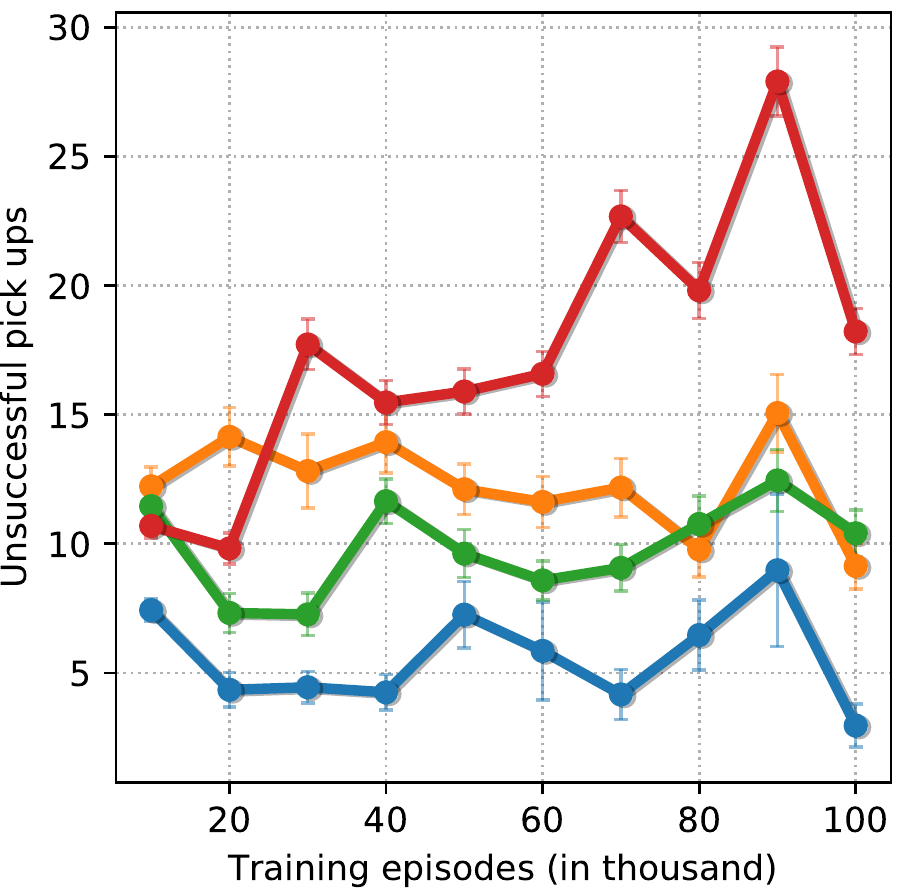}&
    \includegraphics[height=4cm]{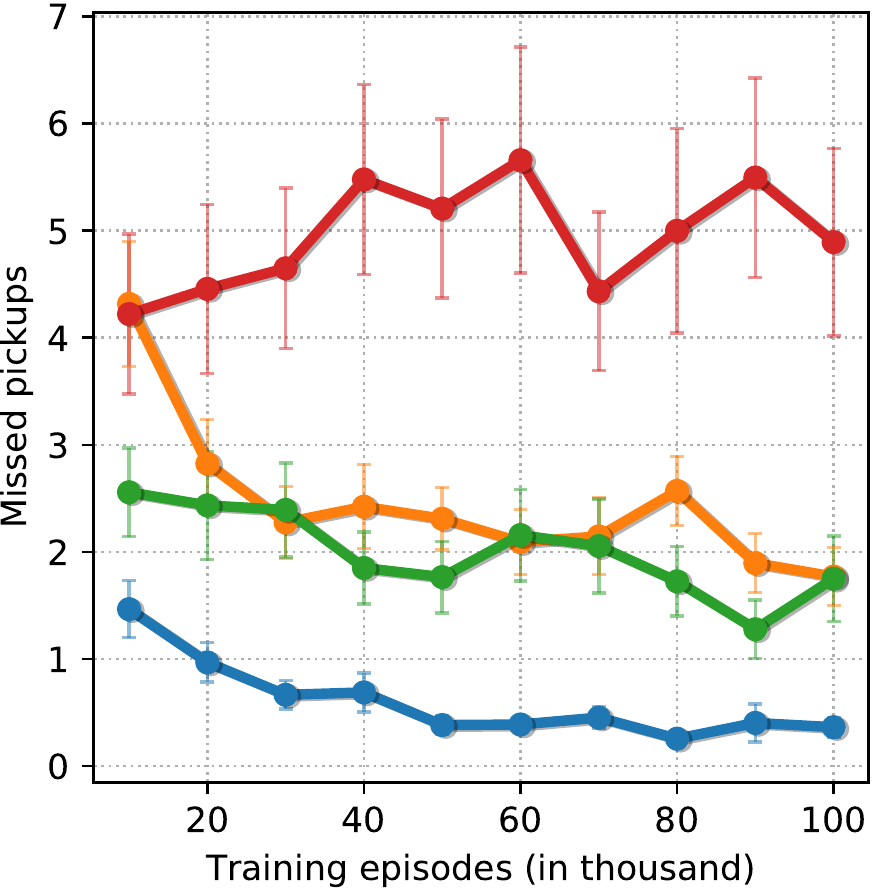}&
    \includegraphics[height=4cm]{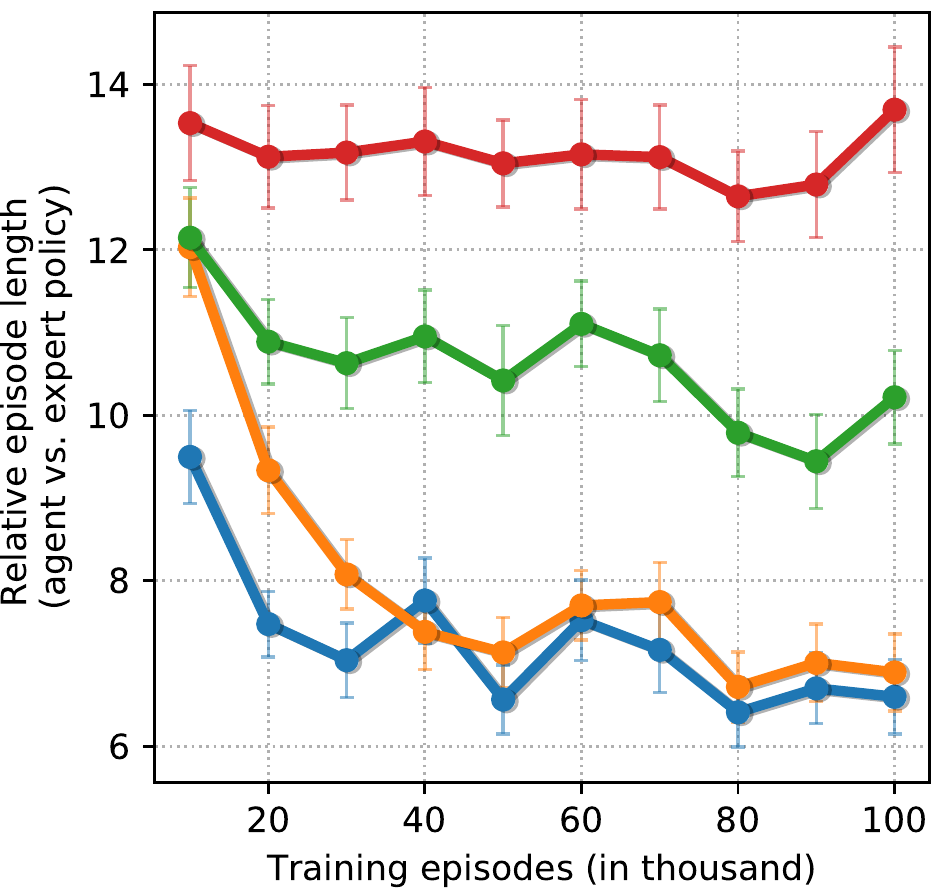}&
    \includegraphics[height=4cm]{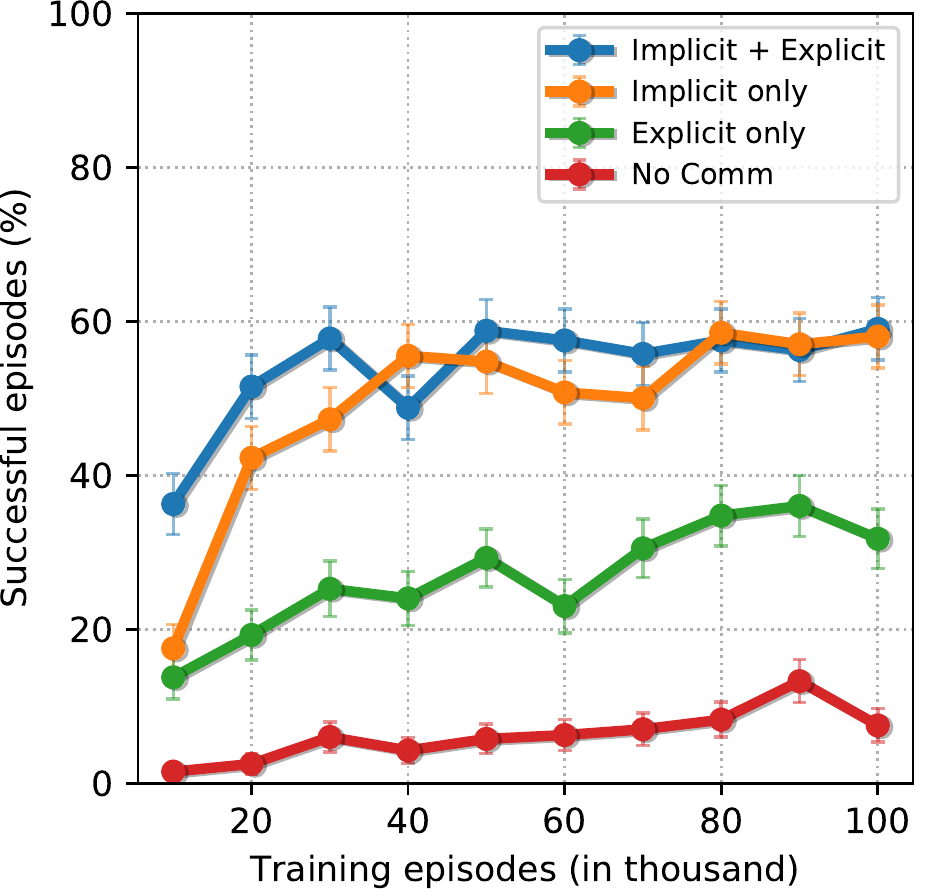}
    \end{tabular}
    \vspace*{-2mm}
    \caption{Unseen scenes metrics (\emph{Constrained} task): (a) Failed pickups (b) Missed pickups (c) Relative ep. len (d) Accuracy.}
    \label{fig:eval_metrics}
    \vspace{-0.5cm}
\end{figure*}

\subsection{Learning} \label{sec:learning}
Similar to others~\cite{DasCVPR2018,GuptaCVPR2017,DasARXIV2016,visdial_rl}, we found  training of our agents from scratch to be infeasible when using a pure reinforcement learning (RL) approach, \eg, with asynchronous actor critic (A3C)~\cite{MnihARXIV2016}, even in simplified settings, without extensive reward shaping. Indeed, often the agents must make upwards of 60  actions to navigate to the object and  will only successfully complete the episode and receive a reward if they jointly pick up the object. This setting of extremely sparse rewards is a well known failure mode of standard RL techniques. Following the above prior work, we use a ``warm-start'' by training with a variant of DAgger~\cite{RossAISTATS2011}. We train our models online using imitation learning for 10,000 episodes with actions for episode $i$ sampled from the mixture $(1-\alpha_{i}) \pi_{\theta_{i-1}} + \alpha_{i}\pi^*$ where $\theta_{i-1}$ are the parameters learned by the model up to episode $i$, $\pi^*$ is an expert policy (described below), and $\alpha_i$ decays linearly from $0.9$ to $0$ as $i$ increases. 
This initial warm-start  allows the agents to learn a policy for which rewards are far less sparse, allowing traditional RL approaches to be applicable. Note that our expert supervision only applies to the actions, there is no supervision for how agents should communicate. Instead the agents must learn to communicate in such a way that would increase the probability of expert actions. 

After the warm-start period, trajectories are sampled purely from the agent's current policy and we train our agents by minimizing the sum of an A3C loss, and a cross entropy loss between the agents' actions and the actions of an expert policy. The A3C and cross entropy losses here are complementary, each helping correct for a deficiency in the other. Namely, the gradients from an A3C loss tend to be noisy and can, at times, derail or slow training; the gradients from the cross entropy loss are 
noise free and thereby stabilize training. A pure cross entropy loss however fails to sufficiently penalize certain undesirable actions. For instance, diverging from the expert policy by taking a \textsc{MoveAhead} action when directly in front of a wall should be more strongly penalized than when the area in front of the agent is free as the former case may result in damage to the agent; both these cases are penalized equally by a cross entropy loss. The A3C loss, on the other hand, accounts for such differences easily so long as they are reflected by the rewards the agent receives. 

We now describe the expert policy. If both agents can see the TV, are within 1.5 meters of it, and are at least a given minimum distance apart from one another then the expert action is to \textsc{Pickup} for both agents. Otherwise given a fixed scene and TV position we obtain, from AI2-THOR, the set $T=\{t_1,\ldots,t_m\}$ of all positions (on a grid with square size $0.25$ meters) and rotations within 1.5 meters of the TV from which the TV is visible. Letting $\ell_{ik}$ be the length of the shortest path from the current position of agent $i\in\{0,1\}$ to $t_k$ we then assign each $(t_j,t_k)\in T\times T$ the score $s_{jk} = \ell_{0j}+\ell_{1k}$. We then compute the lowest scoring tuple $(s, t)\in T \times T$ for which $s$ and $t$ are at least a given minimum distance apart and assign agent 0 the expert action corresponding to the first navigational step along the shortest path from agent 0 to $s$ (and similarly for agent 1 whose expert goal is $t$).

Note that our training strategy and communication scheme can be extended to more than two agents. We defer such an analysis to future work, a careful analysis of the two-agent setting being an appropriate first step.

\noindent \textbf{Implementation Details.}
Each model was trained for 100,000 episodes. Each episode is initialized in a random train (seen) scene of AI2-THOR. Rewards provided to the agents are: 1 to both agents for a successful pickup action, constant -0.01 step penalty to discourage long trajectories,
-0.02 for any failed action (\eg, running into a wall) and -0.1 for a failed pickup action. Episodes run for a maximum of 500  steps (250 steps for each agent) after which the episode is considered failed.

%% file: exp.tex
\section{Experiments} \label{sec:exp}

In this section, we present our evaluation of the effect of communication towards collaborative visual task completion.
We first briefly describe the multi-agent extensions made to AI2-THOR, the environments used for our analysis, the two tasks used as a test bed and metrics considered.
This is followed by a detailed empirical analysis of the tasks. We then provide a statistical analysis of the explicit communication messages used by the agents to solve the tasks, which sheds light on their content. Finally we present qualitative results.

\noindent \textbf{Framework and Data.} 
We extend the AI2-THOR environment to support multiple agents that can each be independently controlled.  In particular, we extend the existing initialization action to accept
an \texttt{agentCount} parameter allowing an arbitrarily large number
of agents to be specified. When additional agents are spawned, each is visually depicted as a capsule
of a distinct color. This allows agents to observe each other's
presence and impact on the environment, a form of implicit communication. We also provide a parameter to render agents invisible to one another, which allows us to study the benefits of implicit communication.
Newly spawned agents have the full capabilities of a single agent, being able to interact with the environment by, for example, picking up and opening objects.
These changes are publicly available with AI2-THOR v1.0.
We consider the 30 AI2-THOR living room scenes for our analysis,
since they are the largest in terms of floor area and also contain a large amount of furniture. We train on 20 and test on the 20 seen scenes as well as the remaining 10 unseen ones.

\begin{figure}
    \centering
    \begin{tabular}{@{\hskip0pt}c@{\hskip2pt}c@{\hskip0pt}}
    \includegraphics[height=4cm]{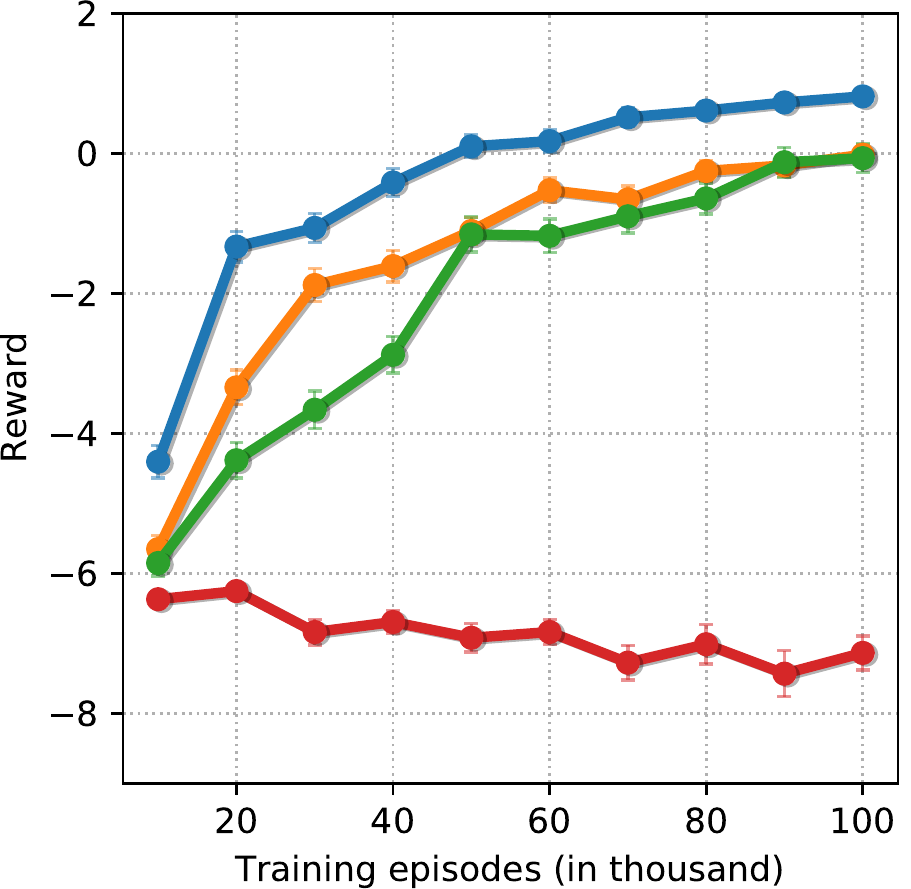}&
    \includegraphics[height=4cm]{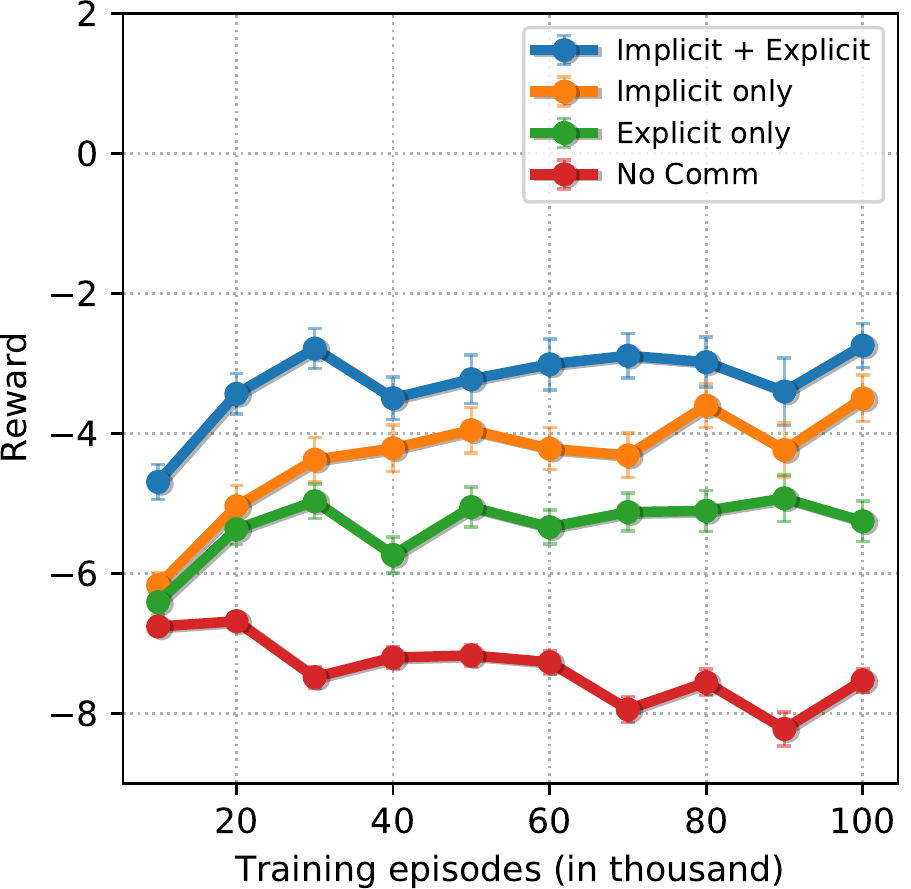}
    \end{tabular}
    \vspace*{-2mm}
    \caption{Reward \vs training episodes on the \emph{Constrained} task. (left) Seen scenes (right) Unseen scenes.}
    \label{fig:seen_unseen}
    \vspace{-0.5cm}
\end{figure}

\noindent \textbf{Tasks.} \label{subsec:tasks}
We consider two tasks, both requiring the two agents to simultaneously pick up the TV in the environment:
(1) \textit{Unconstrained}: No constraints are imposed here with regards to the locations of the agents with respect to each other.
(2) \textit{Constrained}: The
agents must be at least 8 steps from each other (akin to requiring them to stand across each other when they pick up the object).
Intuitively, we expect the \emph{Constrained} setting to be more difficult than the \emph{Unconstrained}, since it requires the agents to spatially reason about themselves and objects in the scene.
For each of the above tasks, we train 4 variants of \modelshort, resulting from switching explicit and implicit communication on and off. Switching off implicit communication amounts to rendering the \emph{other} agent invisible.

\noindent \textbf{Metrics.} \label{subsec:metrics}
We consider the following metrics:
(1) \textit{Reward}, 
(2) \textit{Accuracy}: \% successful episodes, 
(3) Number of \textit{Failed pickups}, 
(4) Number of \textit{Missed pickups}: where both agents could have picked up the object but did not, 
(5) \textit{Relative episode length}: relative to an oracle.
These metrics are aggregated over 400 random initializations (Unseen scenes: 10 scenes $\times$ 40 inits, Seen scenes: 20 scenes $\times$ 20 inits).
Note that accuracy alone isn't revealing enough. Na\"ive agents that wander around and randomly pick up objects will eventually succeed. Also, agents that correctly locate the TV and then keep attempting a pickup in the hope of synchronizing with the other agent will also succeed. Both these cases will however do poorly on the other metrics.

\noindent \textbf{Quantitative analysis.} \label{subsec:quant}
All plots and metrics referenced  in this section contain 
90\% confidence intervals. 

Fig.~\ref{fig:eval_metrics} compares the four metrics: Accuracy, Failed pickups, Missed pickups, and Relative episode length for unseen scenes and the \emph{Constrained} task.
With regards to accuracy, explicit+implicit communication fares only moderately better than implicit communication, but the need for explicit communication is dramatic in the absence of an implicit one.
But when one considers all metrics, the benefits of having both explicit and implicit communication are clearly visible. The number of failed and missed pickups is lower, while episode lengths are a little better than just using implicit communication.
The differences between just explicit \vs just implicit also shrink when looking at all metrics together.
However, across the board, it is clear that communicating is advantageous over not communicating.

Fig.~\ref{fig:seen_unseen} shows the rewards obtained by the 4 variants of our model on seen and unseen environments for the \emph{Constrained} task. While rewards on seen scenes are unsurprisingly higher, the models with communication do generalize well to unseen environments.
Adding the two means of communication is more beneficial than either and far better than not having any means of communication. Interestingly just implicit communication fares better than just explicit, on accuracy.

\begin{figure}
    \centering
    \begin{tabular}{@{\hskip0pt}c@{\hskip2pt}c@{\hskip0pt}}
    \includegraphics[height=4cm]{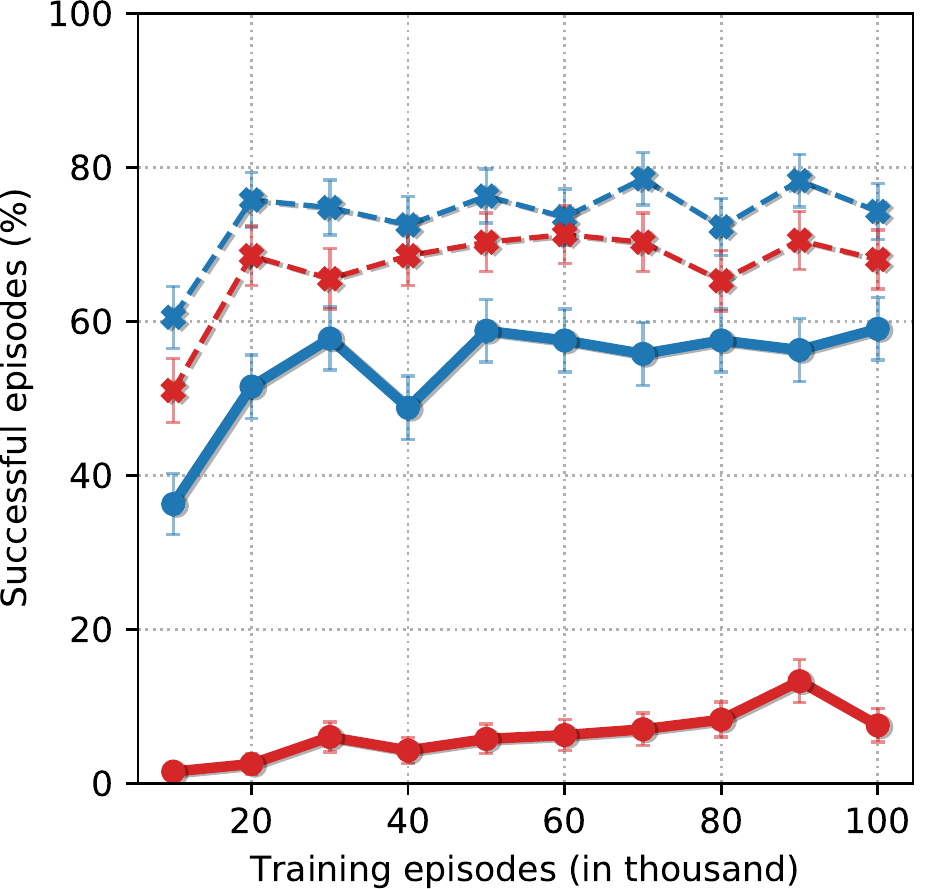}&
    \includegraphics[height=4cm]{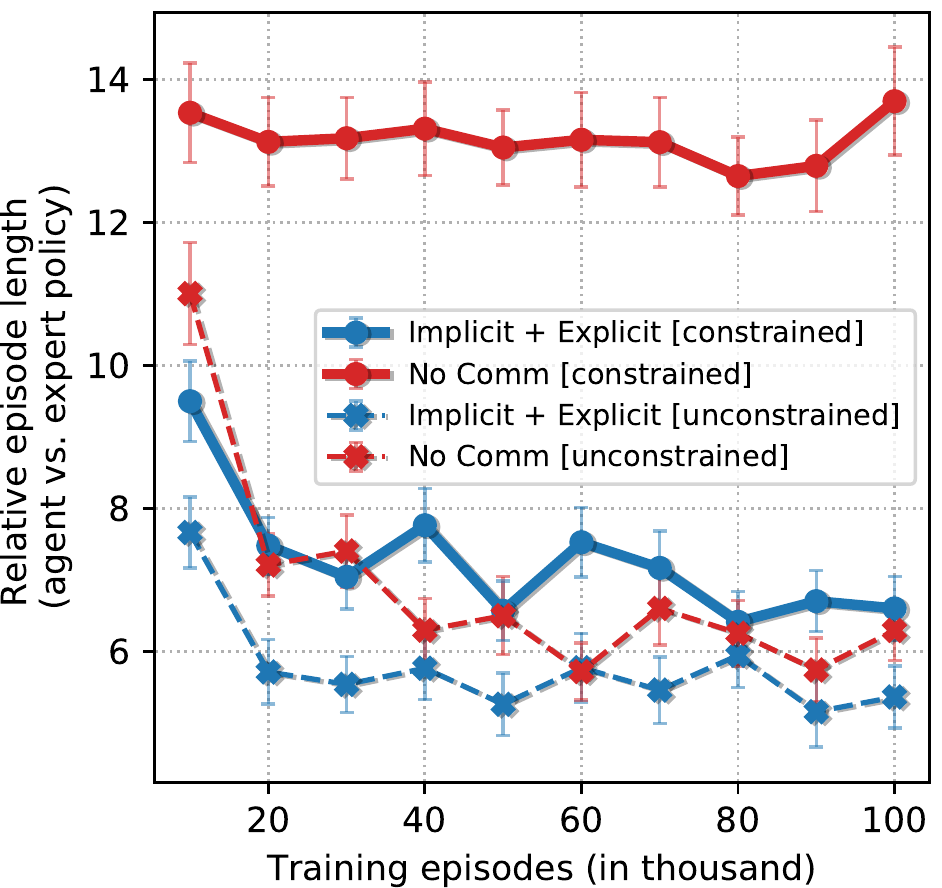}\\
    \end{tabular}
    \vspace*{-2mm}
    \caption{\emph{Constrained} \vs \emph{unconstrained} task (on unseen scenes): (left) Accuracy, (right) Relative episode length.}
    \label{fig:con_uncon}
    \vspace{-0.5cm}
\end{figure}

\begin{figure*}[t]
    \centering
    \begin{tabular}{@{\hskip3pt}c@{\hskip3pt}c}
    \includegraphics[height=4.5cm]{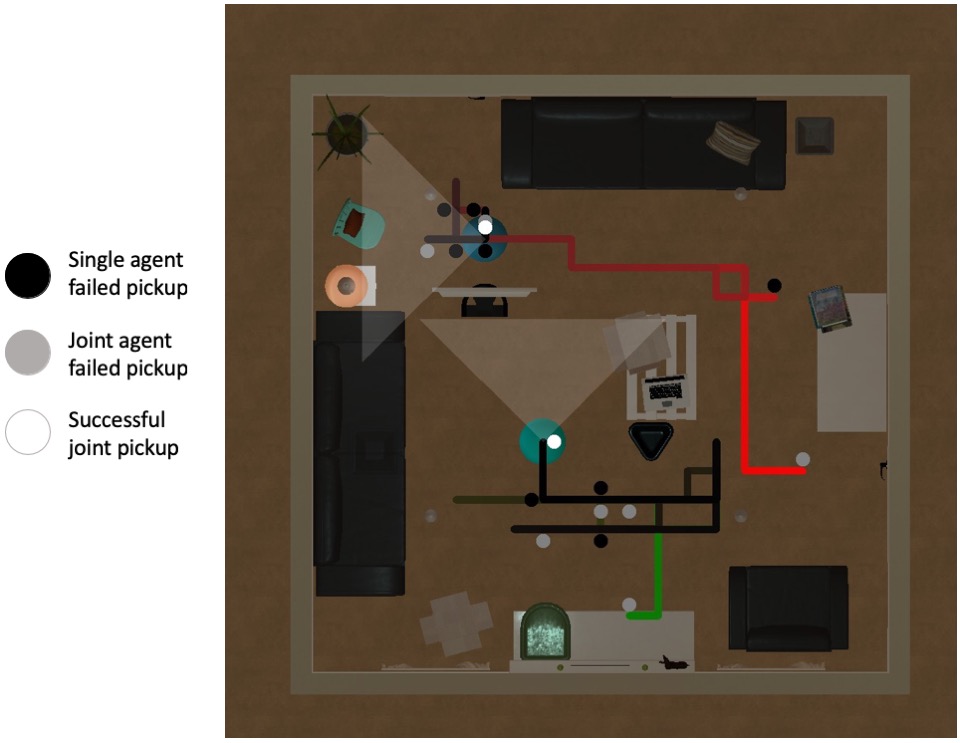} &
    \includegraphics[height=4.5cm]{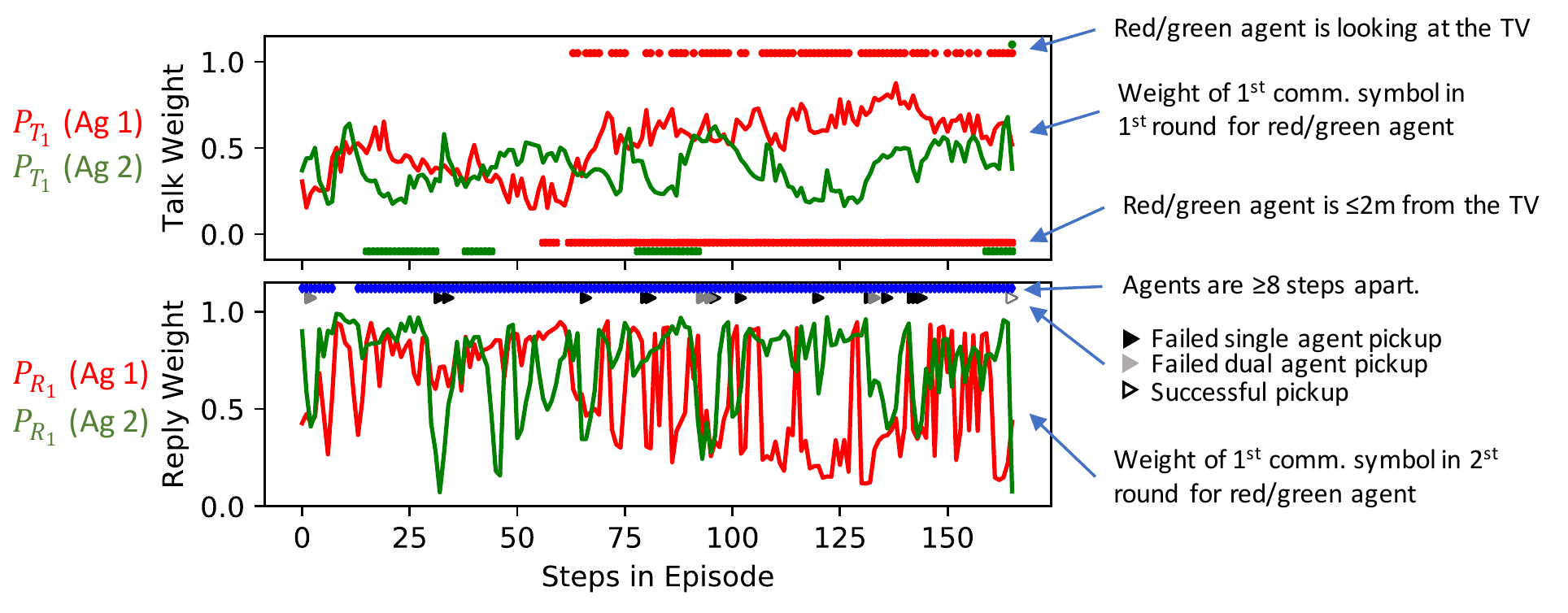}\\
     (a) \emph{Constrained} setting agent trajectories &
    (b) Communication between agents \\
    \end{tabular}
    \vspace*{-2mm}
    \caption{Single episode trajectory with associated agent communication.}
    \label{fig:communication}
    \vspace*{-1mm}
\end{figure*}

Fig.~\ref{fig:con_uncon} presents the accuracy and relative episode lengths metrics for the unseen scenes and \emph{Unconstrained} task in contrast to the \emph{Constrained} task.
In these plots, for brevity we only consider the extreme cases of having full communication \vs no communication. As expected, the \emph{Unconstrained} setting is easier for the agents with higher accuracy and lower episode lengths.
Communication is also advantageous in the \emph{Unconstrained} setting, but its benefits are lesser compared to the \emph{Constrained} setting.

Table~\ref{tab:depth_gridworld} shows a large jump in accuracy when we provide a perfect depth map as an additional input on the \emph{Constrained} task, indicating that improved perception is beneficial to task completion. We also obtained significant jumps in accuracy (from 31.8 $\pm$ 3.8 to 37.2 $\pm$ 4.0) when we increase the size of our vocabulary from 2 to 8. This analysis was performed in the explicit-only communication and \emph{Constrained} environment setup. However, note that even with a vocabulary of 2, agents may be using the full continuous spectrum to encode more nuanced events.

\noindent \textbf{Grid-world abstraction.} In order to assess impact of learning to communicate from pixels rather than, as in most prior work, from grid-world environments, we perform a direct translation of our task into a grid-world and compare its performance to our best model. We transform the 1.25m $\times$ 2.75m area in front of our agent into a $5\times 11$ grid where each square is assigned a 16 dimensional embedding based on whether it is free space, occupied by another agent, occupied by the target object, otherwise unreachable, or unknown (in the case the grid square leaves the environment). The agents then move in AI2-THOR but perceive this partially observable grid-world. Agents in this setting acquire a large bump in accuracy on the \emph{Constrained} task (Table~\ref{tab:depth_gridworld}), confirming our claim that photo-realistic visual environments are more challenging than grid-world like settings. 

\noindent \textbf{Interpreting Communication.}\label{subsec:qualitative}
\begin{table}
\centering
\begin{normalsize}
\hspace{-2.5mm}
\begin{tabular}{|c|ccc|ccc|}\hline
  &$\beta^{\leq}$ & $\beta_{t}^{\leq}$ & $\beta_{r}^{\leq}$
  &$\beta^{\text{see}}$ &$\beta_{t}^{\text{see}}$ &$\beta_{r}^{\text{see}}$ \\ \hline
  Est.
  & 0.35&1.23&-0.35& 
    0.88&0.59&-1.1 \\
  SE & 0.013&0.019&0.013&
       0.013&0.015&0.013 \\ \hline
\end{tabular}
\end{normalsize}\\
\begin{normalsize}
\begin{tabular}{|c|cccccc|}\hline
&$\beta^{\text{pick}}$ & $\beta_{t,0}^{\text{pick}}$ &$\beta_{r,0}^{\text{pick}}$ & $\beta_{t,1}^{\text{pick}}$
  &$\beta_{r,1}^{\text{pick}}$ & $\beta^{\text{pick}}_{\vee,r}$ \\ \hline
Est &1.06 &-0.01&-0.04&0&-0.03&-1.09 \\
SE & 0.012&0.007&0.006&0.007&0.006&0.021 \\\hline
\end{tabular}
\end{normalsize}
\vspace*{-2mm}
\caption{Estimates, and corresponding robust bootstrap standard errors, of the parameters from
  Section \ref{subsec:qualitative}.}\label{table:estimates}
  \vspace*{-4mm}
\end{table}
While we have seen, in Section~\ref{subsec:quant}, that communication can substantially benefit our task, we now investigate \emph{what} these agents have learned to communicate. We focus on the communication strategies learned by agents with a vocabulary of two in the \emph{Constrained} setting. \figref{fig:communication} displays one episode trajectory of the
two agents with the corresponding communication. From
\figref{fig:communication}(b) we generate hypotheses regarding
communication strategies.  Suppressing the dependence on episode and step,
for $i\in\{0,1\}$ let $t_i$ be the weight assigned by agent $i$ to the  $1^\text{st}$
element
  of the vocabulary in the $1^\text{st}$ round of communication,
and similarly let $r_i$ be as $t_i$ but for the $2^\text{nd}$ round of
communication. When the agent with the red trajectory (henceforth
called agent 0 or $A_0$) begins to see the TV the weight $t_0$
increases and remains high until the end of the episode. This
suggests that the $1^\text{st}$ round of communication 
may
be used to signify closeness to or visibility of the TV. On the other
hand, the pickup actions taken by the two agents are associated with
the agents making $r_0$ and $r_1$ simultaneously small.

To add evidence to these hypotheses we fit logistic regression models
to predict, from (functions of) $t_i$ and $r_i$, two oracle values
(\eg, whether the TV is visible) and whether or not the agents will
attempt a pickup action. As the agents are largely symmetric we take
the perspective of $A_0$ and define the models 
$\sigma^{-1}\ P(\substack{\text{$A_0$ is $\leq 2$m from the TV}}) =
  \beta^{\leq} + \beta_{t}^{\leq} t_0 + \beta_{r}^{\leq} r_0$,\  
 $\sigma^{-1}\ P(\substack{\text{$A_0$ sees TV and is $\leq 1.5$m from
  it}})\ =
  \beta^{\text{see}} + \beta_{t}^{\text{see}} t_0 +
  \beta_{r}^{\text{see}} r_0$, and
 $\sigma^{-1}\ P(\substack{\text{$A_0$ attempts a pickup action}})=                
  \beta^{\text{pick}} + \sum_{i\in\{0,1\}}(\beta_{t,i}^{\text{pick}}
  t_i + \beta_{r,i}^{\text{pick}} r_i) +
  \beta^{\text{pick}}_{\vee,r}\max(r_0, r_1)$
where $\sigma^{-1}$ is the logit function. Details of how these models
are fit can be found in the appendix.

From Table~\ref{table:estimates}, which displays the estimates of the
above parameters along with their standard errors, we find strong
evidence for the above intuitions. Note, for all of the estimates
discussed above, the standard errors are very small, suggesting highly
statistically significant results. The large positive coefficients
 associated with $\beta_t^{\leq}$ and
$\beta^{\text{see}}_{t}$ suggest that, conditional on  $r_0$ being
held constant, an increase in the weight $t_0$ is associated with a
higher probability of $A_0$ being near, and seeing, the TV. Note also
that the estimated value of $\beta^\text{see}_r$ is fairly large in
magnitude and negative. This is very much in line with our prior
hypothesis that $r_0$ is made small when agent 0 wishes to signal a
readiness to pickup the object. Finally, essentially all estimates of
coefficients in the final model are close to 0 except for
$\beta^{\text{pick}}_{\vee,r}$ which is large and negative. Hence,
conditional on other values being fixed, $\max(r_0,r_1)$ being small
is associated with a higher probability of a subsequent pickup
action. Of course $r_0,r_1 \leq \max(r_0,r_1)$ again lending evidence
to the hypothesis that the agents coordinate pickup actions by setting
$r_0,r_1$ to small values.

%% file: conc.tex
\section{Conclusion}
\vspace{-0.2cm}

\noindent We study the problem of learning to collaborate in visual environments and demonstrate the benefits of learned explicit and implicit communication to aid task completion. We compare performance of collaborative tasks in photo-realistic visual environments to an analogous grid-world environment, to establish that the former are more challenging. We also provide a statistical interpretation of the communication strategy learned by the agents. 

Future research directions include extensions to more than two agents, more intricate real-world tasks and scaling to more environments. It would be exciting to enable natural language communication between the agents which also naturally extends to involving human-in-the-loop.

%% file: appendix.tex
\appendix
\section{Appendix}
\begin{figure*}
\centering
\includegraphics[height=5.5cm]{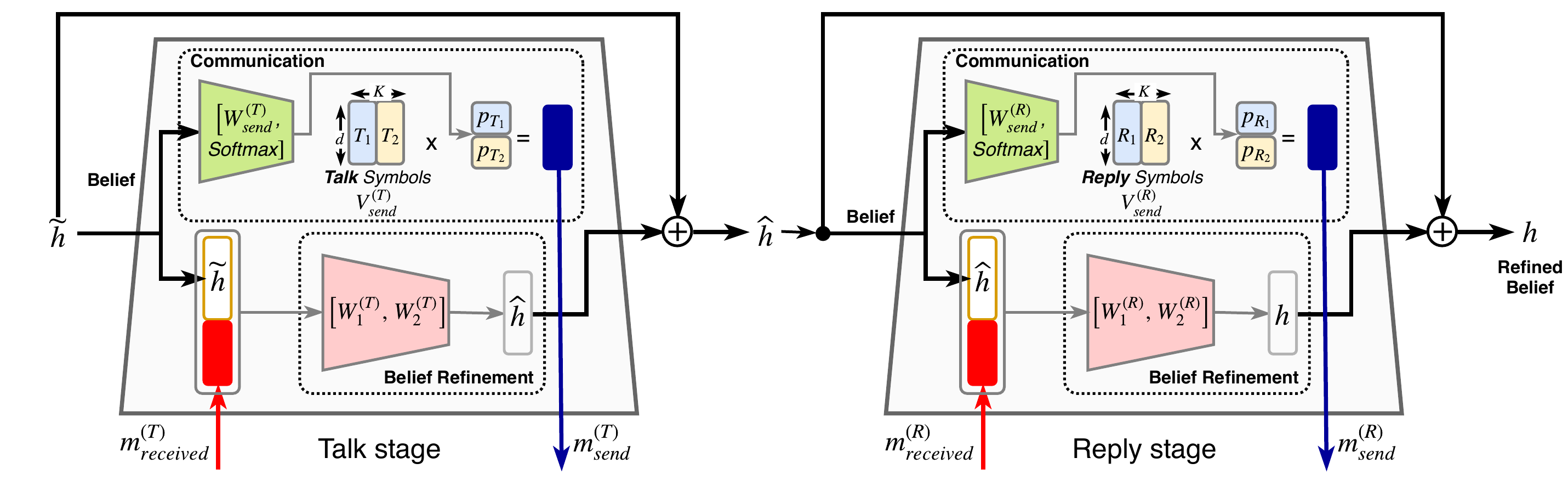}
\caption{Two stages of communication and belief refinement module - \textit{talk} and \textit{reply}. The refined belief from the talk stage is further refined by another round of communication between agents at the reply stage. In this illustration the size of vocabulary is 2 \ie $K=2$.}
\label{fig:talk_and_reply}
\end{figure*}
This appendix presents the following content:
\begin{enumerate}\compresslist
    \item Visualizations of the grid-world abstraction of our task,
    \item Our learning algorithm,
    \item Interplay between \textit{talk} and \textit{reply} stages of the communication and belief refinement module,
    \item Implementation details of  model
    \item A detailed explanation of metrics used in our paper,
    \item Quantitative evaluation of our models but now evaluated on seen scenes,
    \item Statistical analysis of agent communication strategies but now demonstrated on unseen scenes,
    \item Qualitative results of agents with different communication abilities deployed on unseen scenes. This includes clip summaries with agent communication signals for video \url{https://youtu.be/9sQhD_Gin5M}.
\end{enumerate}

\subsection{AI2-THOR to Grid-world}

\begin{figure*}[h!]
    \centering
    \begin{tabular}{p{5cm}p{5cm}}
     \includegraphics[height=5cm,frame]{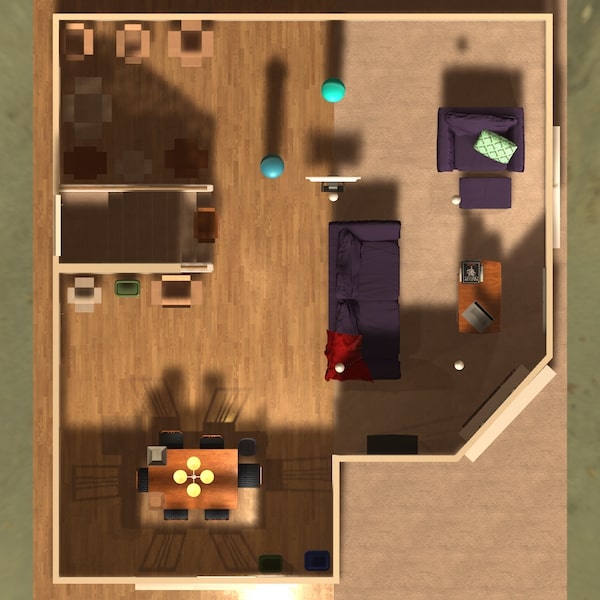}&
    \includegraphics[height=5cm,frame]{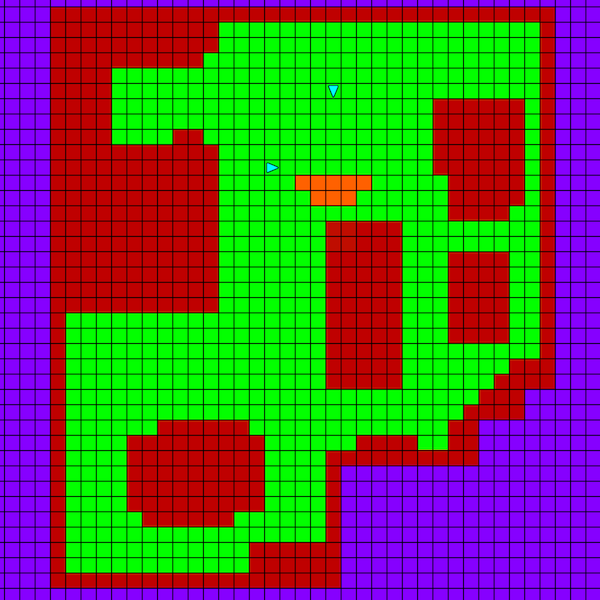} \\
(a) Top view of AI2-THOR scene & (b) Corresponding grid-world \\  
    \end{tabular}
    \caption{An AI2-THOR scene from a top-down view along the corresponding grid-world. Note that each agent (teal triangles) only observes a small portion of the grid-world at any given time-step, see Figure~\ref{fig:agent-view-to-gridworld} for details. Here each color corresponds to a different category: freespace (green), impassable terrain (red), target object (orange), and unknown (purple).}
    \label{fig:gridworld}
\end{figure*}

\begin{figure*}[h!]
    \centering
    \begin{tabular}{p{3.4cm}p{3.4cm}p{3.4cm}}
     \includegraphics[height=3.4cm,frame]{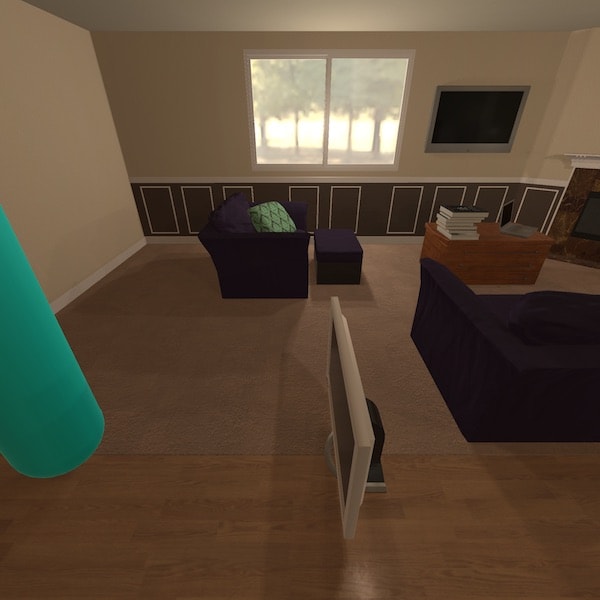}&
    \includegraphics[height=3.4cm,frame]{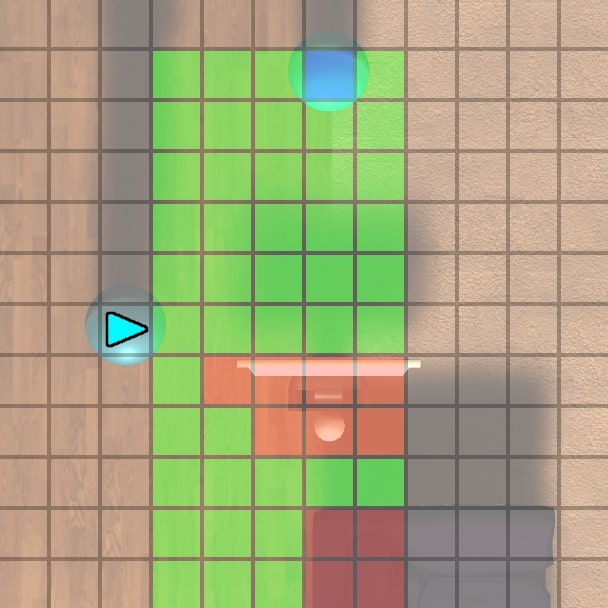}&
\includegraphics[height=3.4cm,frame]{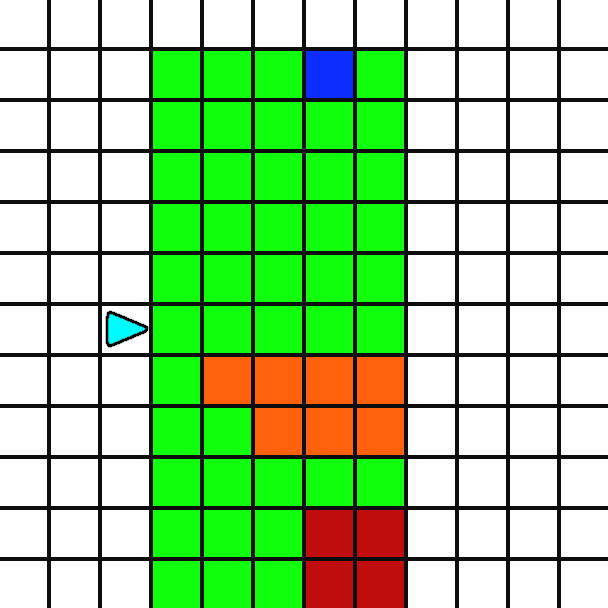} \\[3mm]
\includegraphics[height=3.4cm,frame]{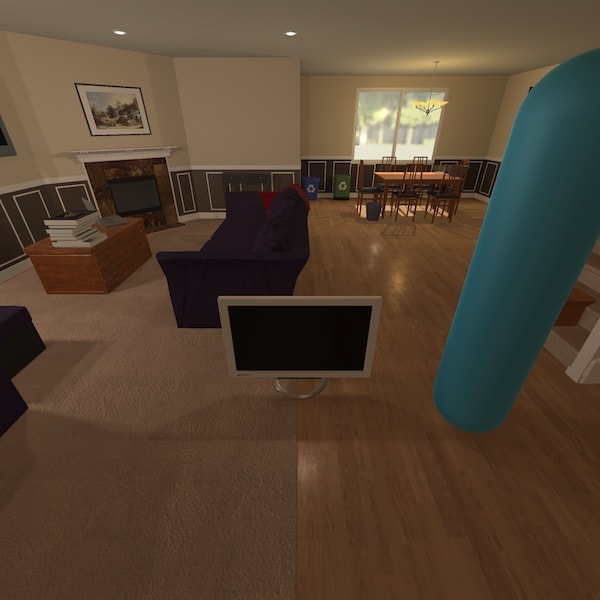}&
    \includegraphics[height=3.4cm,frame]{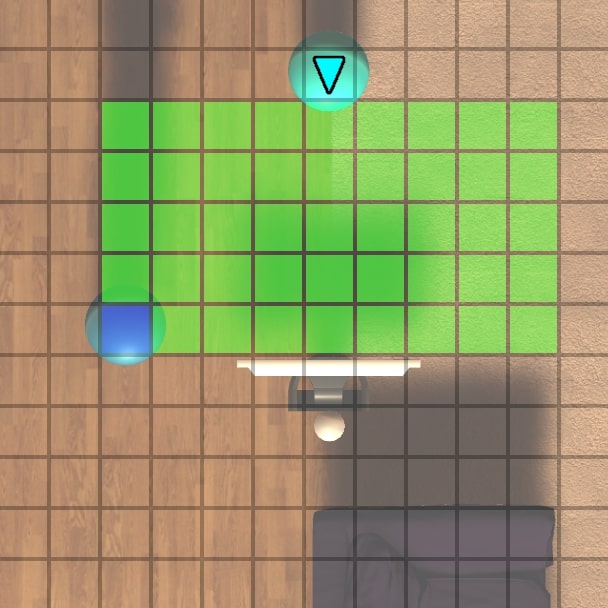}&
\includegraphics[height=3.4cm,frame]{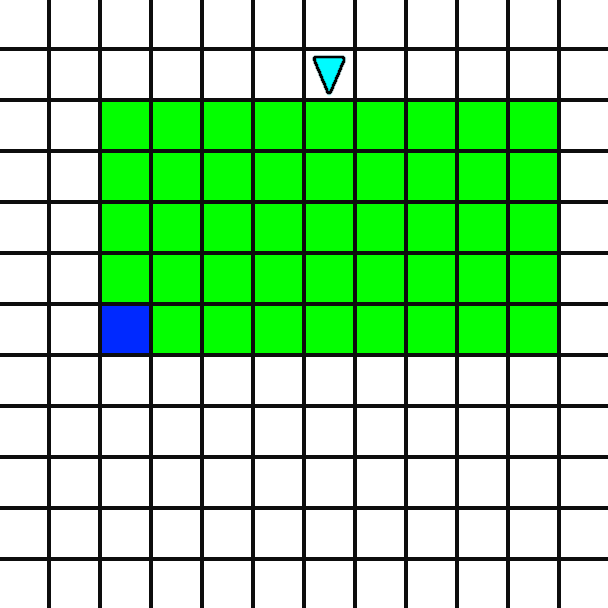} \\
(a) First-person AI2-THOR agent view & (b) Agent partially observed grid-world overlayed on map view & (c) Grid-world corresponding to agent view  \\  
    \end{tabular}
    \caption{First person viewpoints of agents in AI2-THOR and the corresponding grid-world observations. Note that white squares are unobserved and blue squares correspond to another agent, see Figure~\ref{fig:gridworld} for a description of the other colors.}
    \label{fig:agent-view-to-gridworld}
\end{figure*}

In order to assess the impact of learning to communicate directly from pixels rather than, as in most prior work, from grid-world environments, we perform a direct translation of our task into a grid-world and compare its performance to our best model. For this purpose we transform AI2-THOR into a grid-world environment. Figure~\ref{fig:gridworld} visualizes, for a single AI2-THOR scene, this transformation. To make our comparison fair, as our pixel-based agents only obtain partial information about their environment at any given timestep, we impose the same restriction on our grid-world agents by only providing them with an egocentric $5\times11$ view of their environment (see Figure~\ref{fig:agent-view-to-gridworld}).

\begin{algorithm}[t]
\hspace*{-0.5cm}
\begin{minipage}{0.97\linewidth}
\algblockdefx[NAME]{WhileInParallel}{EndWhileInParallel}%
      [1]{\textbf{while} #1 \textbf{in parallel do}}%
          {\textbf{end}}
\begin{algorithmic}[1]

  \State Randomly initialize shared model weights $\theta_{\text{shared}}$
  \State Set global episode counter $c \gets 0$
  \WhileInParallel{$c <$ maxEpisodes}
  \State $\theta \gets \theta_{\text{shared}}$
  \State $c \gets c + 1$
  \State Randomly choose environment
  \State Randomize agents' positions and TV location \label{l1}
  \State Set $\alpha \gets 0.9 \cdot \text{max}(1 - c / 10000, 0)$
  \State Set $\pi \gets (1-\alpha)\cdot \pi_{\theta} + \alpha\cdot\pi^*$
  \State Roll out trajectory of length $\leq 500$ from both agents using $\pi$.
  \State $L_{\text{a3c}} \gets$ A3C loss for trajectory
  \State $L_{\text{cross}} \gets$ cross entropy loss of trajectory w.r.t.\ $\pi^*$
  \If{no expert actions sampled in trajectory}
  \State $g \gets \nabla_{\theta} (L_{\text{a3c}} +L_{\text{cross}})$ 
  \Else
  \State $g \gets \nabla_{\theta} L_{\text{cross}}$
  \EndIf
  \State Perform one gradient update of $\theta_{\text{shared}}$ using ADAM with gradients $g$ and statistics shared across processes
  \EndWhileInParallel
\end{algorithmic}
\end{minipage}
  \caption{Learning Algorithm 
  }
  \label{alg:learn}
\end{algorithm}

\subsection{Learning algorithm}
Algorithm \ref{alg:learn} succinctly summarizes our learning procedure as otherwise described in Section 3.3 of the main paper.

\subsection{Talk and Reply stages}
Explicit communication happens via two stages - talk and reply. As illustrated in \figref{fig:talk_and_reply}, each stage has it's own weights ($V_{send}$, $W_{send}$, $W_1$, $W_2$). These are clearly marked using superscripts of $^{(T)}$ and $^{(R)}$ for the talk and reply stage, respectively.

\subsection{Implementation Details.}
We use the same hyperparameters and embedding dimensionality in  all of our experiments.
In our A3C loss we discount rewards with a factor of $\gamma=0.99$ and weight the entropy maximization term with a factor of $\beta=0.01$. We use the Adam optimizer with a learning rate of $10^{-4}$, momentum values of 0.9 and 0.999 (for the first and second moments respectively), and share optimizer statistics across processes. Gradient steps are made in the hogwild approach, that is without explicit synchronization or locks between processes~\cite{RechtNIPS2011}.

Each model was trained for 100,000 episodes. Each episode is initialized in a random train (seen) scene of AI2-THOR. Rewards provided to the agents are: 1 to both agents for a successful pickup action, constant -0.01 step penalty to discourage long trajectories,
-0.02 for any failed action (\eg, running into a wall) and -0.1 for a failed pickup action. Episodes run for a maximum of 500 total steps (250 steps for each agent) after which the episode is considered failed.
The minimum aggregate achievable reward in an episode, obtained by successive attempting failed pickup actions by both agents is -65 while the maximum reward is 1.98 achieved by both
agents immediately picking up the object as their first action and only receiving a single step penalty.

\subsection{Metrics}
We now present a more detailed explanation of the metrics we use to evaluate our models.

\begin{enumerate}[(1)]
    \item Per agent reward structure:
    \begin{itemize}
        \item +1 for performing a successful joint pickup,
        \item -0.1 for a failed pickup action,
        \item -0.02 for any other failed action (trying to move into walls, furniture, etc.), and
        \item -0.01 for each step to encourage short trajectories.
    \end{itemize}
    \item Accuracy: the percentage of episodes which led to the  successful pickup action by both agents.
    \item Number of unsuccessful pickups: total number of pickup actions attempted by both agents which didn't lead to the target being picked up. The three preconditions necessary for a successful joint pickup action are as follows.
    \begin{enumerate}[(i)]
        \item Both agents perform the pickup action simultaneously,
        \item \label{item:dist_target}Both agents are closer than 1.5m to the target and the target is visible, and
        \item \label{item:dist_agents}Both agents are  a minimum distance apart from each other (0 for the \emph{Unconstrained} and 8 steps = 2 meters in the manhattan distance for the \emph{Constrained} setting).
    \end{enumerate}
    \item Number of missed pickups: total number of episode steps where both agents
could have picked up the object but did not. This is the number of opportunities where \ref{item:dist_target} and \ref{item:dist_agents} were met, but the agents didn't perform simultaneous pickup actions.
    \item Relative episode length: the quantity 
    \[
    \frac{\text{Episode length following agent policy } (\pi)}{\text{Episode length following oracle policy } (\pi^*)}
    \]
    As it has access to information not available to the agents, our expert policy is also referred to as the oracle policy. As mentioned in the paper, the oracle plans a shortest path from each agent location to the target. This is achieved by leveraging the full  map of the scene (\ie, free space, occupied areas, location of other agent, and the target location).
\end{enumerate}

\subsection{Quantitative evaluation}
In this section we provide quantitative evaluation results of variants of TBONE. We provide results on seen (train) and unseen (test) scenes. Many of the unseen scenes results are already included in the main paper, but we reproduce the full suite of graphs here, for ease of comparison. 

\begin{figure*}[h!]
    \centering
    \begin{tabular}{@{\hskip0pt}c@{\hskip2pt}c@{\hskip2pt}c@{\hskip2pt}c@{\hskip2pt}c}
    (1) Reward&
    (2) Accuracy&
    (3) \begin{tabular}{c}\# Unsuccessful \\ pickups\end{tabular}&
    (4) \begin{tabular}{c}\# Missed \\ pickups\end{tabular}&
    (5) \begin{tabular}{c}Relative episode \\ lengths\end{tabular}\\
    \includegraphics[height=3.4cm]{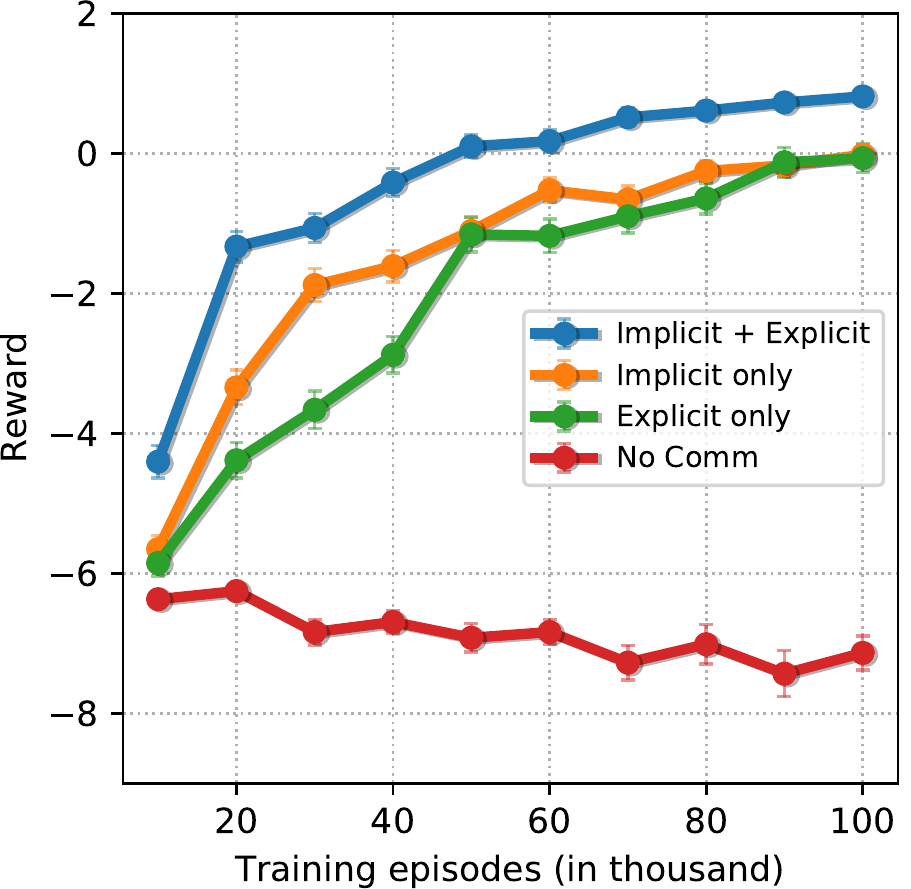}&
    \includegraphics[height=3.4cm]{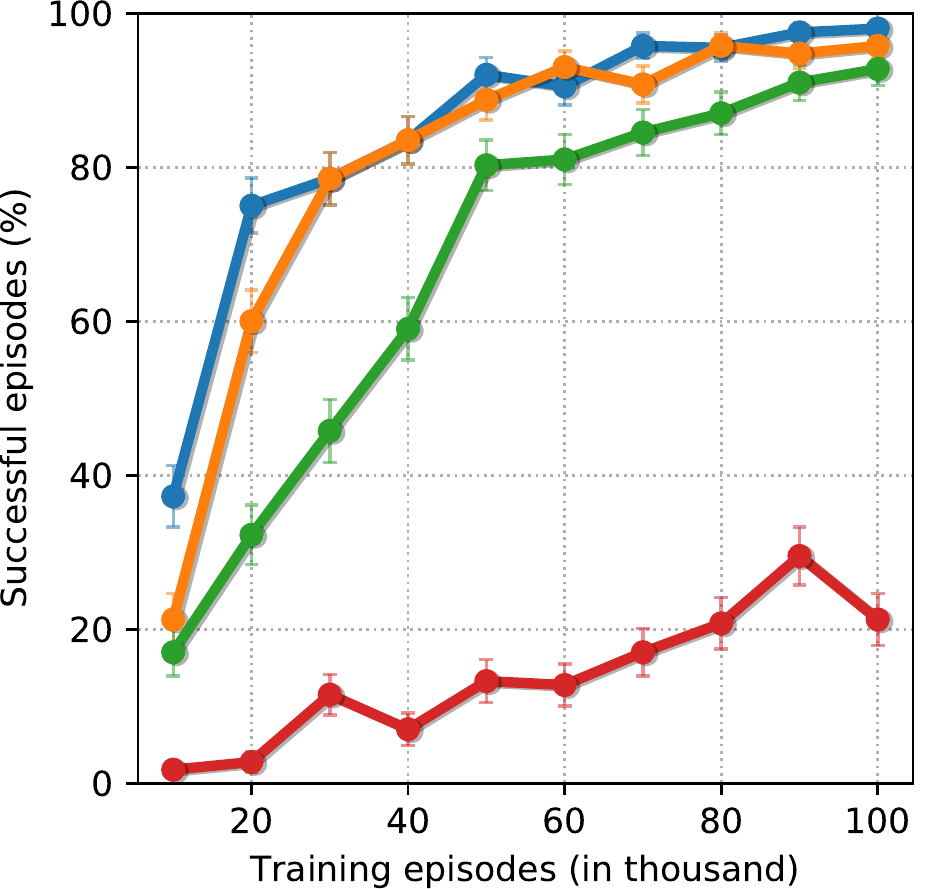}&
    \includegraphics[height=3.4cm]{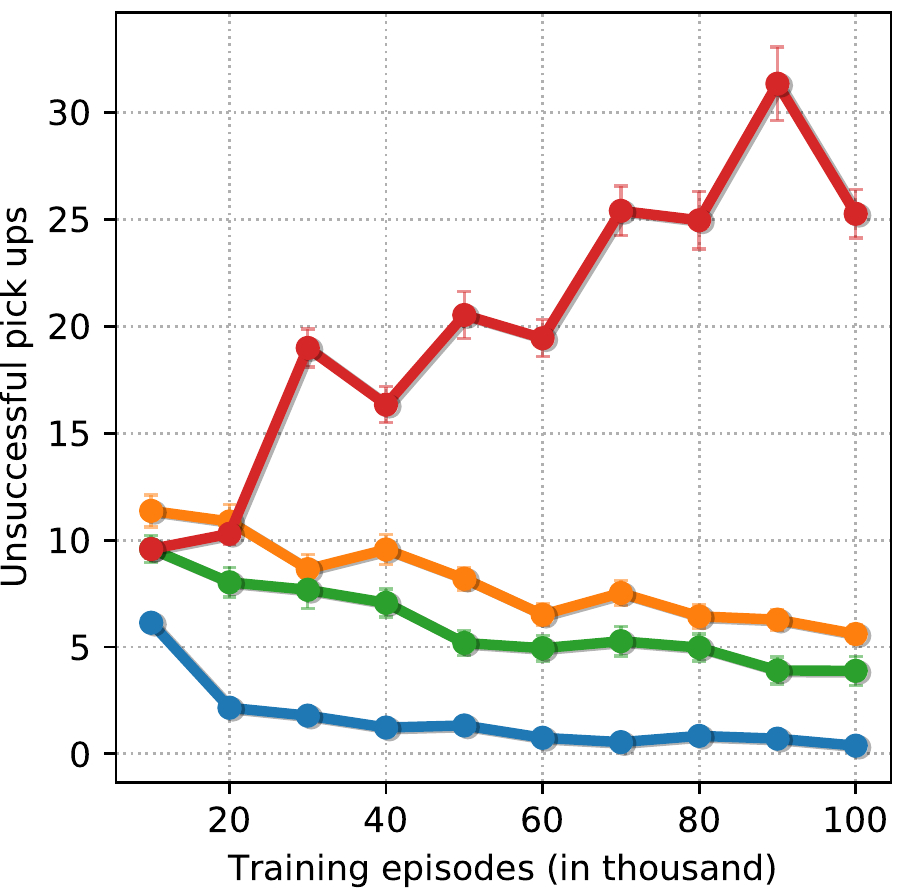}&
    \includegraphics[height=3.4cm]{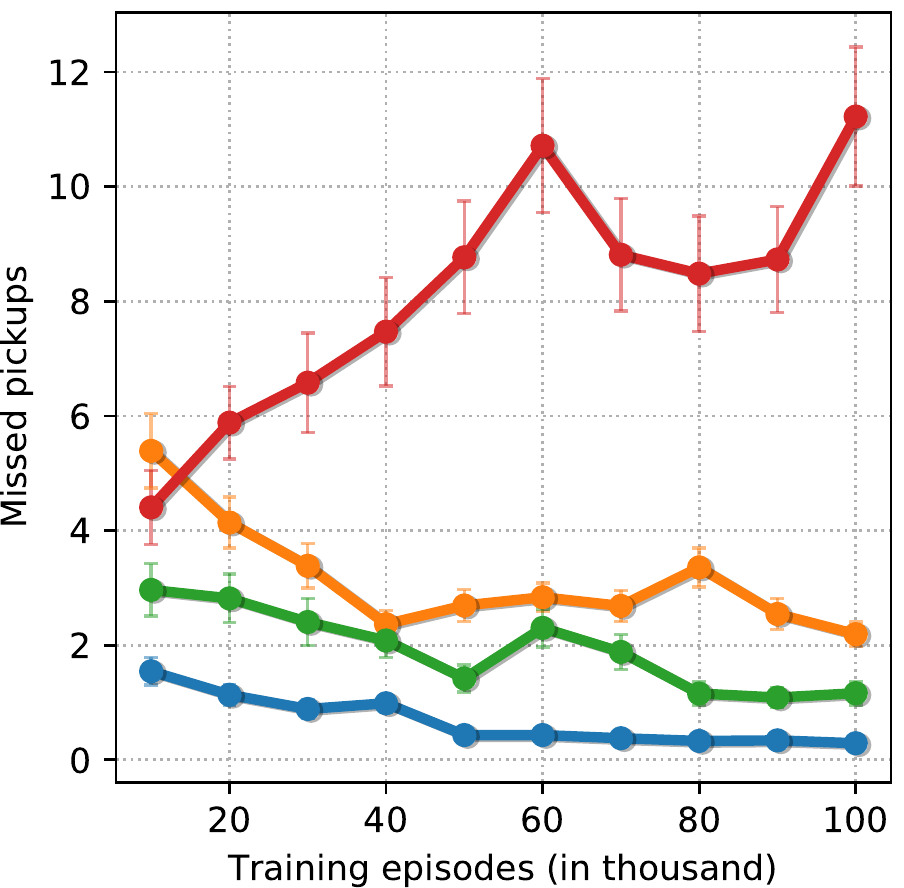}&
    \includegraphics[height=3.4cm]{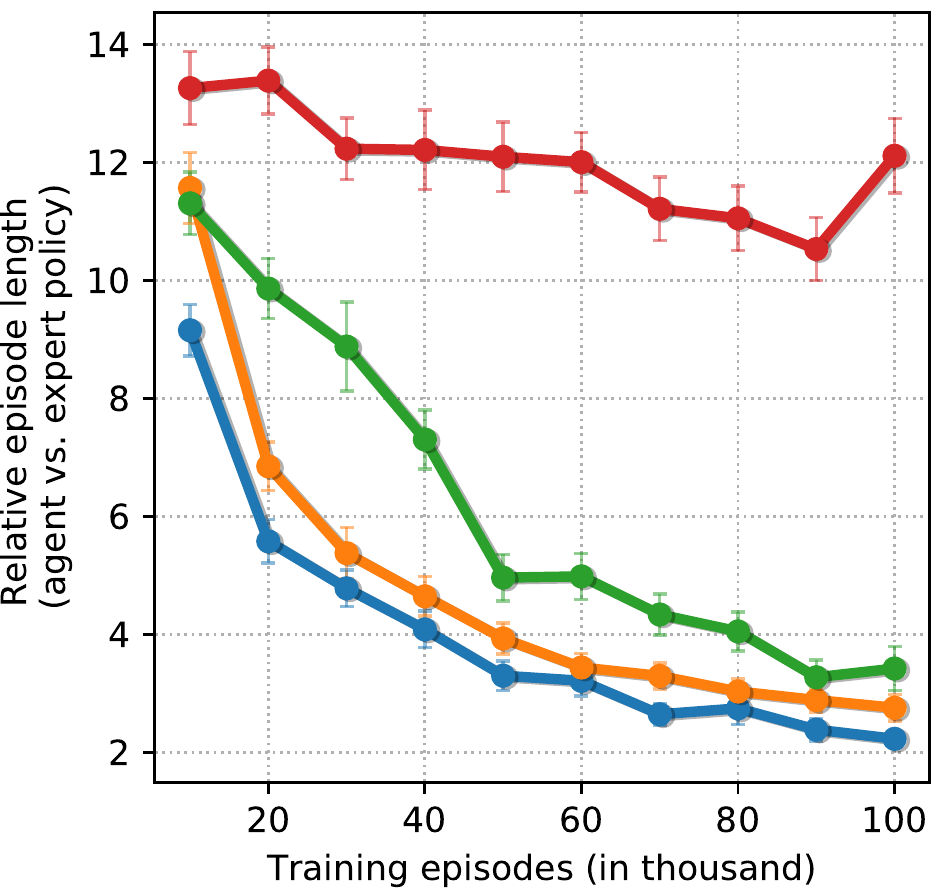}
    \end{tabular}
    \caption{\emph{Constrained} task, seen scenes.}
    \label{fig:constained_seen}
\end{figure*}

\begin{figure*}[h!]
    \centering
    \begin{tabular}{@{\hskip0pt}c@{\hskip2pt}c@{\hskip2pt}c@{\hskip2pt}c@{\hskip2pt}c}
    (1) Reward&
    (2) Accuracy&
    (3) \begin{tabular}{c}\# Unsuccessful \\ pickups\end{tabular}&
    (4) \begin{tabular}{c}\# Missed \\ pickups\end{tabular}&
    (5) \begin{tabular}{c}Relative episode \\ lengths\end{tabular}\\
    \includegraphics[height=3.4cm]{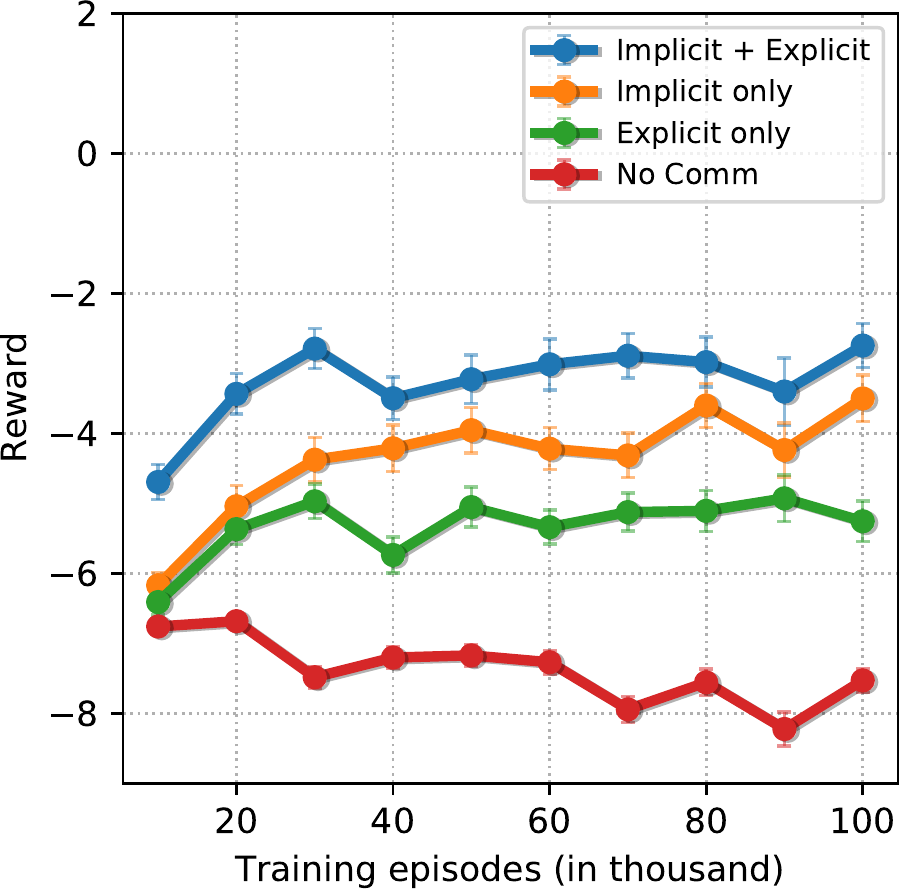}&
    \includegraphics[height=3.4cm]{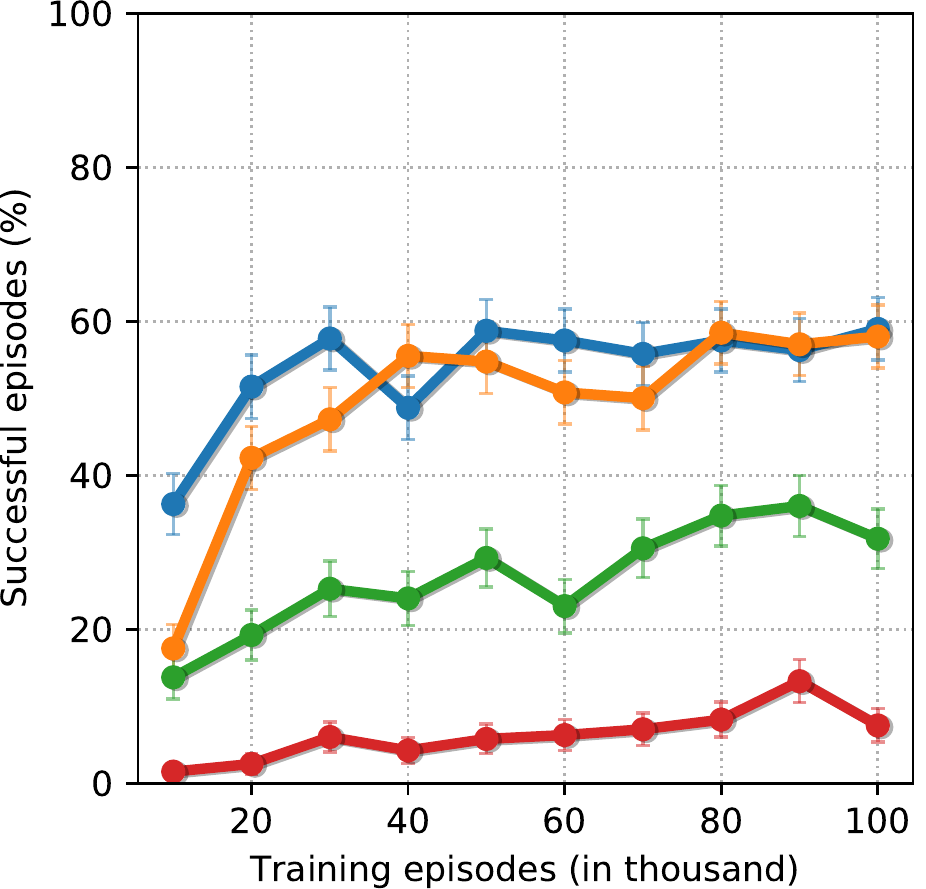}&
    \includegraphics[height=3.4cm]{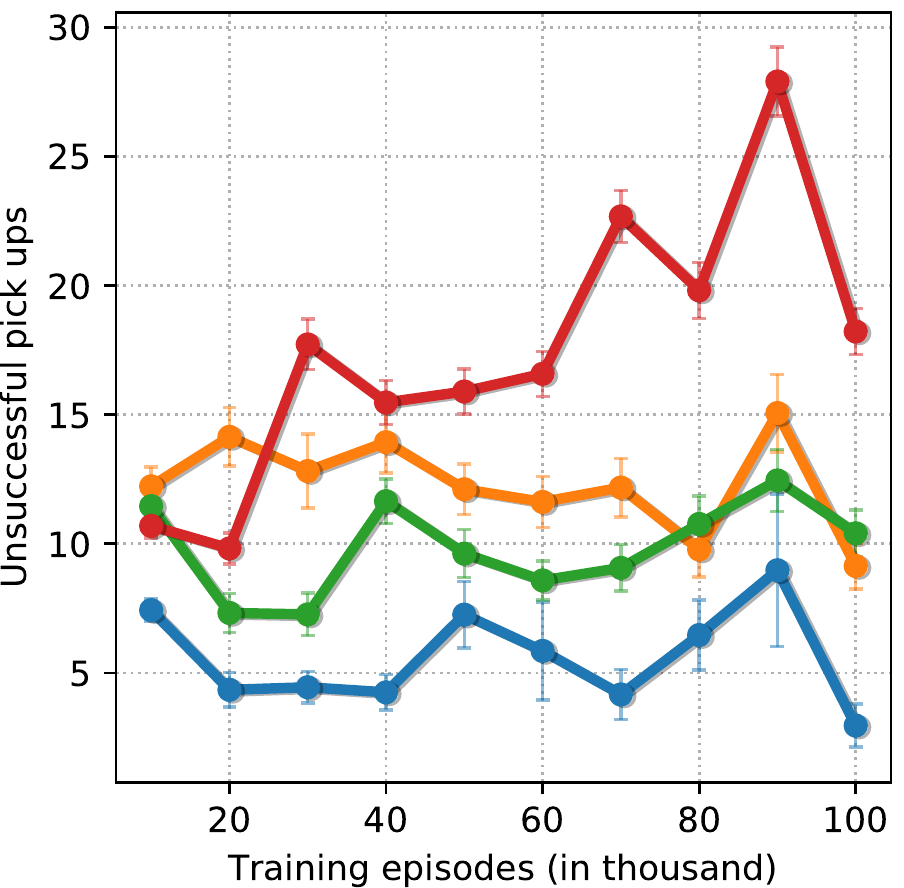}&
    \includegraphics[height=3.4cm]{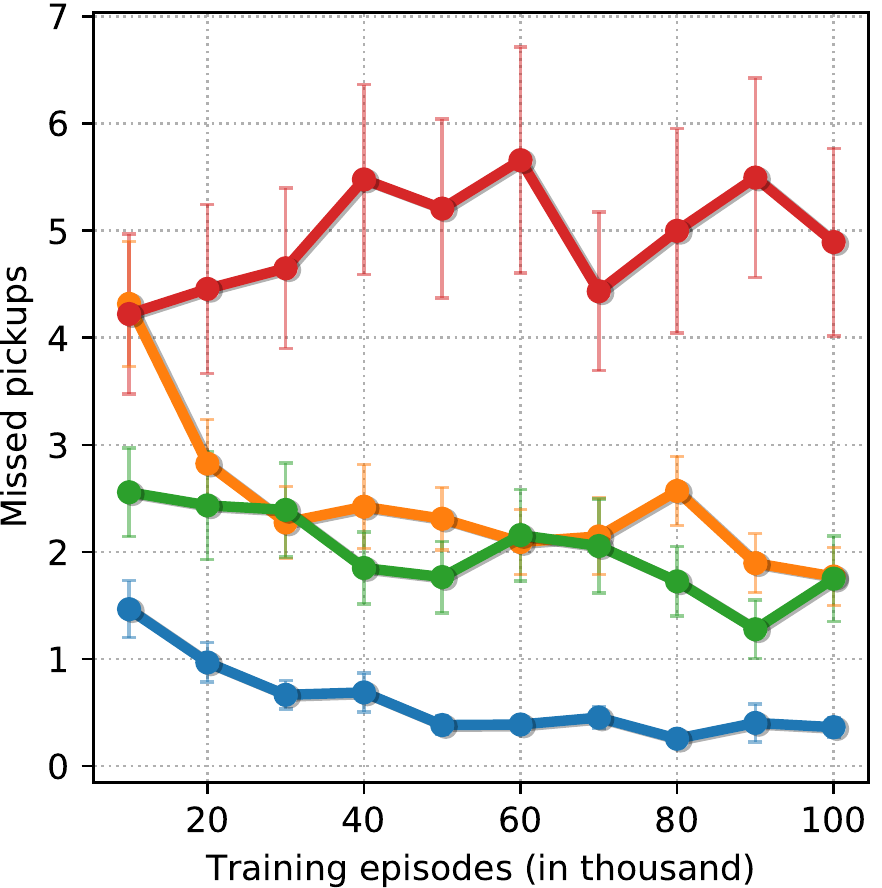}&
    \includegraphics[height=3.4cm]{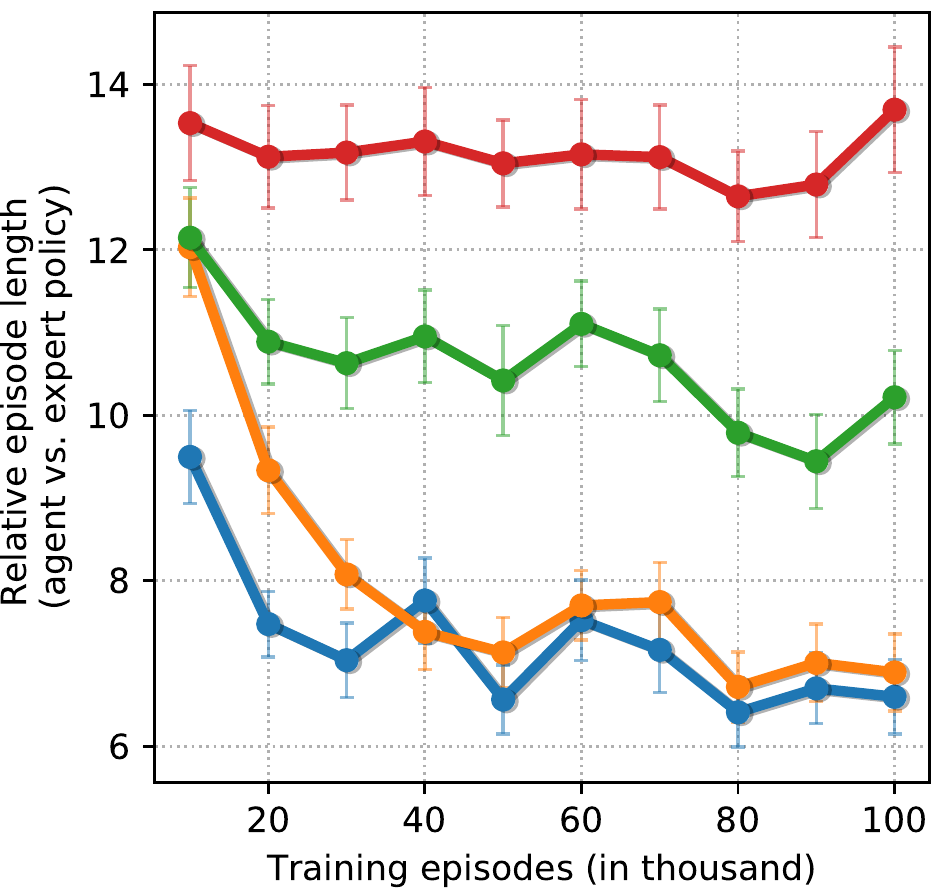}
    \end{tabular}
    \caption{\emph{Constrained} task, unseen scenes.}
    \label{fig:constained_unseen}
\end{figure*}

\begin{figure*}[h!]
    \centering
    \begin{tabular}{@{\hskip0pt}c@{\hskip2pt}c@{\hskip2pt}c@{\hskip2pt}c@{\hskip2pt}c}
    (1) Reward&
    (2) Accuracy&
    (3) \begin{tabular}{c}\# Unsuccessful \\ pickups\end{tabular}&
    (4) \begin{tabular}{c}\# Missed \\ pickups\end{tabular}&
    (5) \begin{tabular}{c}Relative episode \\ lengths\end{tabular}\\
    \includegraphics[height=3.4cm]{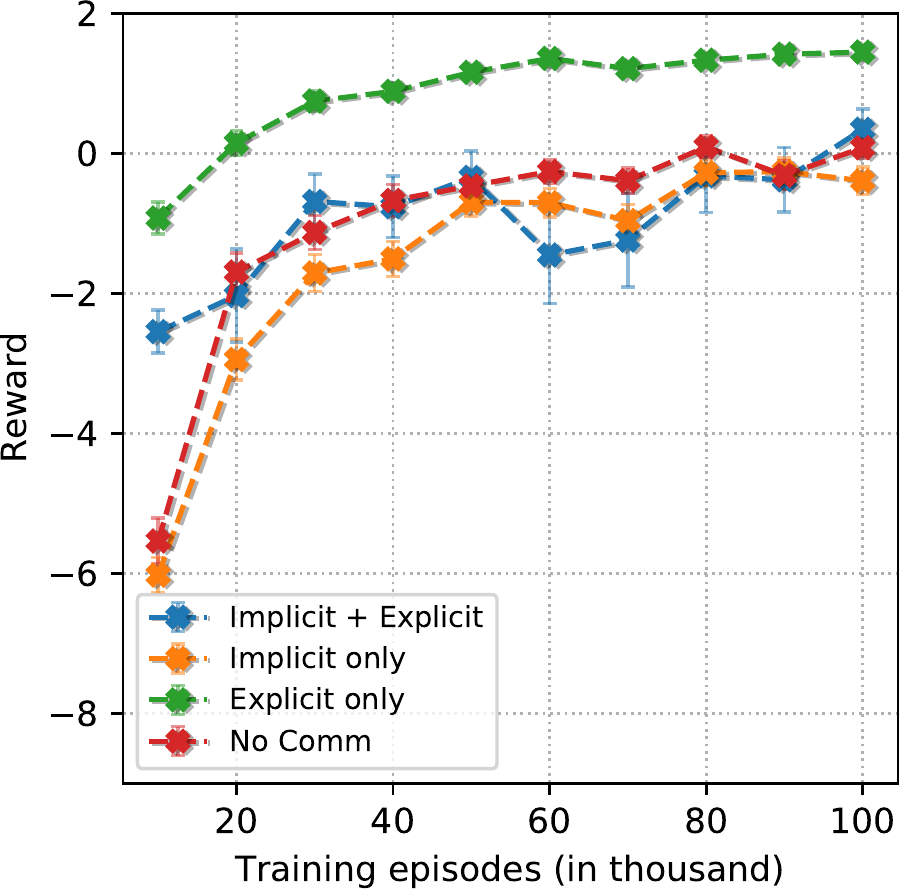}&
    \includegraphics[height=3.4cm]{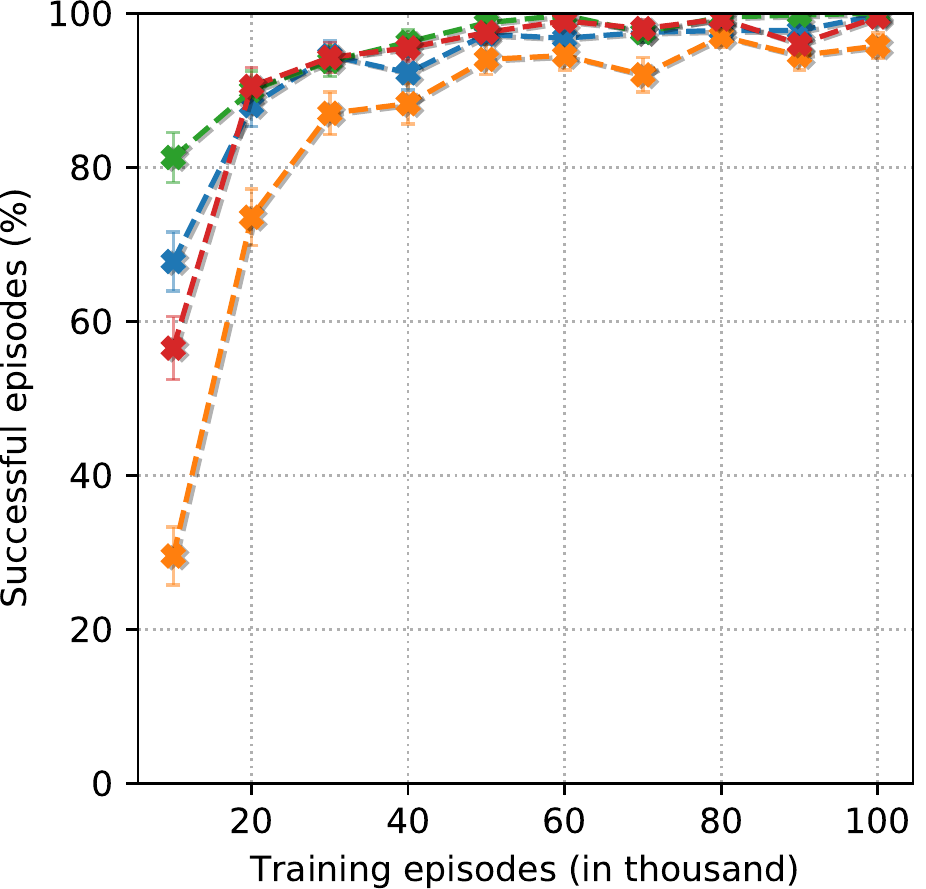}&
    \includegraphics[height=3.4cm]{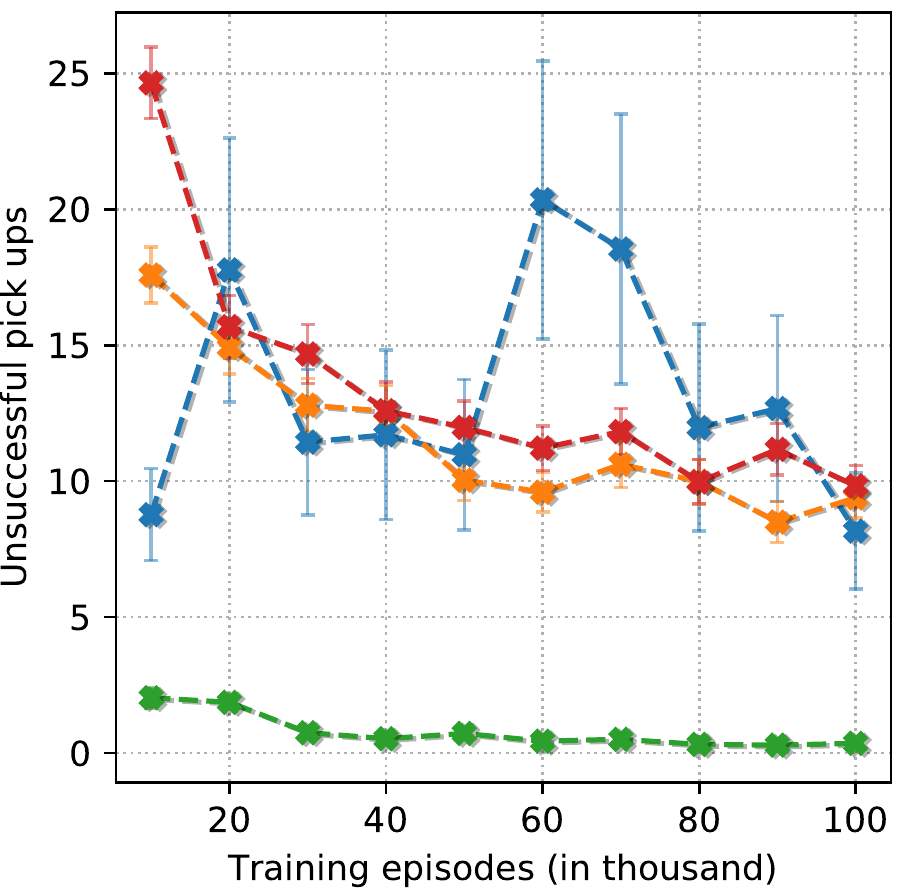}&
    \includegraphics[height=3.4cm]{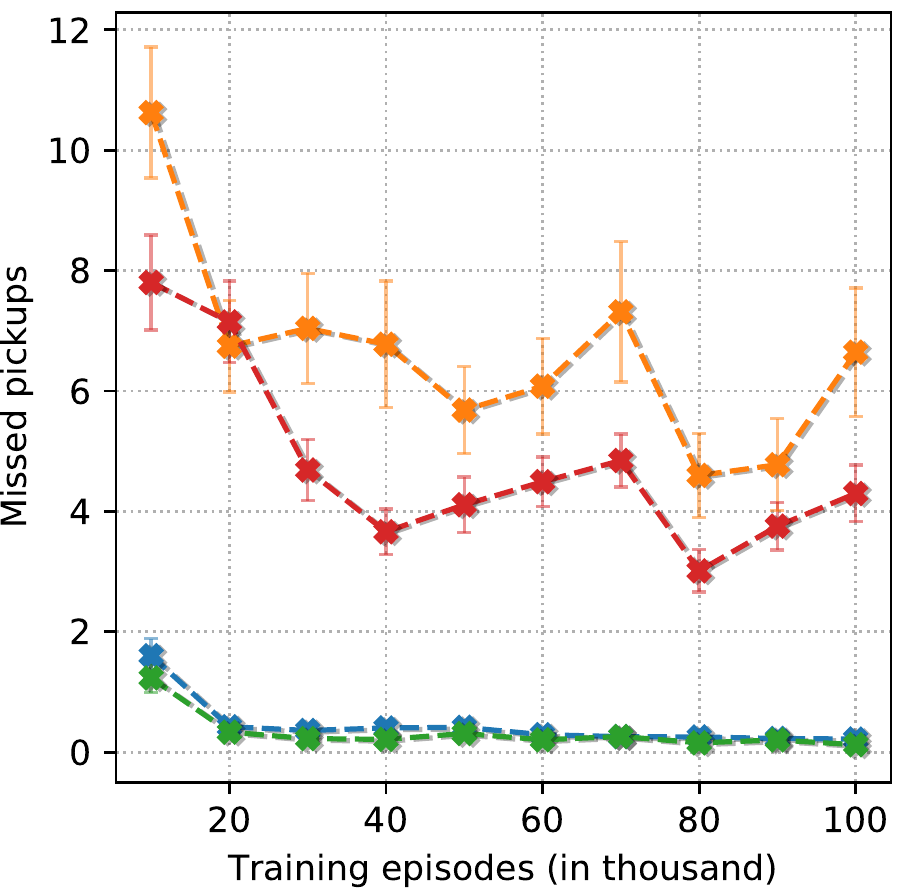}&
    \includegraphics[height=3.4cm]{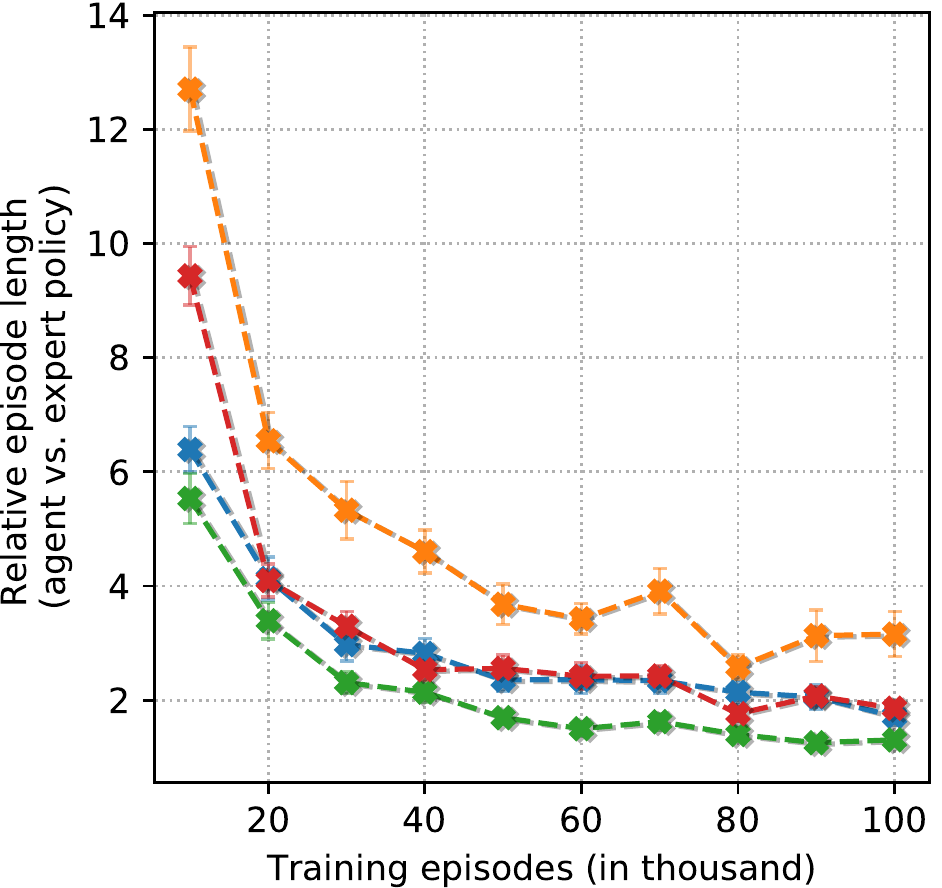}
    \end{tabular}
    \caption{\emph{Unconstrained} task, seen scenes.}
    \label{fig:unconstained_seen}
\end{figure*}

\begin{figure*}[h!]
    \centering
    \begin{tabular}{@{\hskip0pt}c@{\hskip2pt}c@{\hskip2pt}c@{\hskip2pt}c@{\hskip2pt}c}
    (1) Reward&
    (2) Accuracy&
    (3) \begin{tabular}{c}\# Unsuccessful \\ pickups\end{tabular}&
    (4) \begin{tabular}{c}\# Missed \\ pickups\end{tabular}&
    (5) \begin{tabular}{c}Relative episode \\ lengths\end{tabular}\\
    \includegraphics[height=3.4cm]{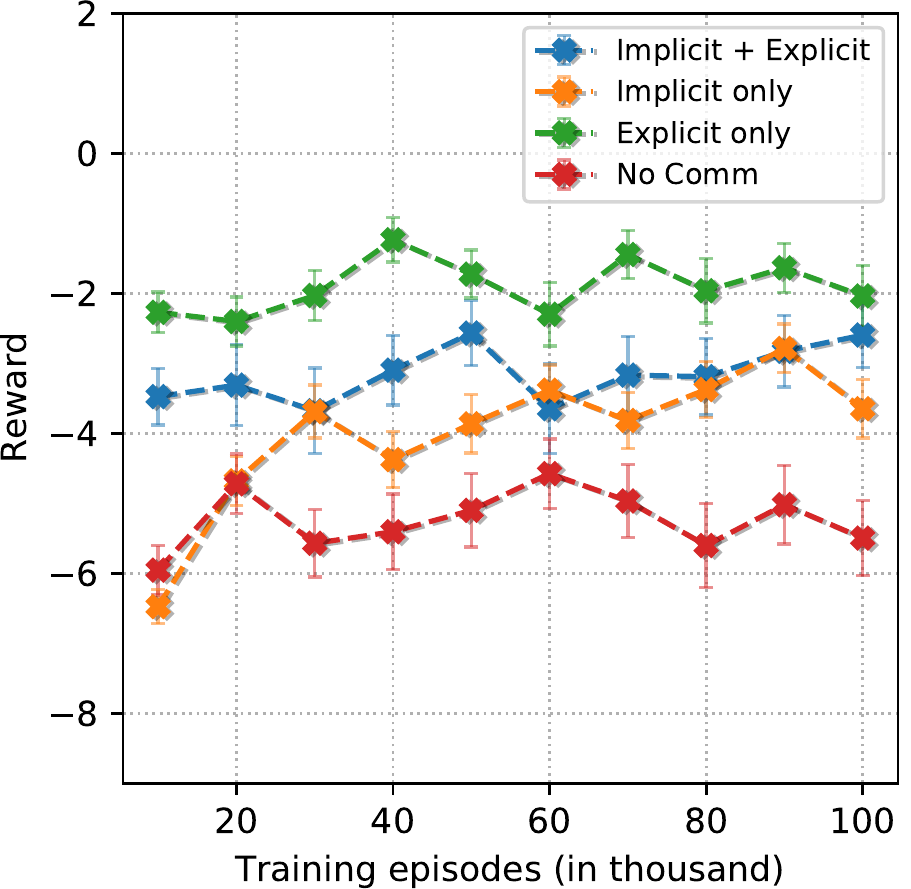}&
    \includegraphics[height=3.4cm]{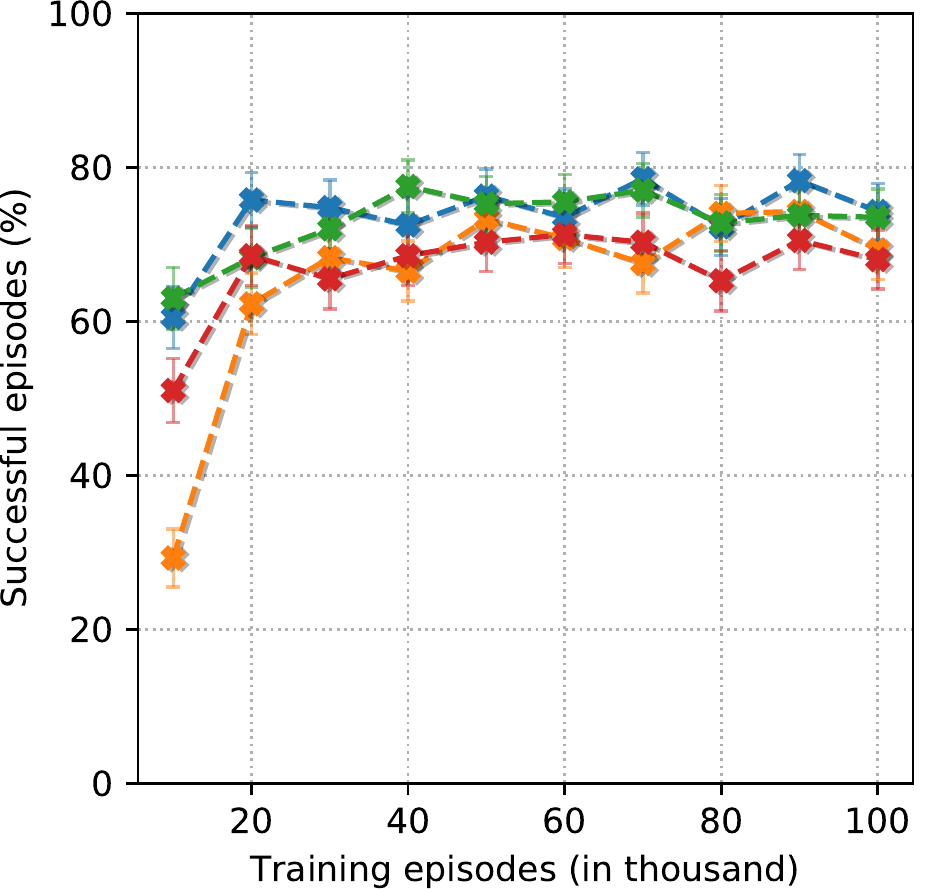}&
    \includegraphics[height=3.4cm]{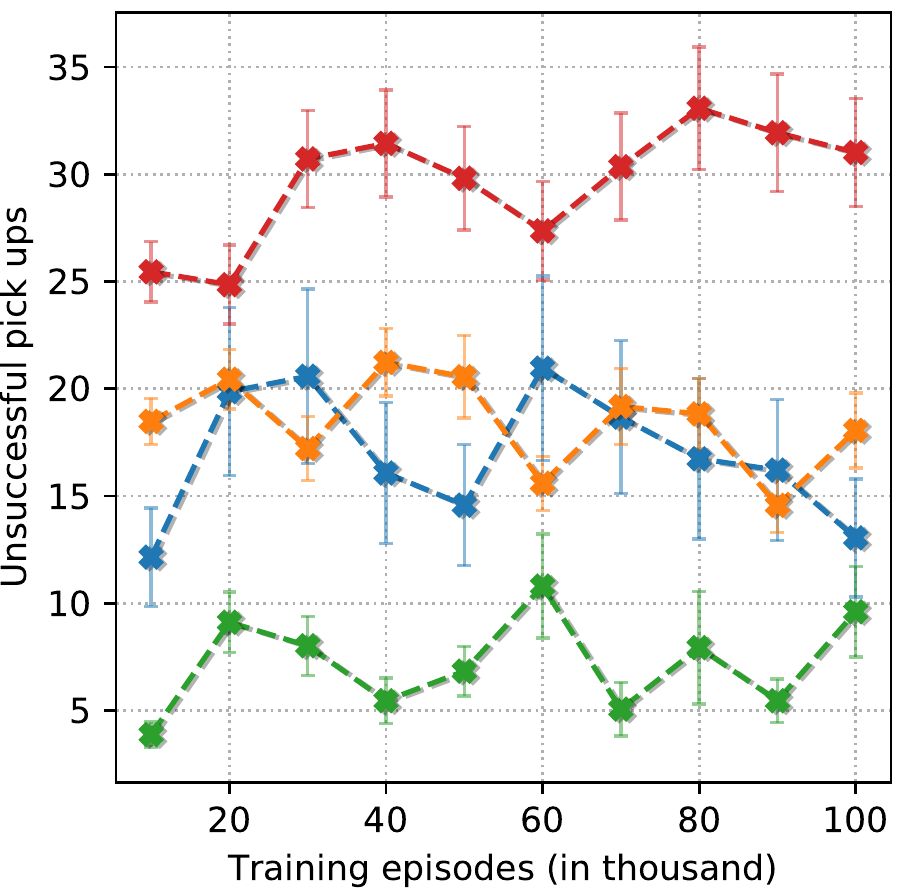}&
    \includegraphics[height=3.4cm]{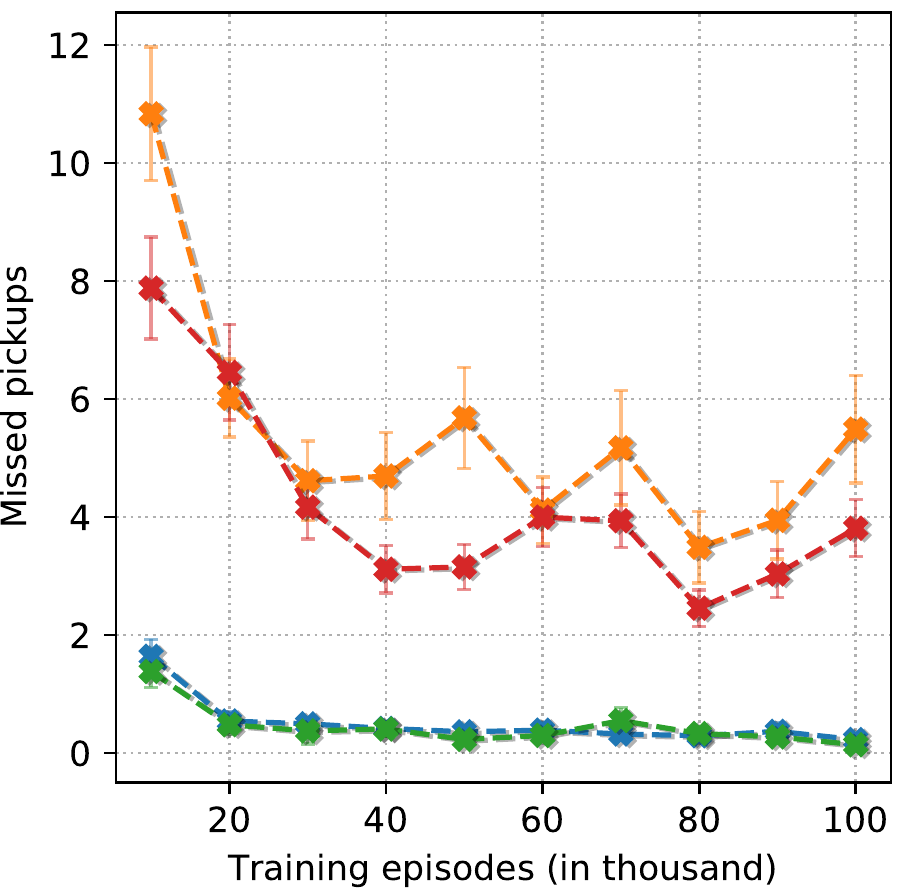}&
    \includegraphics[height=3.4cm]{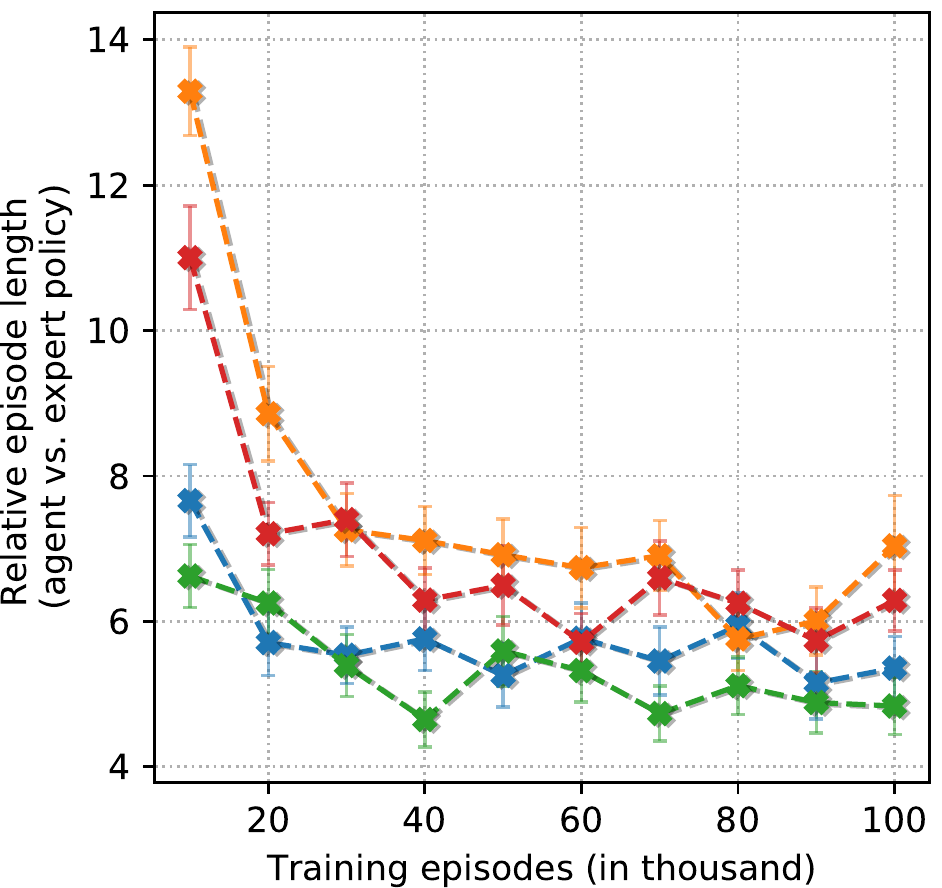}
    \end{tabular}
    \caption{\emph{Unconstrained} task, unseen scenes.}
    \label{fig:unconstained_unseen}
\end{figure*}

For the \emph{Constrained} task, \figref{fig:constained_seen} and \figref{fig:constained_unseen} show the above metrics on seen and unseen scenes, respectively. For the \emph{Unconstrained} task, \figref{fig:unconstained_seen} and \figref{fig:unconstained_unseen} show the above metrics on seen and unseen scenes, respectively.

On the \emph{Constrained} task in seen scenes (\figref{fig:constained_seen}), having both modes of communication clearly produces better rewards. And having either or both modes of communication easily outperforms agents with no means of communication. While the accuracy metric is similar to having only implicit means of communication, the number of unsuccessful pickups, missed pickups, and relative episode lengths metrics show the benefit of having both modes of communication over any one of them. A similar trend is seen in unseen scenes for the same task (\figref{fig:constained_unseen}).

On the \emph{Unconstrained} task, the benefits of communication are, as expected, less dramatic (\figref{fig:unconstained_seen} and \figref{fig:unconstained_unseen}). Since the task is simpler and potentially can be solved without communication, agents with no means of communication are able to obtain high accuracies. But in the absence of communication, agents end up having a large number of unsuccessful pickups. This is expected. With no means of communication, agents simply go close to the TV and start attempting pickups. Only with communication can they lower this metric by coordinating with each other.

\subsection{Interpreting Communication} \label{sec:qualitative}

To fit the logistic models described in Section 4 of the main paper we randomly initialize 2,687 episodes on the 20 training scenes from which we obtain a corresponding number of agent 
trajectories. Treating each step in these trajectories as a single
observation, this results in a dataset containing 143,401 samples. We fit these
logistic models using the \texttt{statsmodels} package~\cite{SeaboldPSC2010} in Python. As observations within a single
episode are highly correlated, we use the bootstrap~\cite{EfronAoS1979}
to obtain robust standard errors for our estimates. 

As the analysis above is done on the seen scenes, it begs the question of whether the same trends occur when agents communicate in unseen environments. To address this, we sample 1,333 agent episodes on the 10 test scenes resulting in a dataset of  201,738 samples. We fit identical logistic regression models to this dataset as in the main paper and report the resulting estimates and standard errors in Table~\ref{table:test-estimates}. While several estimates differ, in a statistically significant way, from those on the seen scenes, all trends remain the same suggesting that agents communicate in largely the same way in unseen environments as they do in previously seen environments.

\begin{table}[h!]
\centering
\begin{small}
\hspace{-2.5mm}
\begin{tabular}{|c|ccc|ccc|}\hline
  &$\beta^{\leq}$ & $\beta_{t}^{\leq}$ & $\beta_{r}^{\leq}$
  &$\beta^{\text{see}}$ &$\beta_{t}^{\text{see}}$ &$\beta_{r}^{\text{see}}$ \\ \hline
  Est.
  & 0.07 & 1.29 & -0.14 &
    0.65 & 0.57 & -0.88 \\
  SE & 0.033 & 0.027 & 0.031 &
       0.041 & 0.027 & 0.042 \\ \hline
\end{tabular}
\end{small}\\
\begin{small}
\begin{tabular}{|c|cccccc|}\hline
&$\beta^{\text{pick}}$ & $\beta_{t,0}^{\text{pick}}$ &$\beta_{r,0}^{\text{pick}}$ & $\beta_{t,1}^{\text{pick}}$
  &$\beta_{r,1}^{\text{pick}}$ & $\beta^{\text{pick}}_{\vee,r}$ \\ \hline
Est &1.15 & -0.0 & -0.04 & -0.01 & -0.04 & -1.17 \\
SE & 0.037 & 0.009 & 0.009 & 0.009 & 0.011 & 0.041  \\\hline
\end{tabular}
\end{small}
\vspace*{-2mm}
\caption{Estimates, and corresponding robust bootstrap standard errors, of the parameters from
  the main paper's Section 4 when using trajectories sampled from the unseen scenes as described in Section \ref{sec:qualitative}.}\label{table:test-estimates}
  \vspace*{-6mm}
\end{table}

\subsection{Qualitative results}
\subsubsection{Effect of communication}
\begin{figure*}
  \centering
  \includegraphics[width=.8\linewidth]{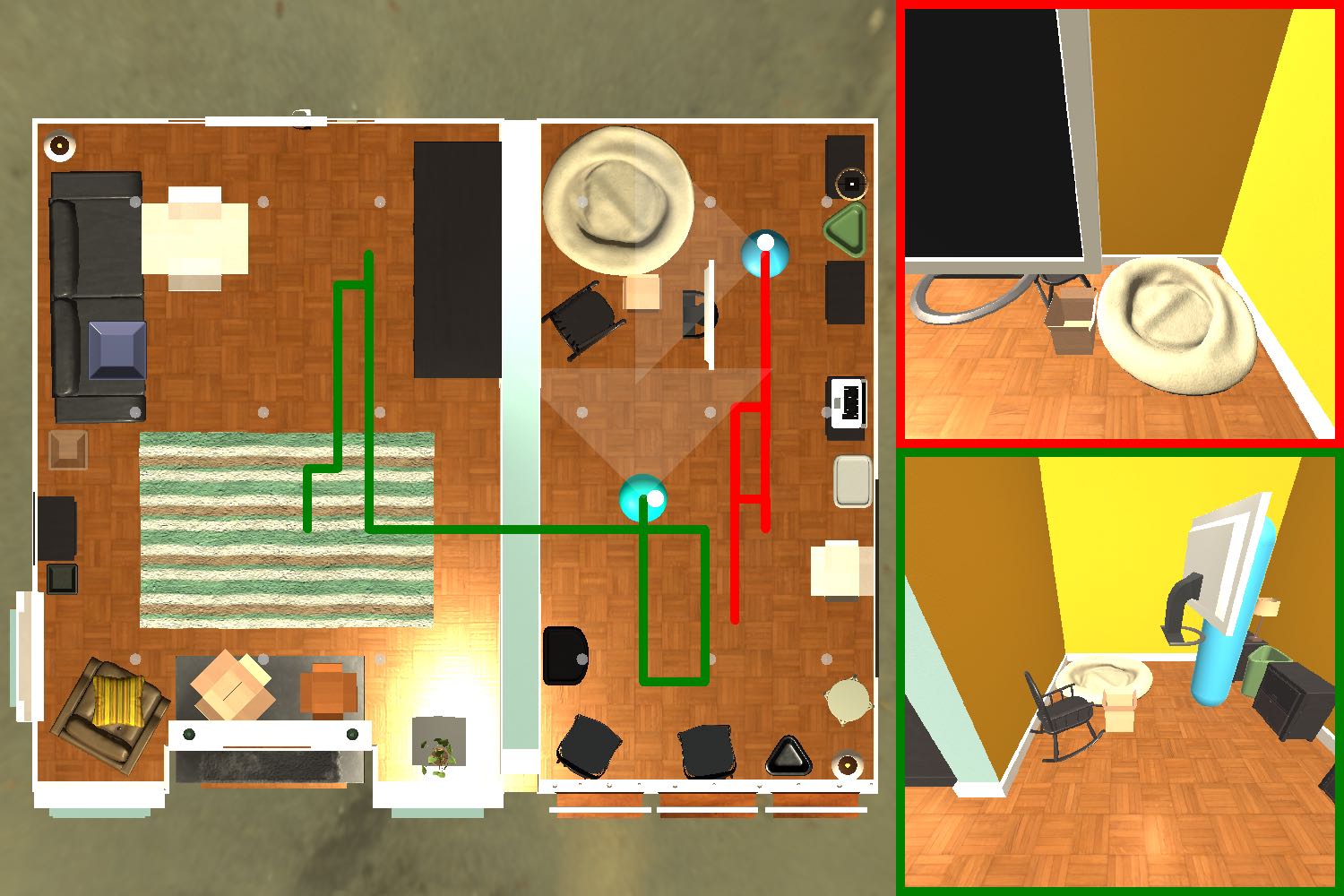} \\
  \includegraphics[width=.8\linewidth]{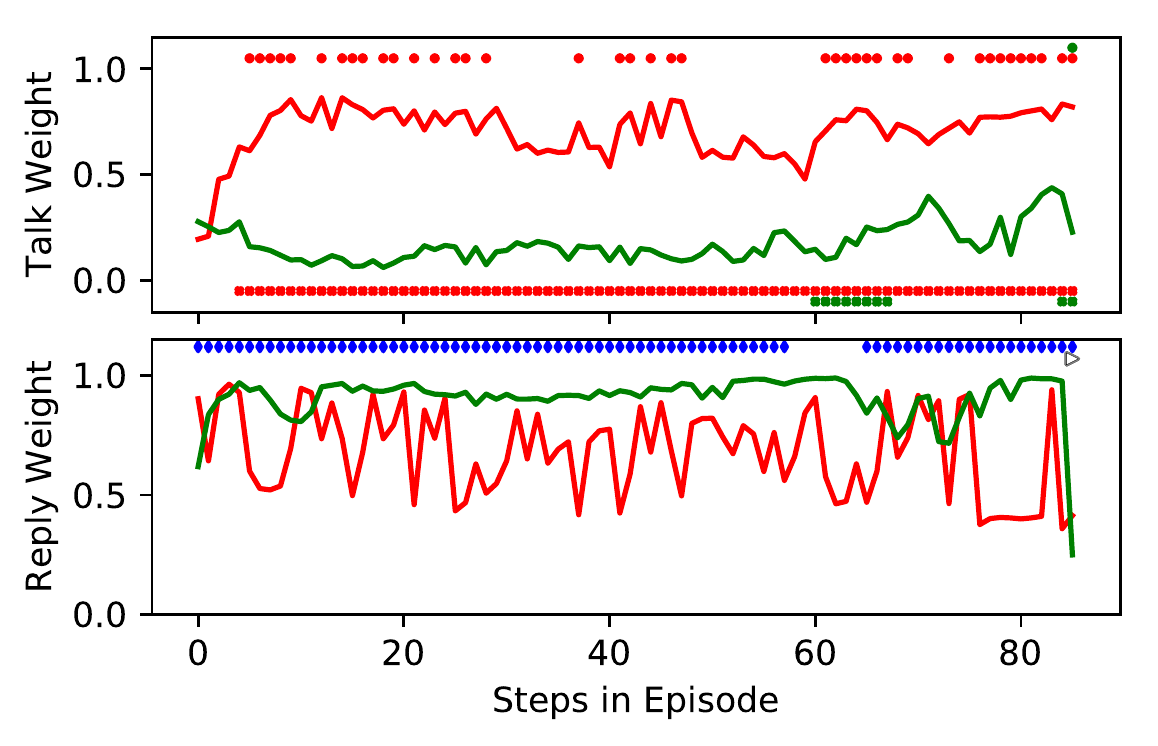}
  \caption{Initialization 1: With explicit and implicit communication, episode length is 86 per agent.
  Associated agent communication in plot below, see Figure 8 in the main paper for a legend.}
  \label{fig:example1a}
\end{figure*}

\begin{figure*}
  \centering
  \includegraphics[width=.8\linewidth]{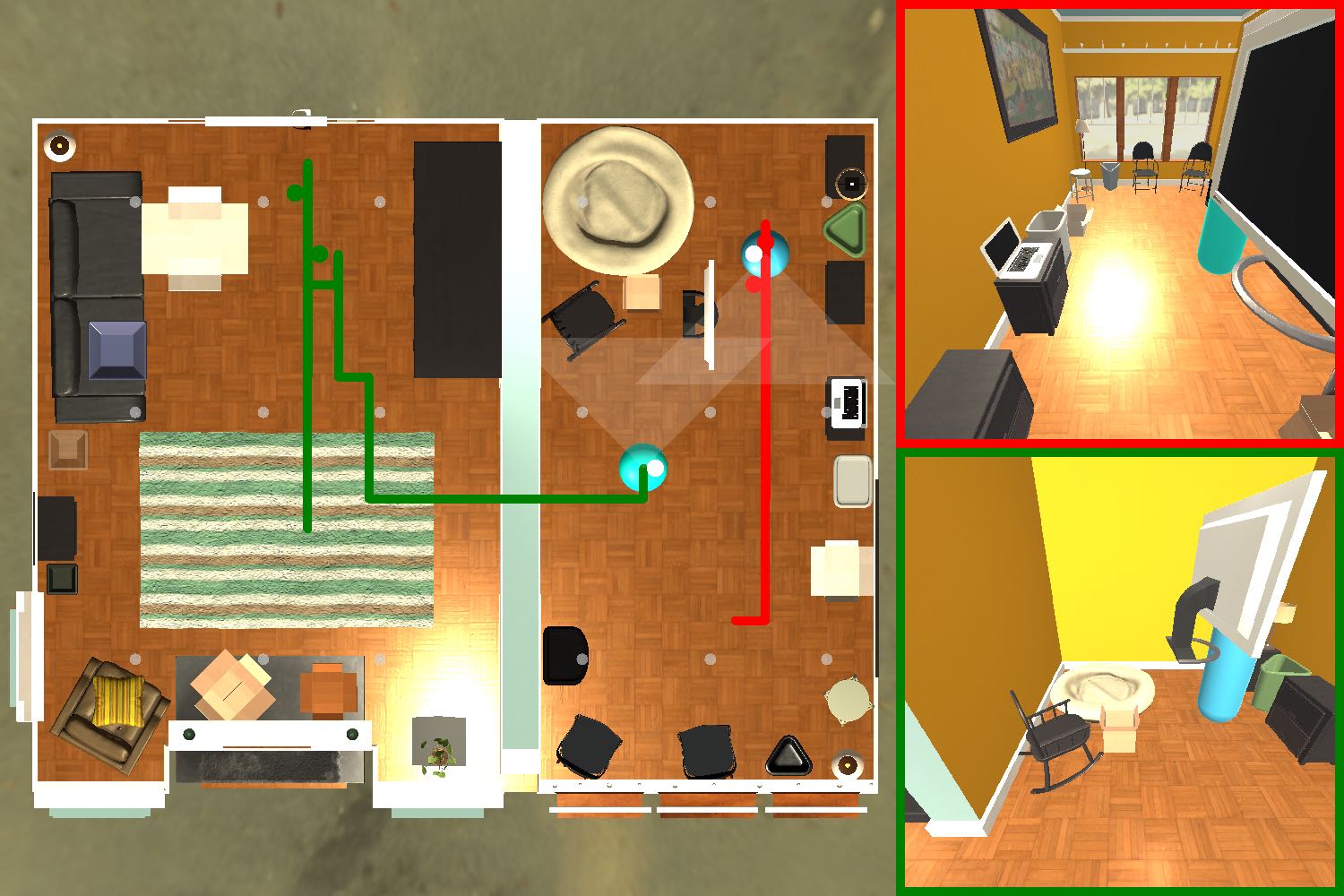}
  \caption{Initialization 1: With only implicit communication, episode length is 165 per agent.}
  \label{fig:example1b}
\end{figure*}

\begin{figure*}
  \centering
  \includegraphics[width=.8\linewidth]{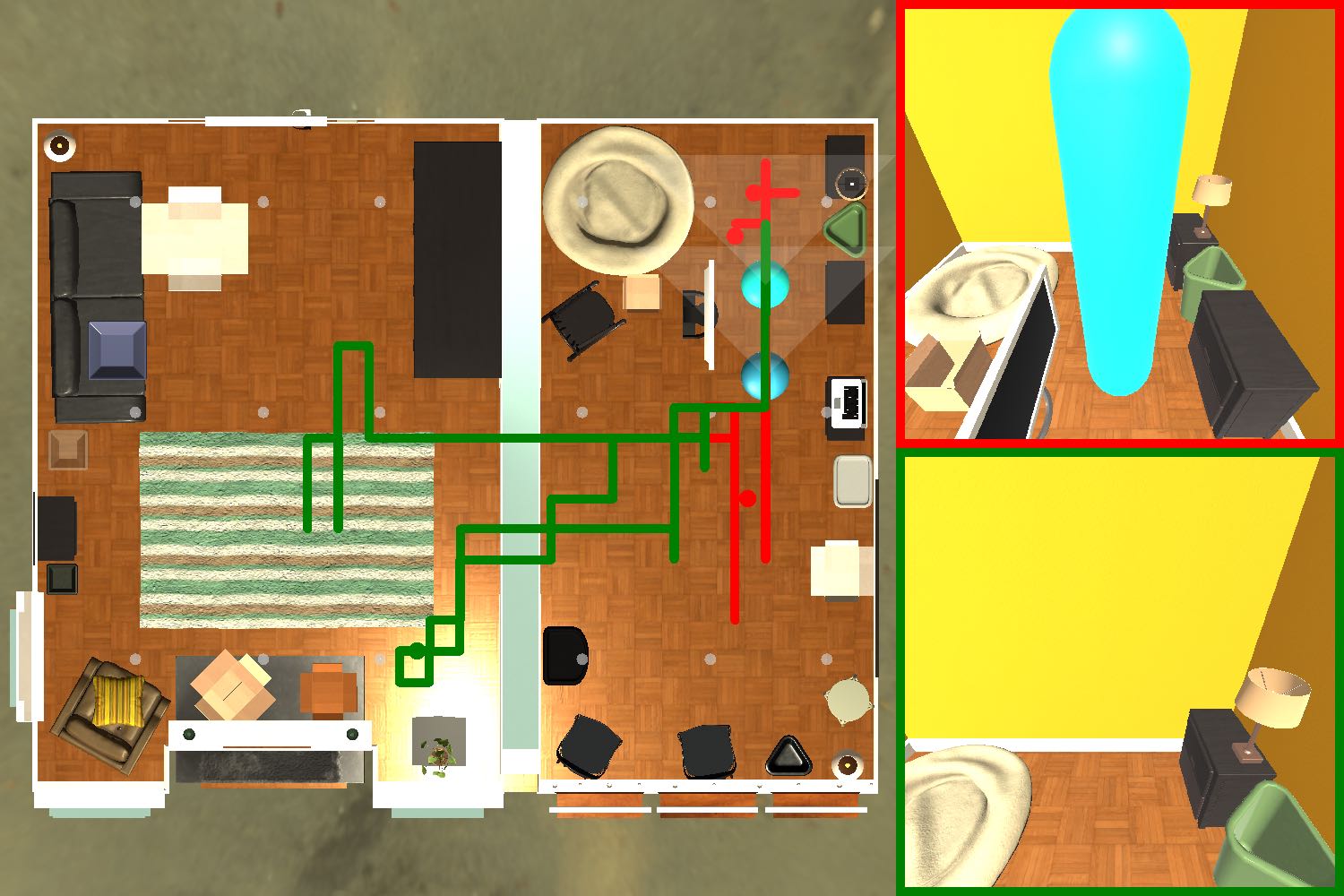}
  \caption{Initialization 1: With no communication, episode length is 250 per agent (unsuccessful).}
  \label{fig:example1c}
\end{figure*}

\begin{figure*}
  \centering
  \includegraphics[width=.8\linewidth]{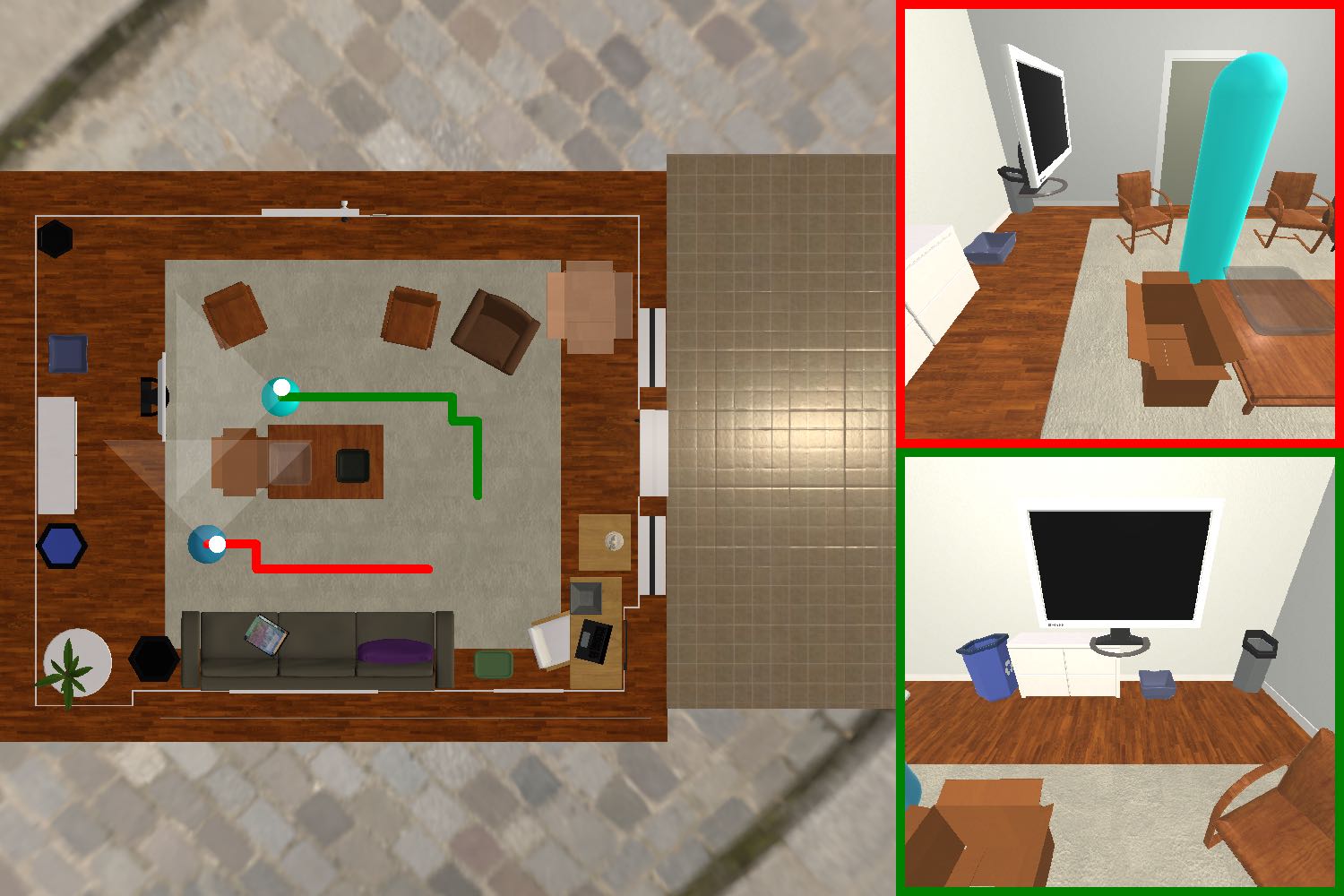} \\
  \includegraphics[width=.8\linewidth]{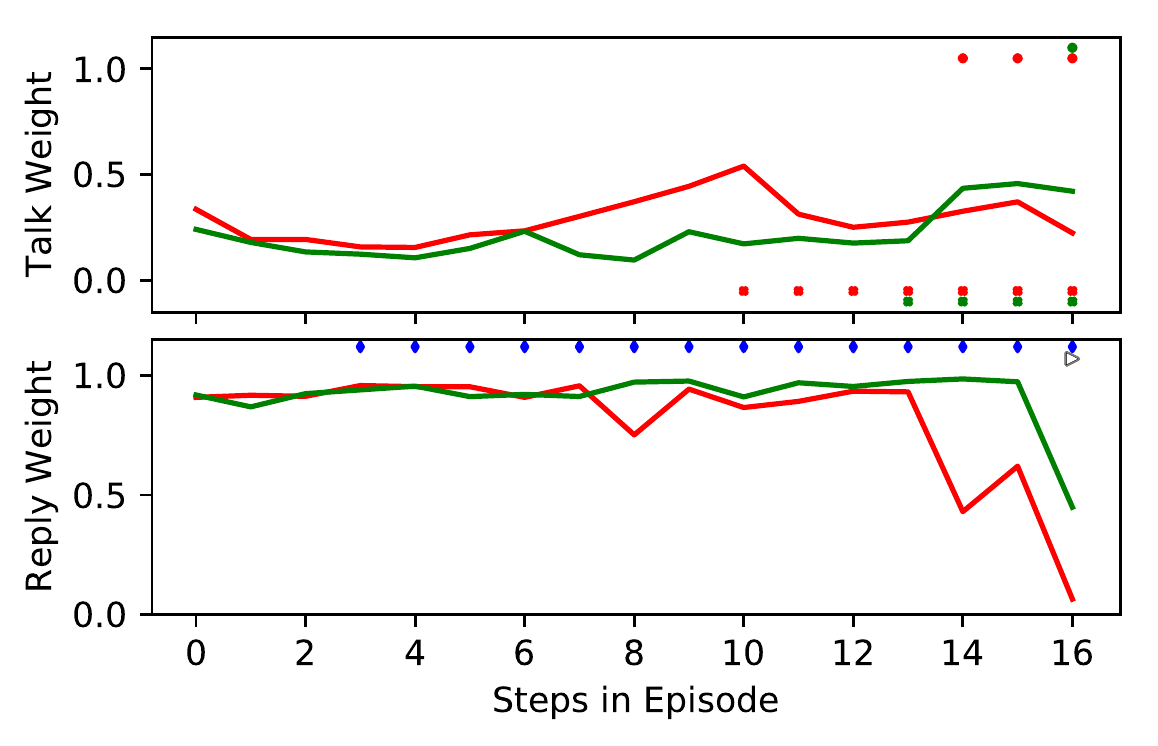}
  \caption{Initialization 2: With explicit and implicit communication, episode length is 17 per agent.
  Associated agent communication in plot below, see Figure 8 in the main paper for a legend.}
  \label{fig:example2a}
\end{figure*}

\begin{figure*}
  \centering
  \includegraphics[width=.8\linewidth]{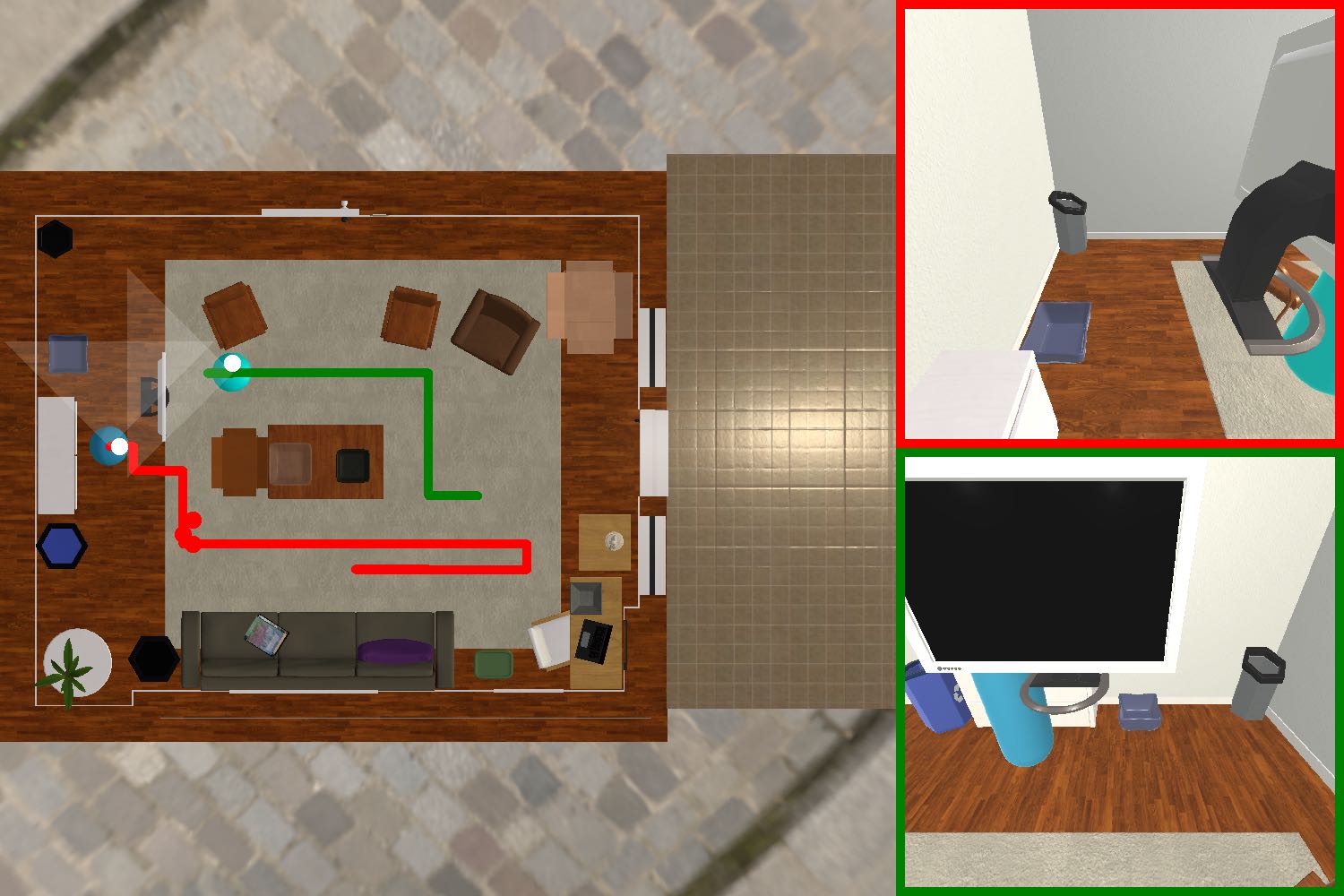}
  \caption{Initialization 2: With only implicit communication, episode length is 72 per agent.}
  \label{fig:example2b}
\end{figure*}

\begin{figure*}
  \centering
  \includegraphics[width=.8\linewidth]{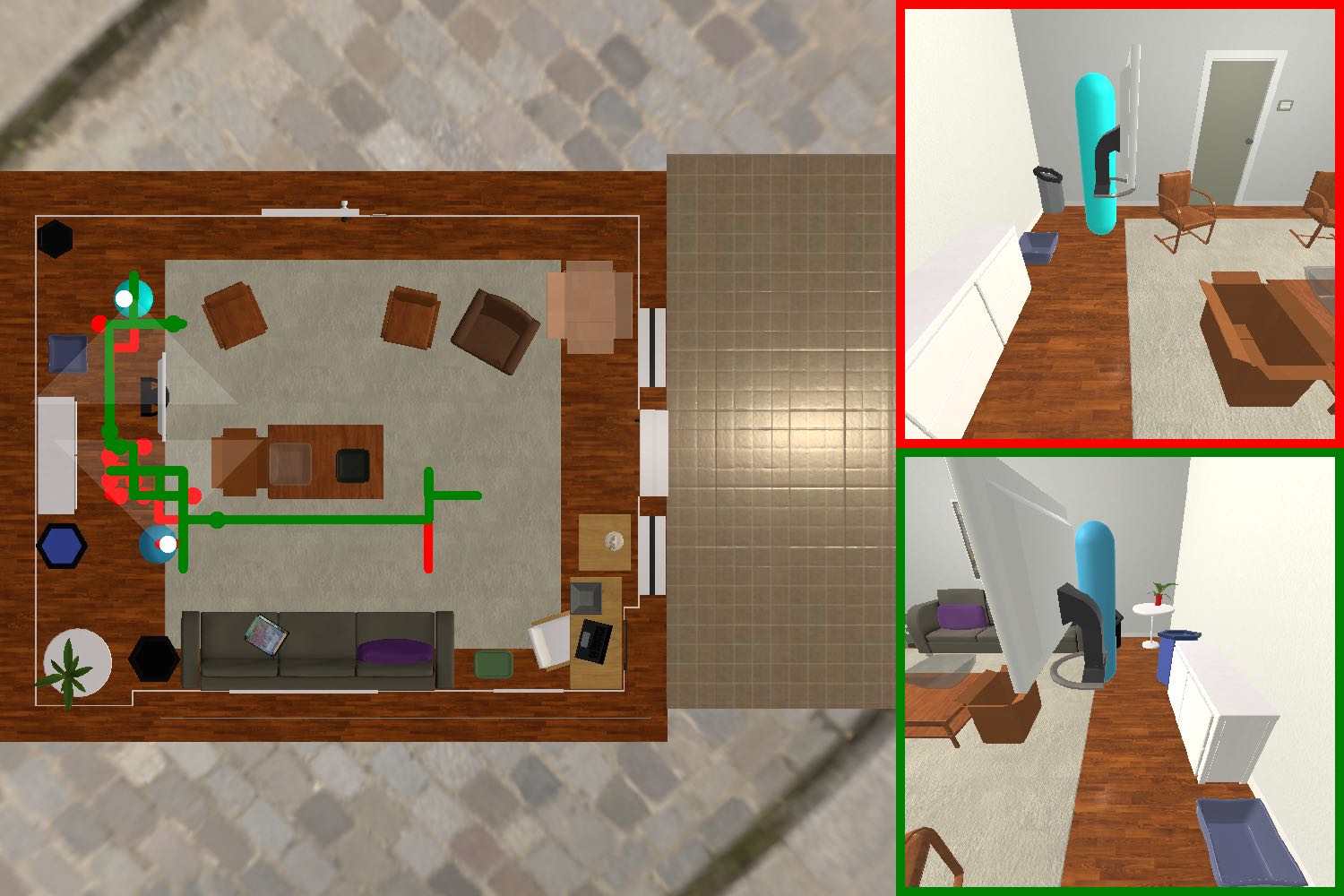}
  \caption{Initialization 2: With no communication, episode length is 217 per agent.}
  \label{fig:example2c}
\end{figure*}

We present qualitative results of agents with three communication abilities: implicit + explicit vs.\ implicit only vs.\ no communication. We compare the effect by deploying this agents for a particular initialization of an episode i.e. the same scene, agents' start locations and target object location. We find both explicit and implicit communication help achieve the task faster as seen \figref{fig:example1a}, \figref{fig:example1b} and \figref{fig:example1c} which have episode lengths of 86, 165 and 250 respectively. Another such initialization is compared in \figref{fig:example2a}, \figref{fig:example2b} and \figref{fig:example2c} which have episode lengths of 17, 72 and 217 respectively.
\subsubsection{Video}
\begin{figure*}
  \centering
  \includegraphics[width=.8\linewidth]{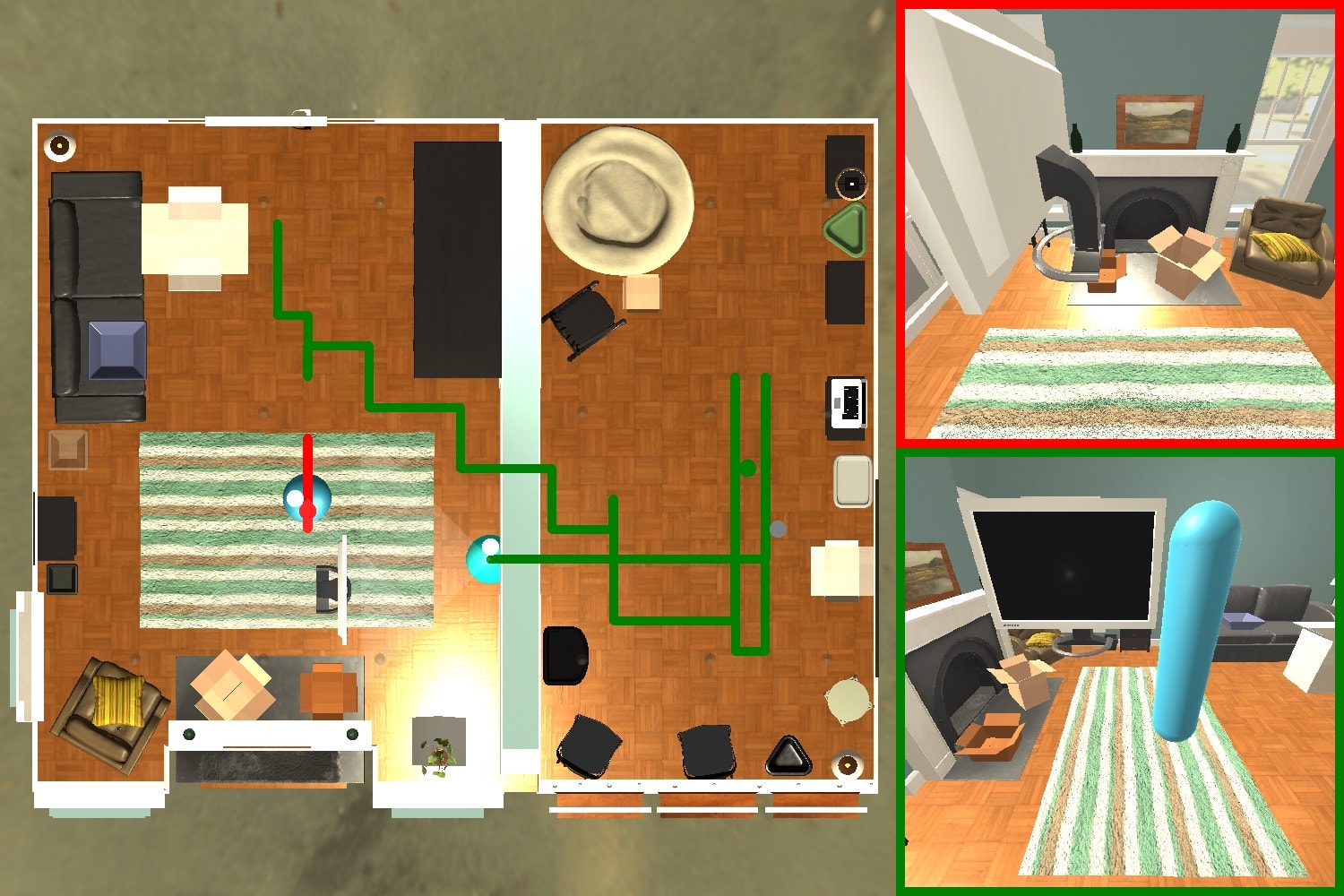} \\
  \includegraphics[width=.8\linewidth]{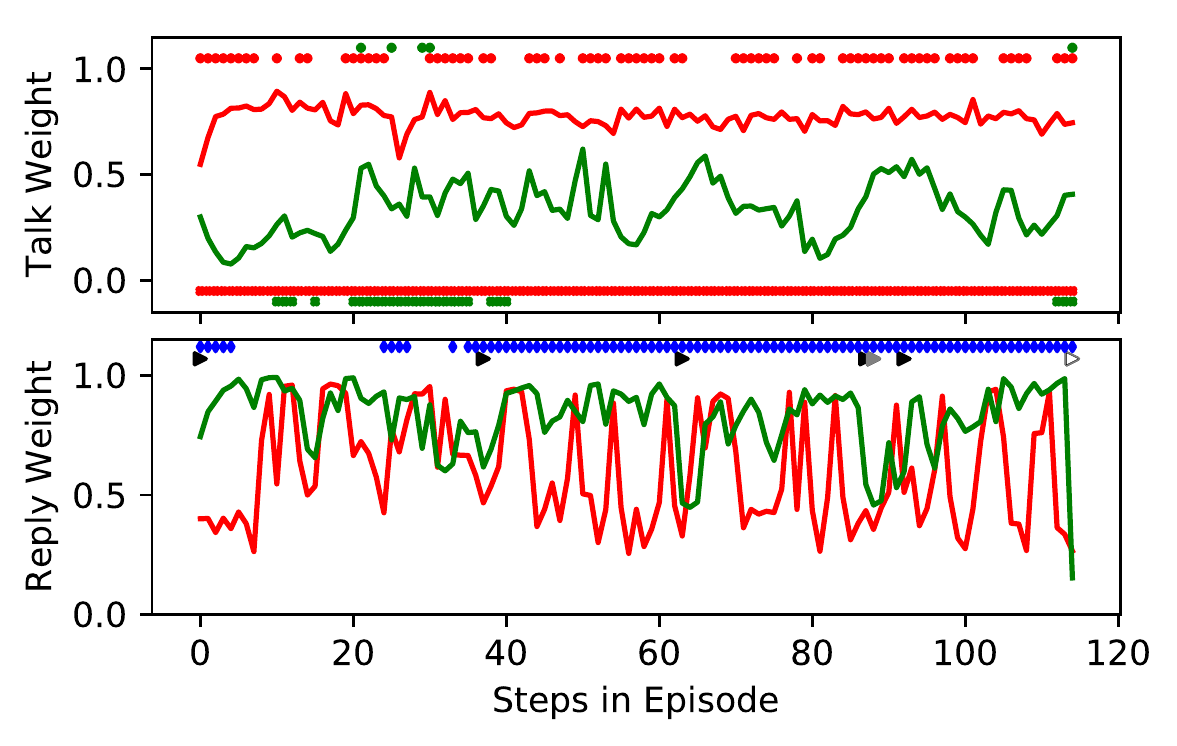}
  \caption{Clip 1 summary, see Figure 8 in the main paper for a legend.}
  \label{fig:clip1}
\end{figure*}

\begin{figure*}
  \centering
  \includegraphics[width=.8\linewidth]{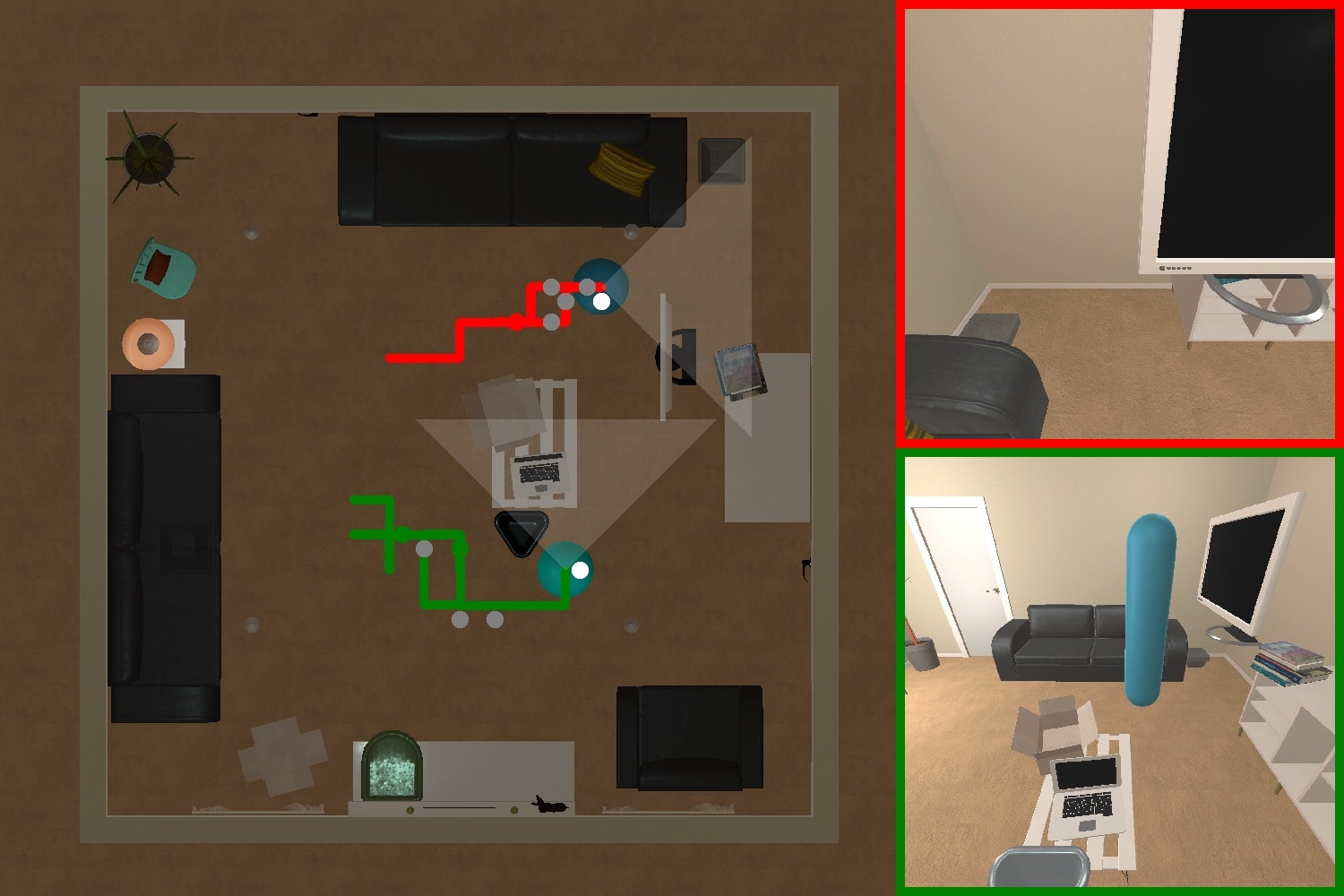} \\
  \includegraphics[width=.8\linewidth]{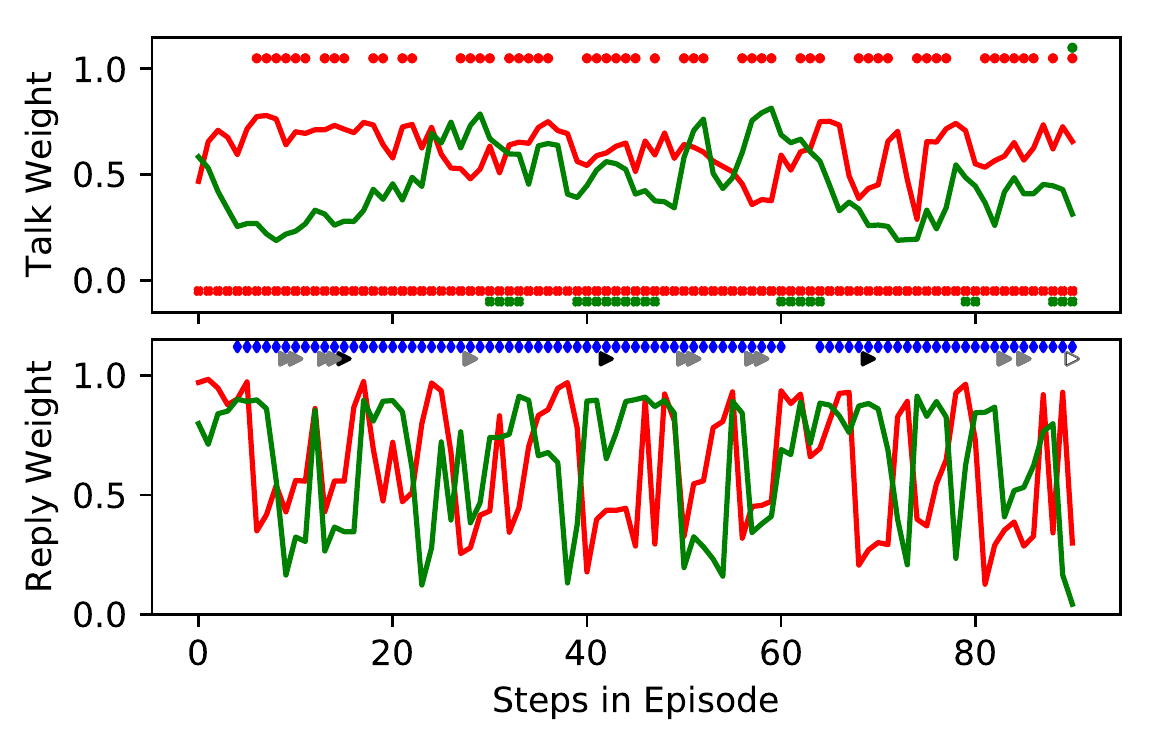}
  \caption{Clip 2 summary, see Figure 8 in the main paper for a legend.}
  \label{fig:clip2}
\end{figure*}

\begin{figure*}
  \centering
  \includegraphics[width=.8\linewidth]{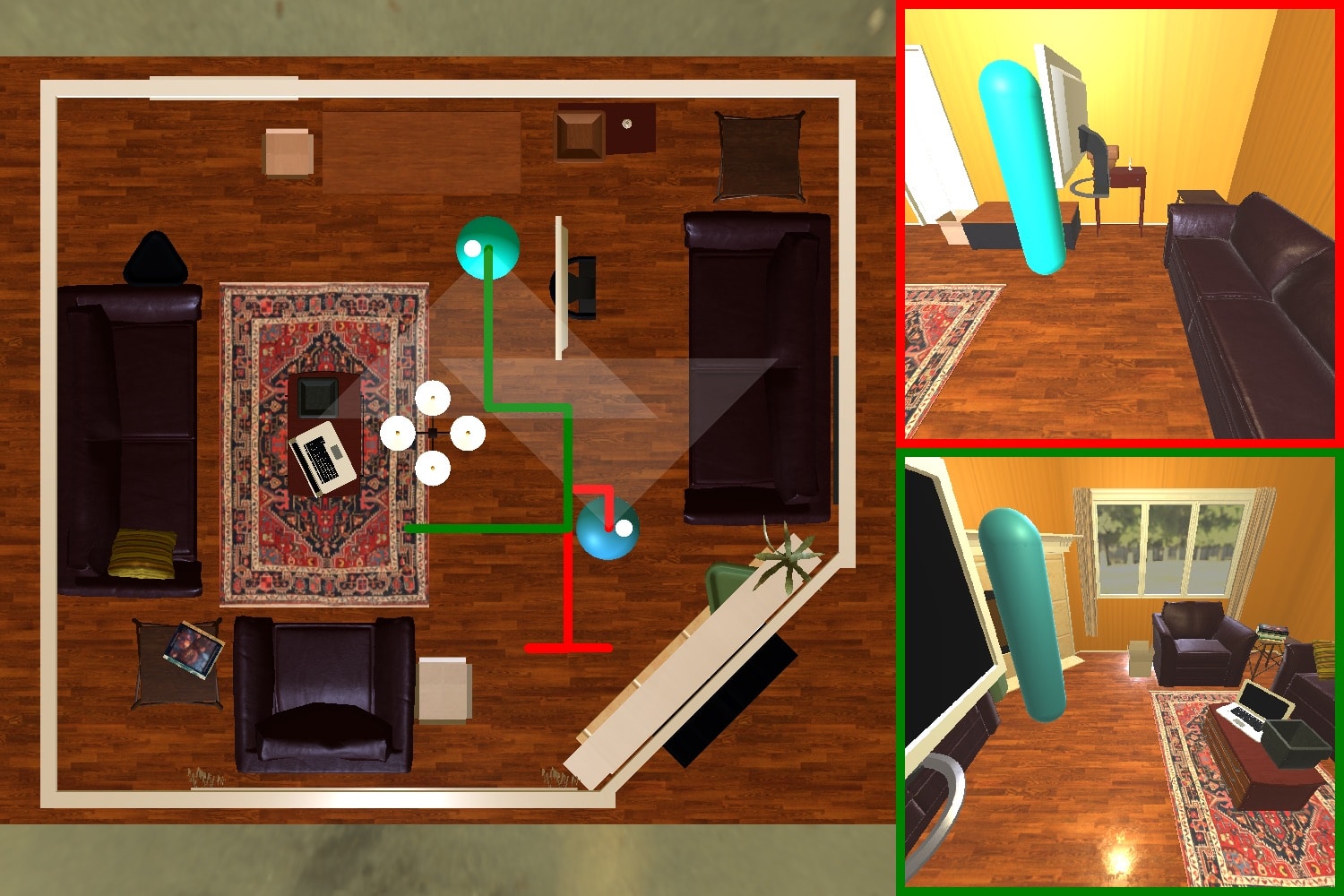} \\
  \includegraphics[width=.8\linewidth]{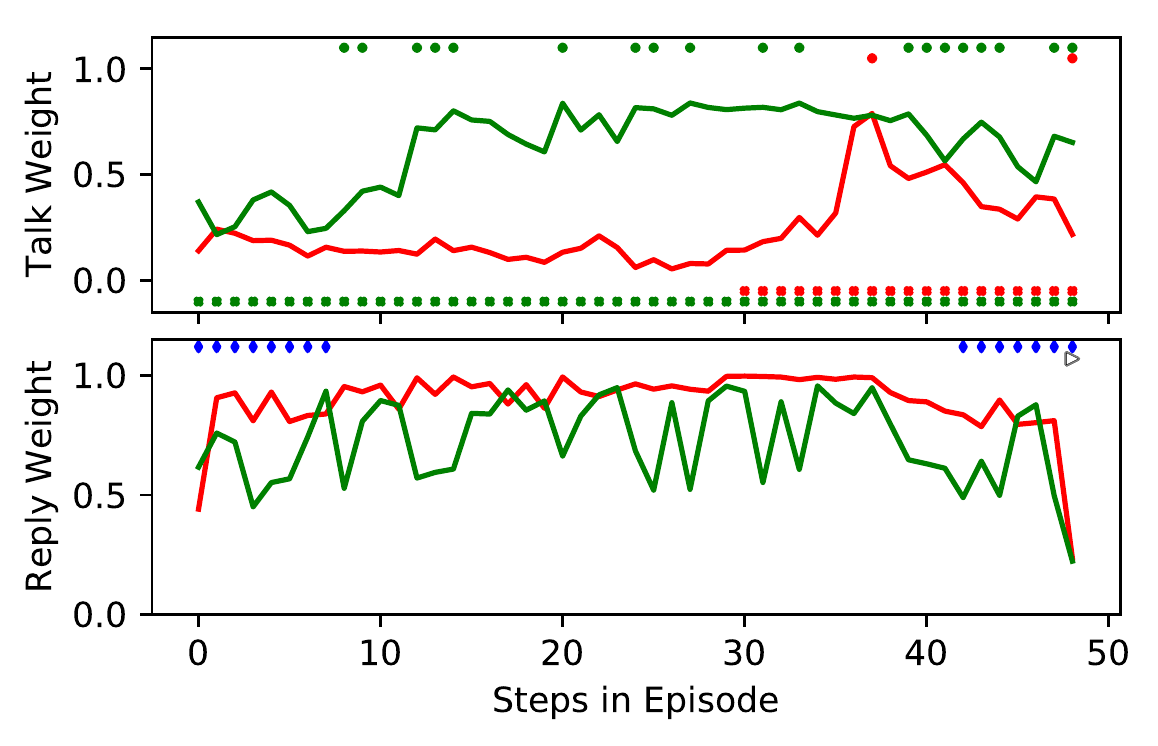}
  \caption{Clip 3 summary, see Figure 8 in the main paper for a legend.}
  \label{fig:clip3}
\end{figure*}

\begin{figure*}
  \centering
  \includegraphics[width=.8\linewidth]{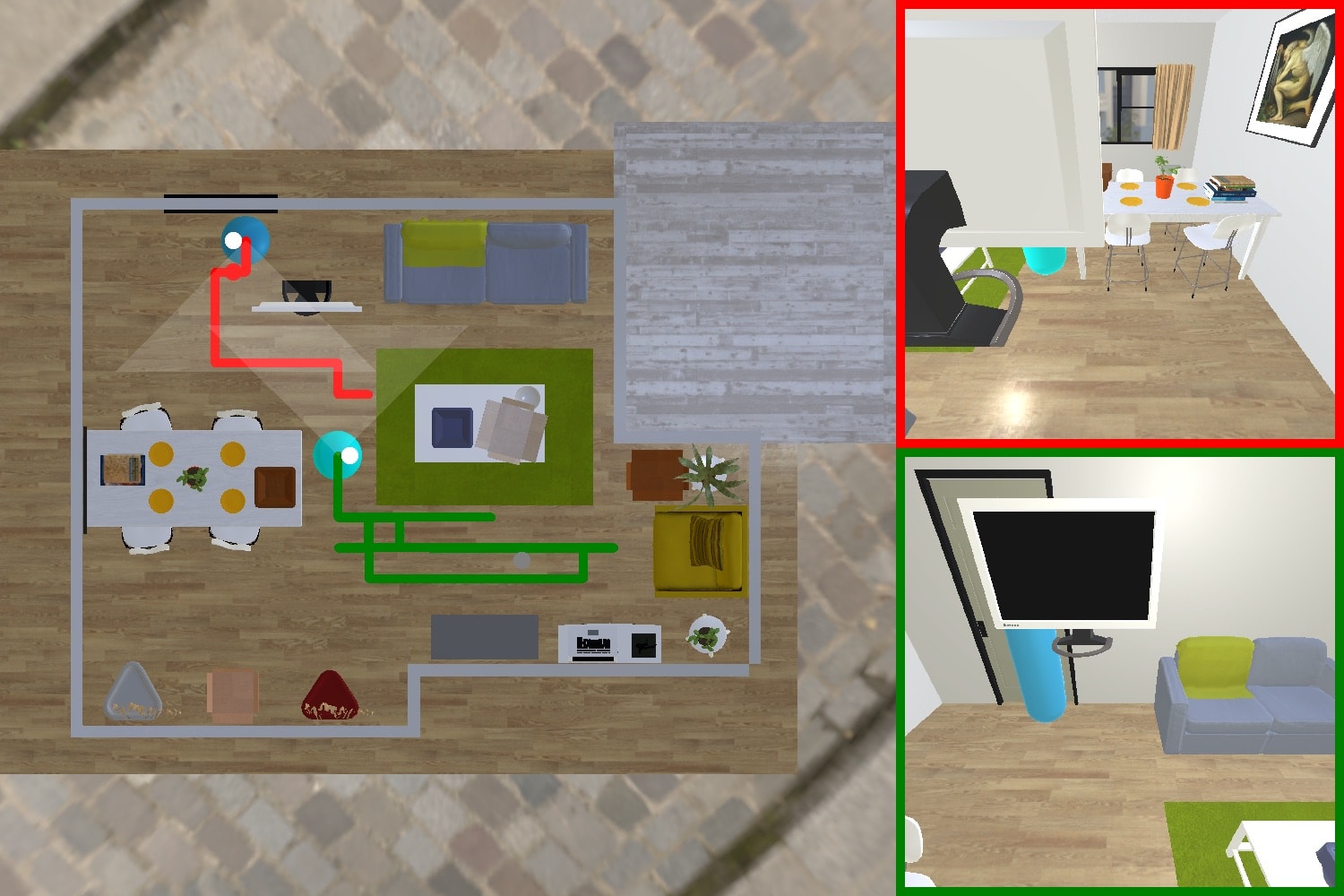} \\
  \includegraphics[width=.8\linewidth]{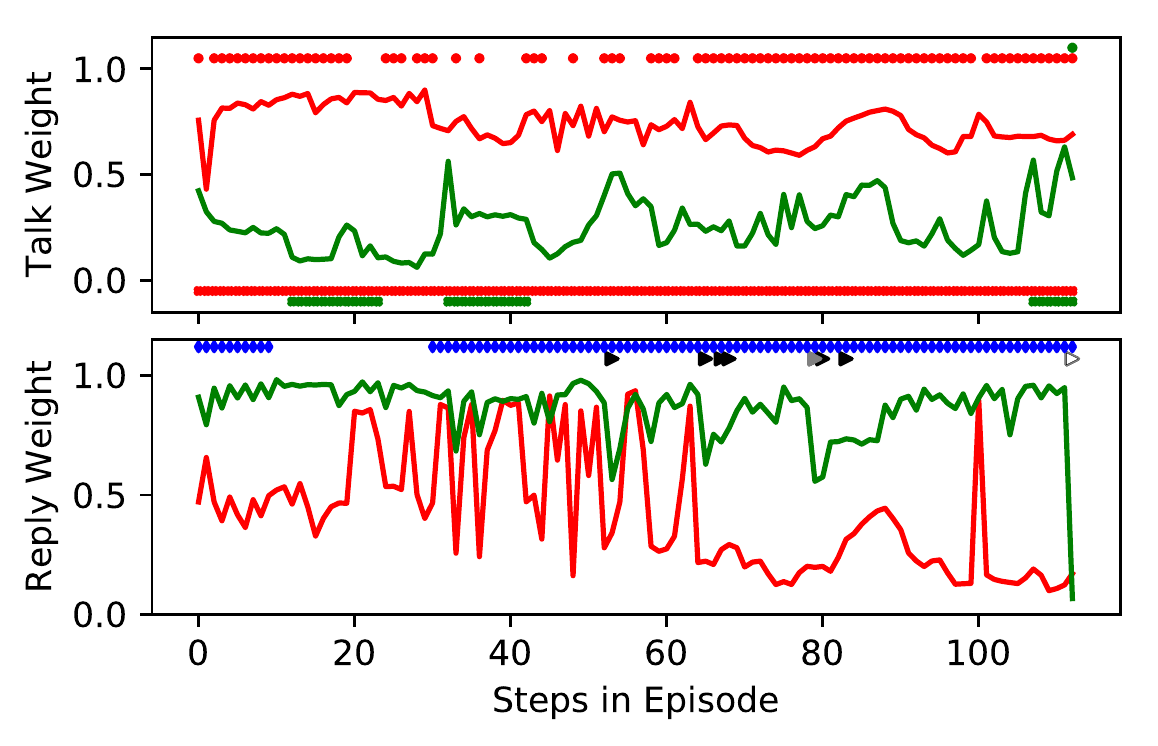}
  \caption{Clip 4 summary, see Figure 8 in the main paper for a legend.}
  \label{fig:clip4}
\end{figure*}

\begin{figure*}
  \centering
  \includegraphics[width=.8\linewidth]{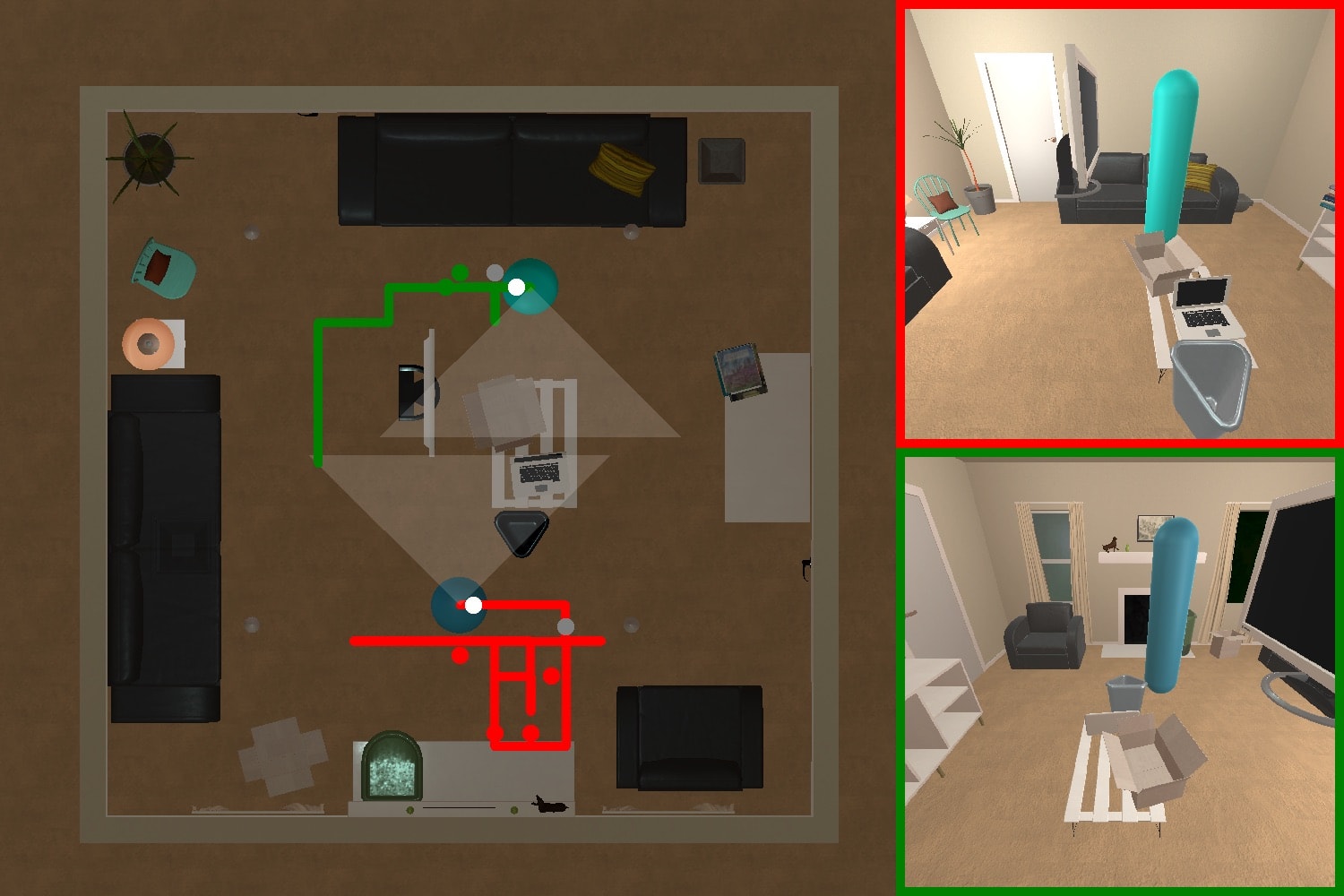} \\
  \includegraphics[width=.8\linewidth]{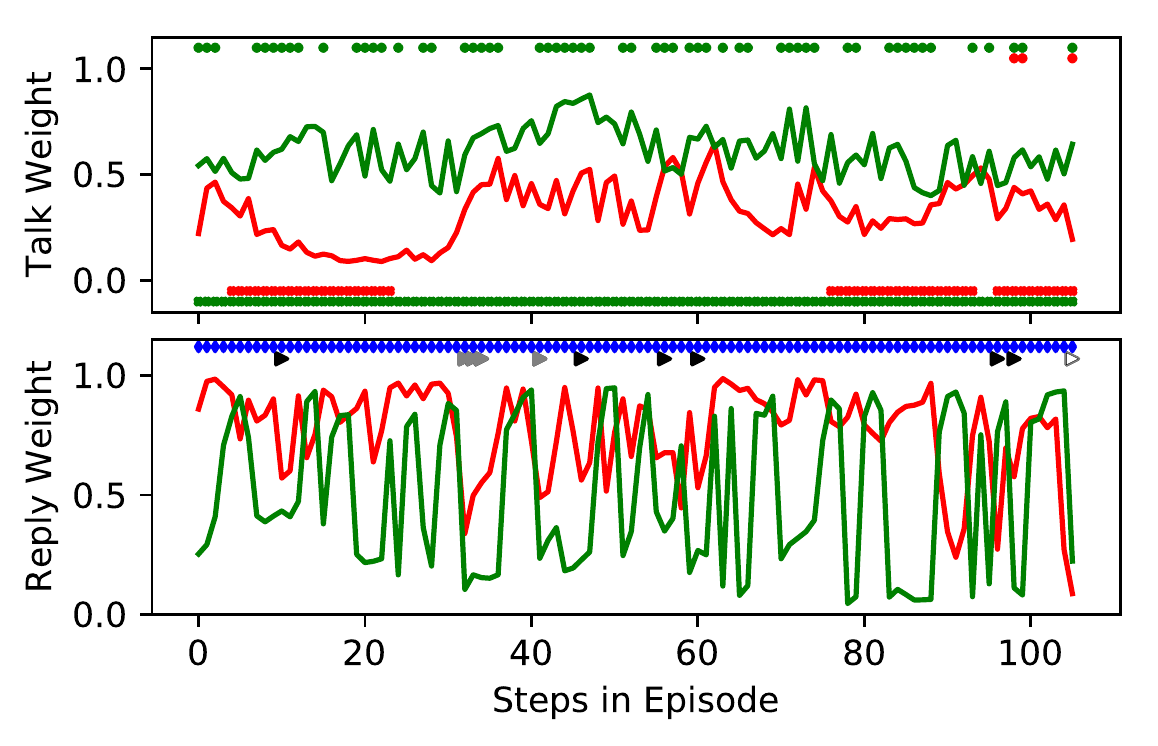}
  \caption{Clip 5 summary, see Figure 8 in the main paper for a legend.}
  \label{fig:clip5}
\end{figure*}

\begin{figure*}
  \centering
  \includegraphics[width=.8\linewidth]{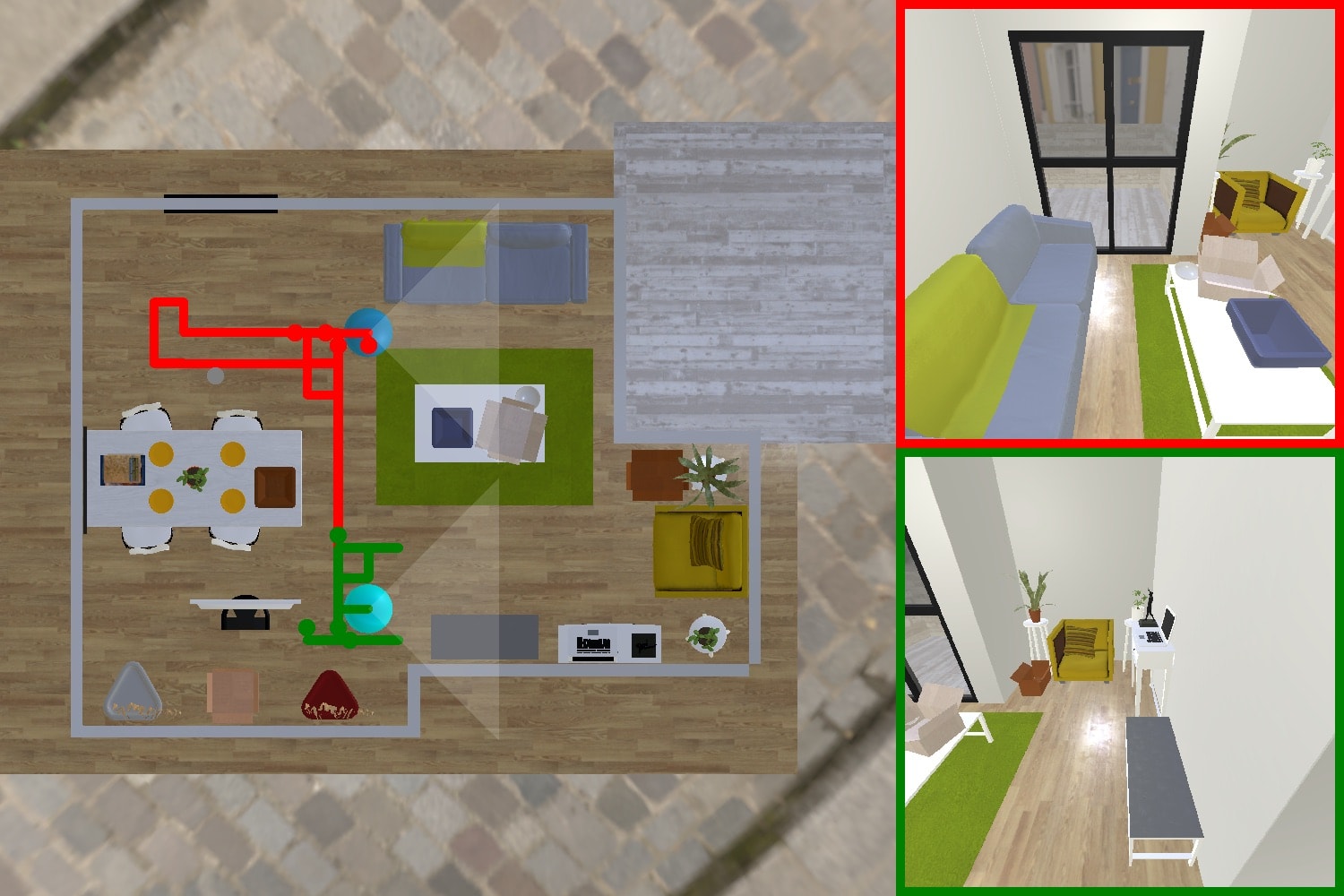} \\
  \includegraphics[width=.8\linewidth]{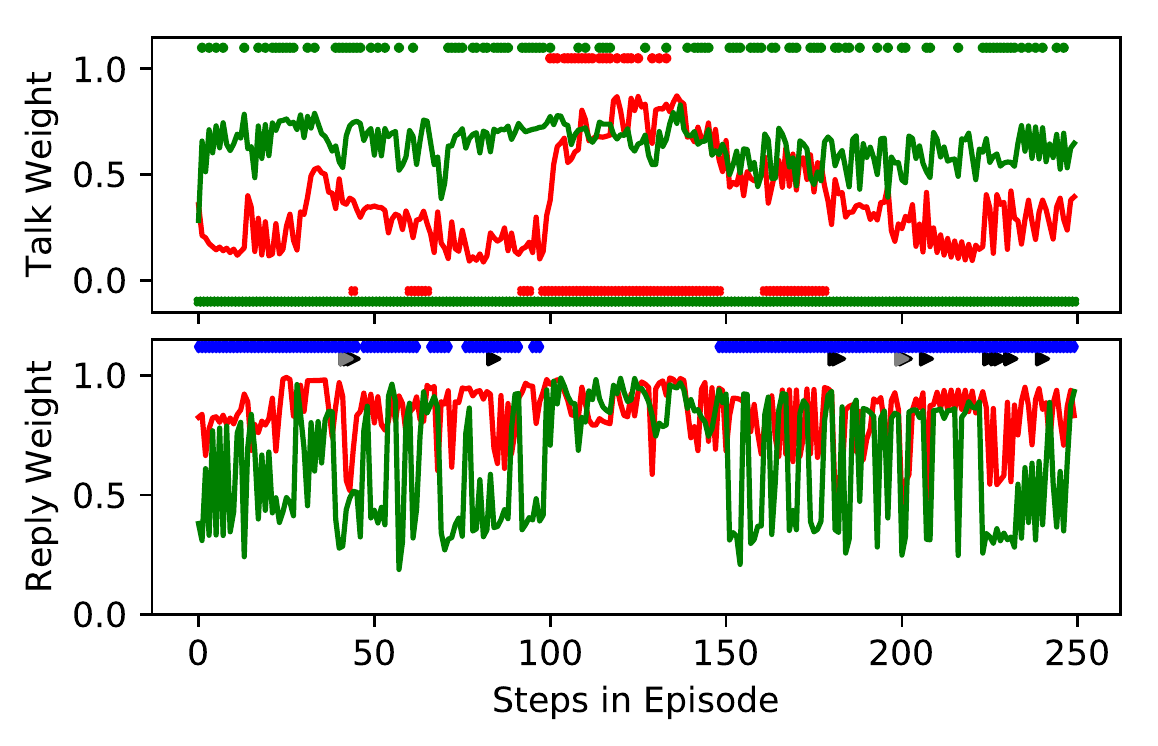}
  \caption{Clip 6 summary, see Figure 8 in the main paper for a legend.}
  \label{fig:clip6}
\end{figure*}

The associated video includes episode visualizations for the \emph{Constrained} task on \textbf{Unseen scenes}, and can be found here: \url{https://youtu.be/9sQhD_Gin5M}. For these episodes we ran inference on the model with both explicit and implicit communication. The six clips in the video are summarized in \figref{fig:clip1}, \figref{fig:clip2}, \figref{fig:clip3}, \figref{fig:clip4}, \figref{fig:clip5} and \figref{fig:clip6}. The first four culminated in successful pickup of the target object. The last two videos highlight typical error modes. 



